\documentclass{svjour3}
\usepackage[noadjust]{cite}
\usepackage{url}
\usepackage[pagebackref=true]{hyperref} 
\usepackage{xcolor}
\usepackage[most]{tcolorbox}
\usepackage{graphics} 
\usepackage{epsfig, psfig} 
\usepackage{amsmath} 
\usepackage{amssymb} 
\usepackage{morefloats}
\usepackage{bchart}
\usepackage{array,multirow,graphicx}
\usepackage{enumitem}
\usepackage{booktabs}

\usepackage{tikz}

\begin{document}

\title{New Performance Measures for Object Tracking under Complex Environments}

\author{Ajoy~Mondal}
\institute{
CVIT,
International Institute of Information Technology, \\
Hyderabad, India \\
email: ajoy.mondal83@gmail.com }
\maketitle

\begin{abstract}
Various performance measures based on the ground truth and without ground truth exist to evaluate the quality of a developed tracking algorithm. The existing popular measures - average center location error ({\sc acle}) and average tracking accuracy ({\sc ata}) based on ground truth, may sometimes create confusion to quantify the quality of a developed algorithm for tracking an object under some complex environments (e.g., scaled or oriented or both scaled and oriented object). In this article,  we propose three new auxiliary performance measures based on ground truth information to evaluate the quality of a developed tracking algorithm under such complex environments. Moreover, one performance measure is developed by combining both two existing measures ({\sc acle} and {\sc ata}) and three new proposed measures for better quantifying the developed tracking algorithm under such complex conditions. Some examples and experimental results conclude that the proposed measure is better than existing measures to quantify one developed algorithm for tracking objects under such complex environments.


\end{abstract}

\section{Introduction} \label{introduction}

Object tracking in a video sequence has been an active research area in computer vision during the last few decades due to its several real-life application~\cite{yilmaz_survey_2006,survey_2013,benchmark_wu2013,smeulders2014visual,li2018deep,fiaz2019handcrafted}. Every year, large number of tracking algorithms have been developed and corresponding papers are published in journal and conferences. The quality of the developed algorithm is compared with state-of-the-art techniques based on some performance measures. When the developed algorithm is compared with state-of-the-art techniques, several questions may be arises. Are there standard sequences that can be
used for evaluation purposes? Is there a standard evaluation protocol? What kind of performance measure should be used? Unfortunately, there are currently no definite answers to all these questions.

However, several performance measures are proposed to quantify the quality of a newly developed tracking algorithm as compared to state-of-the-art techniques~\cite{performance_evaluation_2008,performance_evaluation_2009,vcehovin2016visual,yin2007performance}. Center location error ({\sc cle}) and area overlap ({\sc aol}) are basically considered to evaluate the performance of a tracking algorithm. These two measures are useful
to compare performances of different tracking algorithms. Other measures like precision plot ({\sc pp}) based on center
location error and success plot ({\sc sp}) based on area overlap are also considered to evaluate the performance of the tracking algorithm throughout the complete sequence. All these measures may sometimes be unable or create confusion to quantify
the quality of an algorithm for tracking single object under complex environments (like object with shape deformation,
scaled object, oriented object, both scaled and oriented object). In this article, we propose three new
complementary measures: error in estimated height of optimal tracker ($E_{h}^{t}$), error in estimated width of
optimal tracker ($E_{w}^{t}$) and error in estimated orientation angle of optimal tracker ($E_{\theta}^{t}$); and
another composite measure based on three proposed and two existing measures to evaluate the performance of a tracking
algorithm under such complex environments. Various examples and experimental results conclude that the proposed
composite measure is better to quantify the quality of an algorithm than the existing measures.

\section{Existing Performance Measures based on Ground Truth Information} \label{epmgti}

Two performance measures namely center location error ($E_{C}^{t}$) and area overlap ($E_{A}^{t}$) are popularly considered to quantify a tracking algorithm. They can be calculated as
\begin{eqnarray}
\begin{array}{l}
E_{C}^{t}  =\|C_{g}^{t}-C^{t} \|; \\
\\
E_{A}^{t}=\frac{T_{g}^{t} \cap T^{t}}{T_{g}^{t} \cup T^{t}},
\end{array}
\end{eqnarray}
where $T_{g}^{t}$ is a ground truth rectangular tracker with center $C_{g}^{t}$ at $t^{th}$ frame and $T^{t}$ is
a rectangular tracker with center $C^{t}$ at $t^{th}$ frame generated by a tracking algorithm. Similarly average
center location error ($E_{C}^{avg}$) and average area overlap or average tracking accuracy ($E_{A}^{avg}$) over all
the frames can be calculated as
\begin{eqnarray}
\begin{array}{l}
E_{C}^{avg}  =\frac{1}{N}\sum_{t=1}^{N} \|C_{g}^{t}-C^{t} \|; \\
\\
E_{A}^{avg}=\frac{1}{N}\sum_{t=1}^{N} \frac{T_{g}^{t} \cap T^{t}}{T_{g}^{t} \cup T^{t}},
\end{array}
\end{eqnarray}
if video contains $N$ number of frames. For a good tracking algorithm, $E_{C}^{t}$ and $E_{C}^{avg}$ should
be $0$ whereas $E_{A}^{t}$ and $E_{A}^{avg}$ should be $1$. Some plots (e.g., precision plot and success plot)
based on center location error and area overlap, respectively are considered to evaluate performance of the tracking algorithms.

\paragraph{\textbf{Precision Plot:}} It is considered to measure the overall tracking performance~\cite{Babenko2010,henriques2015high,benchmark_wu2013}. It shows that the percentage of frames where the estimated location of the object is within the given threshold distance with respect to the ground truth.

\paragraph{\textbf{Success Plot:}} To measure the performance of an algorithm on a sequence of frames, we count the
number of successful frames whose overlap $E_{A}^{t}$ is larger than a given threshold value. The success plot shows
the ratios of successful frames at different thresholds varied from $0$ to $1$. Success rate value at a specific threshold
(e.g., $0.5$) may not always be a good representative for evaluating the tracker’s performance~\cite{benchmark_wu2013}.
Instead of that, the area under the curve ({\sc auc}) of each success plot is considered to evaluate the performance of the tracking
algorithm~\cite{benchmark_wu2013}.

Though these measures are popularly considered to evaluate the performance of a tracking algorithm, they may sometimes
unable or create confusion to quantify the quality of a developed algorithm for tracking an object under some
complex environments (e.g., scaled or oriented or both scaled and oriented object). The following section presents
some examples where the existing performance measures fail or create confusion to evaluate the performance of a tracking
algorithm under such complex environments.

\section{Problems of Existing Evaluation Measures for Scaled Object Tracking} \label{peem_scaled_object_tracking}

\begin{figure*}
\centerline{
\psfig{figure=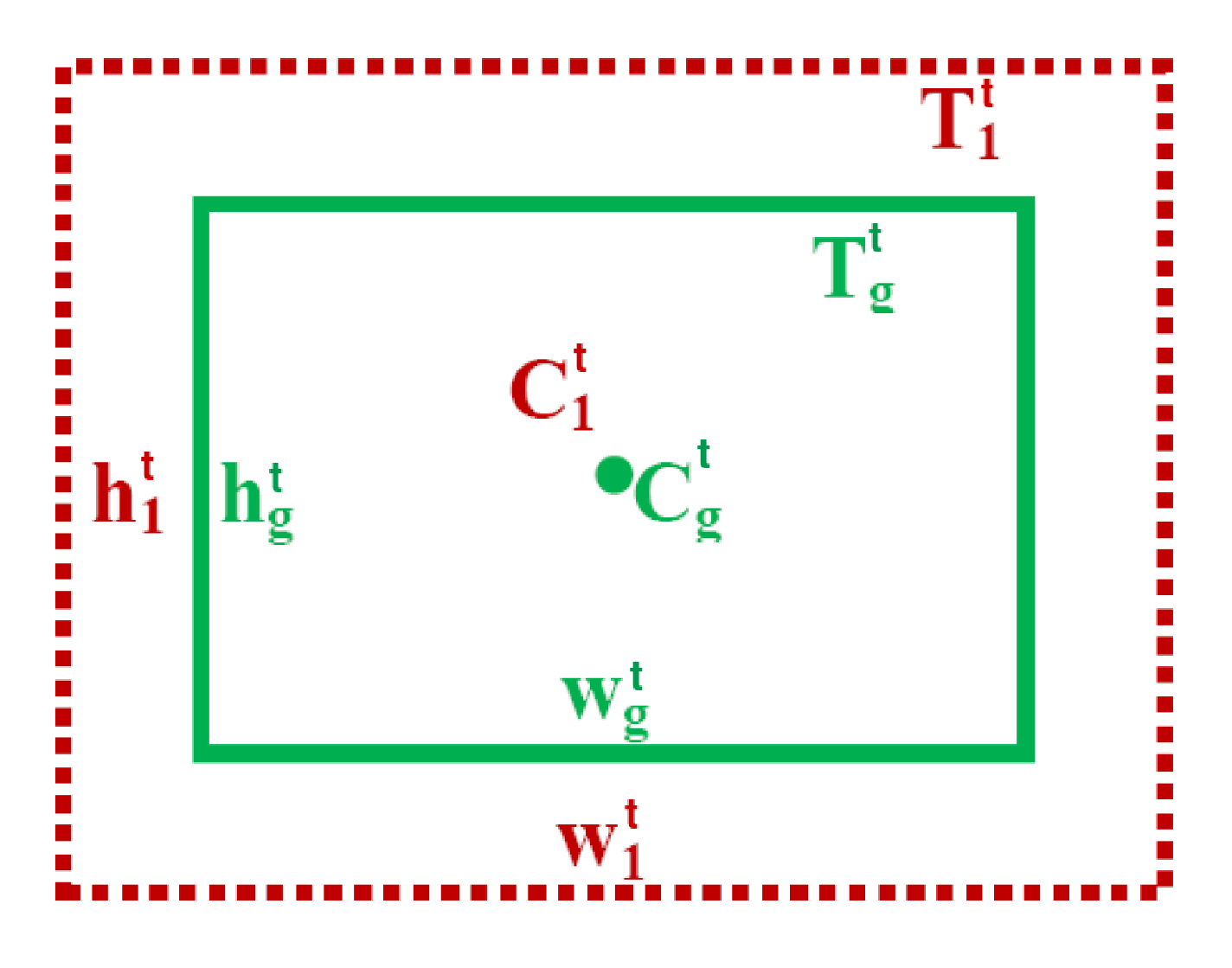,height=0.4\textwidth, width=0.5\textwidth}
\hspace{0.0001\textwidth}
\psfig{figure=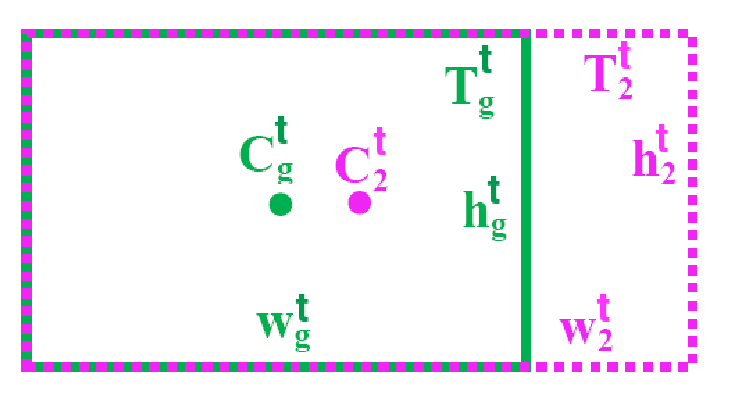,height=0.4\textwidth, width=0.5\textwidth}}
\centerline{(a) \hspace{0.5\textwidth} (b)}
\caption{\textbf{Example 1 -} area overlap between ground truth tracker and trackers obtained by
(a) Alg$_{1}$ and (b) Alg$_{2}$. \label{example1}}
\end{figure*}

\subsection{Example1:} Here, we consider a simple example (see Figure~\ref{example1}). Suppose $T_{g}^{t}$
(green colored rectangle) with center $C_{g}^{t}$, height $h_{g}^{t}$ and width $w_{g}^{t}$ be the
ground truth tracker. Again let $T_{1}^{t}$ (red colored rectangle) with center $C_{1}^{t}$,
height $h_{1}^{t}$ and width $w_{1}^{t}$ and $T_{2}^{t}$ (pink colored rectangle) with center $C_{2}^{t}$,
height $h_{2}^{t}$ and width $w_{2}^{t}$ be two trackers generated by two different (scaled object) tracking algorithms
Alg$1$ and Alg$2$. Here, we also assume that the trackers $T_{g}^{t}$ and $T_{1}^{t}$ have same center locations but
different heights and widths (see Figure~\ref{example1}(a)). Therefore, we have $C_{g}^{t}=C_{1}^{t}$, $h_{g}^{t}< h_{1}^{t}$
and $w_{g}^{t}< w_{1}^{t}$. Now let $w_{1}^{t}=w_{g}^{t}+w^{t}$ and $h_{1}^{t}=h_{g}^{t}+h^{t}$. We also assume that
the trackers $T_{g}^{t}$ and $T_{2}^{t}$ have same heights but different center locations and widths
(see Figure~\ref{example1}(b)). Therefore, we have $C_{g}^{t}<C_{2}^{t}$, $h_{g}^{t}=h_{2}^{t}$ and
$w_{g}^{t}<w_{2}^{t}$. Now let $C_{2}^{t}=C_{g}^{t}+\alpha^{t}$ and $w_{2}^{t}=w_{g}^{t}+w^{t}$.

Now area overlap ($E_{A_{1}}^{t}$) between ground truth tracker $T_{g}^{t}$ and tracker $T_{1}^{t}$ generated
by Alg$1$ at $t^{th}$ frame, is calculated as
\begin{eqnarray}
\begin{array}{l}
E_{A_{1}}^{t}  = \frac{{T_g^{t}  \cap T_1^{t} }}{{T_g^{t}  \cup T_1^{t} }}
        = \frac{{w_g^{t} h_g^{t} }}{{\left( {w_g^{t}  + w^{t}} \right)\left( {h_g^{t}  + h^{t}} \right)}} \\
        = \frac{{w_g^{t} h_g^{t} }}{{w_g^{t} h_g^{t}  + h^{t} w_g^{t}  + w^{t} h_g^{t}  + w^{t} h^{t}}} 
        = \frac{{w_g^{t} h_g^{t} }}{{s^{t} + h^{t} w_g^{t}  + w^{t} h^{t}}}, \label{examp1_scale_area1}
\end{array}
\end{eqnarray}
where $s^{t}=w_g^{t} h_g^{t} + w^{t} h_{g}^{t}$.

Again area overlap ($E_{A_{2}}^{t}$) between ground truth tracker $T_{g}^{t}$ and tracker $T_{2}^{t}$ generated
by Alg$2$ at $t^{th}$ frame, is calculated as
\begin{eqnarray}
\begin{array}{l}
E_{A_{2}}^{t}  = \frac{{T_g^{t}  \cap T_2^{t} }}{{T_g^{t}  \cup T_2^{t} }}
               = \frac{{ w_g^{t} h_{g}^{t} }} {{w_2^{t} h_2^{t} }}
               = \frac{{w_g^{t} h_g^{t} }}{{\left( {w_g^{t}  + w^{t}} \right)\left( {h_g^{t} } \right)}}
               = \frac{{ w_g^{t} h_{g}^{t} }} {{ s^{t} }}, \label{examp1_scale_area2}
\end{array}
\end{eqnarray}
since $s^{t}=w_g^{t} h_g^{t} + w^{t} h_{g}^{t}$.

Now center location errors $E_{C_{1}}^{t}$ and $E_{C_{2}}^{t}$ of trackers $T_{1}^{t}$ and $T_{2}^{t}$ are calculated as
\begin{eqnarray}
\begin{array}{l}
 E_{C_{1}}^{t}  = \left\| {C_{g}^{t}  - C_{1}^{t} } \right\|
                = 0, \label{examp1_scale_center1}
 \end{array}
\end{eqnarray}
since $C_{g}^{t}=C_{1}^{t}$.

\begin{eqnarray}
\begin{array}{l}
 E_{C_{2}}^{t}  = \left\| {C_{g}^{t}  - C_{2}^{t} } \right\|
  = \left\| {C_{g}^{t}  - (C_{g}^{t}+\alpha^{t}) } \right\|
  = \alpha^{t}. \label{examp1_scale_center2}
\end{array}
\end{eqnarray}

Eqs. (\ref{examp1_scale_area1}) and (\ref{examp1_scale_area2}) conclude that $E_{A_{2}}^{t}>E_{A_{1}}^{t}$,
i.e., tracker $T_{2}^{t}$ has more area overlap with ground truth tracker $T_{g}^{t}$ than tracker $T_{1}^{t}$
at $t^{th}$ frame. On the other hand, Eqs. (\ref{examp1_scale_center1}) and (\ref{examp1_scale_center2}) conclude
that $E_{c_{1}}^{t}< E_{C_{2}}^{t}$, i.e., tracker $T_{1}^{t}$ has lesser center location error than
tracker $T_{2}^{t}$ at $t^{th}$ frame. If we consider center location error as a performance measure then
tracker $T_{1}^{t}$ is better than tracker $T_{2}^{t}$ at $t^{th}$ frame. If we consider area overlap as a
performance measure then tracker $T_{2}^{t}$ is better than tracker $T_{1}^{t}$ at $t^{th}$ frame. If we consider both these measures no decision can be made. However, tracker $T_{1}^{t}$ has lesser center location
error with lesser area overlap and tracker $T_{2}^{t}$ has higher center location error with higher area overlap
at $t^{th}$ frame. Therefore, minimum center location error indicates maximum area overlap for the fixed sized tracker is not true for the scaled tracker and performance measures: center location error and area pverlap individually can make ambiguous decision for the scaled tracker.

\subsection{Example2:}

\begin{figure*}
\centerline{
\psfig{figure=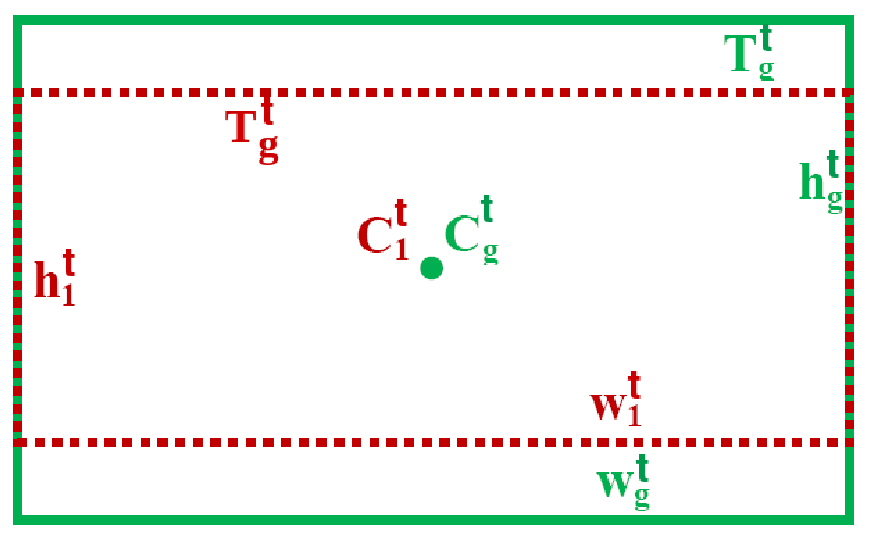,height=0.4\textwidth, width=0.5\textwidth}
\hspace{0.0001\textwidth}
\psfig{figure=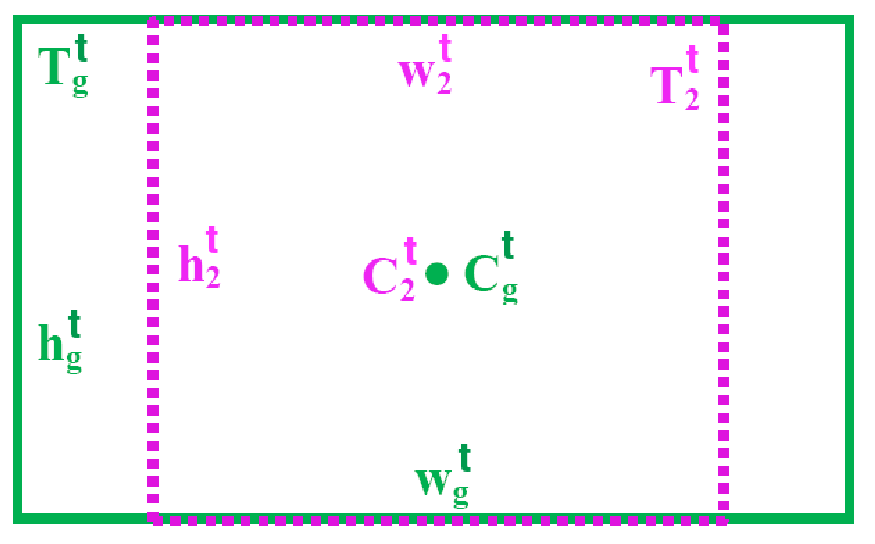,height=0.4\textwidth, width=0.5\textwidth}}
\centerline{a \hspace{0.5\textwidth} b}
\caption{\textbf{Example 2 -} area overlap between ground truth tracker and tracker obtained by
(a) Alg$1$ (b) Alg$2$. \label{example2}}
\end{figure*}

Here, we consider another simple example (see Figure~\ref{example2}). Suppose $T_{g}^{t}$ (green colored rectangle)
with center $C_{g}^{t}$, height $h_{g}^{t}$ and width $w_{g}^{t}$ be the ground truth tracker at $t^{th}$ frame.
Again let $T_{1}^{t}$ (red colored rectangle) with center $C_{1}^{t}$, height $h_{1}^{t}$ and width $w_{1}^{t}$
and $T_{2}^{t}$ (pink colored rectangle) with center $C_{2}^{t}$, height $h_{2}^{t}$ and width $w_{2}^{t}$ be two
trackers generated by two different (scaled object) tracking algorithms Alg$1$ and Alg$2$, respectively. Here, we also
assume that the trackers $T_{g}^{t}$ and $T_{1}^{t}$ have same center location and width but different height
(see Figure~\ref{example2}(a)). Therefore, we have $C_{g}^{t}=C_{1}^{t}$, $w_{g}^{t}=w_{1}^{t}$ and $h_{g}^{t}>h_{1}^{t}$.
Now let $h_{g}^{t}=h_{1}^{t}+h^{t}$. We also assume that the trackers $T_{g}^{t}$ and $T_{2}^{t}$ have same center
location and height but different width (see Figure~\ref{example2}(b)). Therefore, we have $C_{g}^{t}=C_{2}^{t}$,
$h_{g}^{t}=h_{2}^{t}$ and $w_{g}^{t}>w_{2}^{t}$. Now let $w_{g}^{t}=w_{2}^{t}+w^{t}$.

Now area overlap ($E_{A_{1}}^{t}$) between ground truth tracker $T_{g}^{t}$ and tracker $T_{1}^{t}$ generated by
Alg$1$ at $t^{th}$ frame, is calculated as
\begin{eqnarray}
\begin{array}{l}
E_{A_{1}}^{t}  = \frac{{T_g^{t}  \cap T_1^{t} }}{{T_g^{t}  \cup T_1^{t} }}
               =\frac{{h_{1}^{t} w_{1}^{t} }} {{ h_{g}^{t} w_{g}^{t} }}
               =\frac{{(h_{g}^{t}-h^{t}) w_{g}^{t} }} {{ h_{g}^{t} w_{g}^{t} }}
               =1-\frac{h^{t}}{{h_g^{t}}}. \label{examp2_scale_area1}
\end{array}
\end{eqnarray}

Again area overlap ($E_{A_{2}}^{t}$) between the ground truth tracker $T_{g}^{t}$ and tracker $T_{2}^{t}$ generated
by Alg$2$ at $t^{th}$ frame, is calculated as
\begin{eqnarray}
\begin{array}{l}
E_{A_{2}}^{t}  = \frac{{T_g^{t}  \cap T_2^{t} }}{{T_g^{t}  \cup T_2^{t} }}
               = \frac{{ w_{2}^{t} h_{2}^{t} }} {{w_{g}^{t} h_{g}^{t} }}
               =\frac{{(w_{g}^{t}-w^{t}) h_{g}^{t} }} {{w_{g}^{t} h_{g}^{t} }}
               =1-\frac{w^{t}}{{w_g^{t}}}.\label{examp2_scale_area2}
\end{array}
\end{eqnarray}

Now center location errors $E_{C_{1}}^{t}$ and $E_{C_{2}}^{t}$ of the trackers $T_{1}^{t}$ and $T_{2}^{t}$ at $t^{th}$
frame, respectively are calculated as
\begin{eqnarray}
\begin{array}{l}
 E_{C_{1}}^{t}  = \left\| {C_{g}^{t}  - C_{1}^{t} } \right\|
                = 0, \label{examp2_scale_center1}
 \end{array}
\end{eqnarray}
since $C_{g}^{t}=C_{1}^{t}$.

\begin{eqnarray}
\begin{array}{l}
 E_{C_{2}}^{t}  = \left\| {C_{g}^{t}  - C_{2}^{t} } \right\|
  =0, \label{examp2_scale_center2}
\end{array}
\end{eqnarray}
since $C_{g}^{t}=C_{2}^{t}$.

Now if,
\begin{eqnarray}
\begin{array}{l}
1-\frac{ h^{t} } {h_g^{t}}=1-\frac{w^{t}} {w_g^{t}}
\Rightarrow \frac{h^{t}}{w^{t}} = \frac{h_g^{t}} {w_g^{t}},
 \end{array}
\end{eqnarray}
then $E_{C_{1}}^{t} = E_{C_{2}}^{t}$. Therefore, this example shows that both these algorithms Alg$1$ and Alg$2$ are equally good.

\section{Problems of Existing Evaluation Measures for Oriented Object Tracking} \label{peem_oriented_object_tracking}

\subsection{Example1:}

\begin{figure*}
\centerline{
\psfig{figure=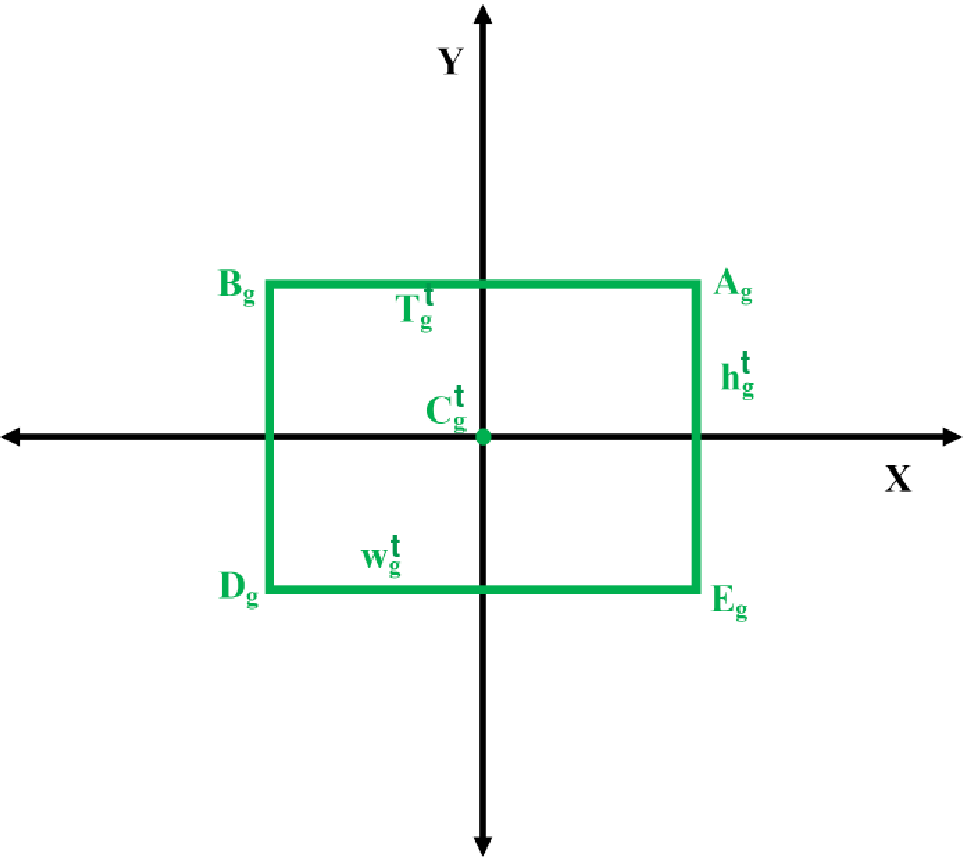, height=0.4\textwidth, width=0.5\textwidth}}
\centerline{a}
\vspace{0.0001\textwidth}
\centerline{
\psfig{figure=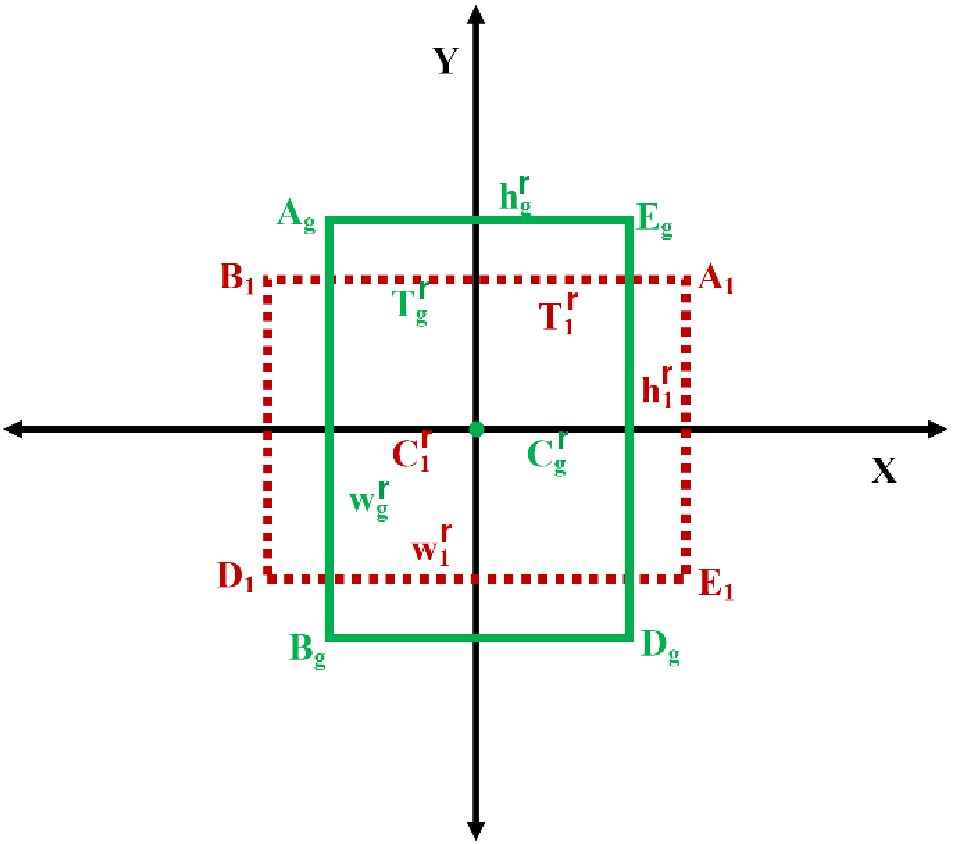,height=0.4\textwidth, width=0.5\textwidth}
\hspace{0.0001\textwidth}
\psfig{figure=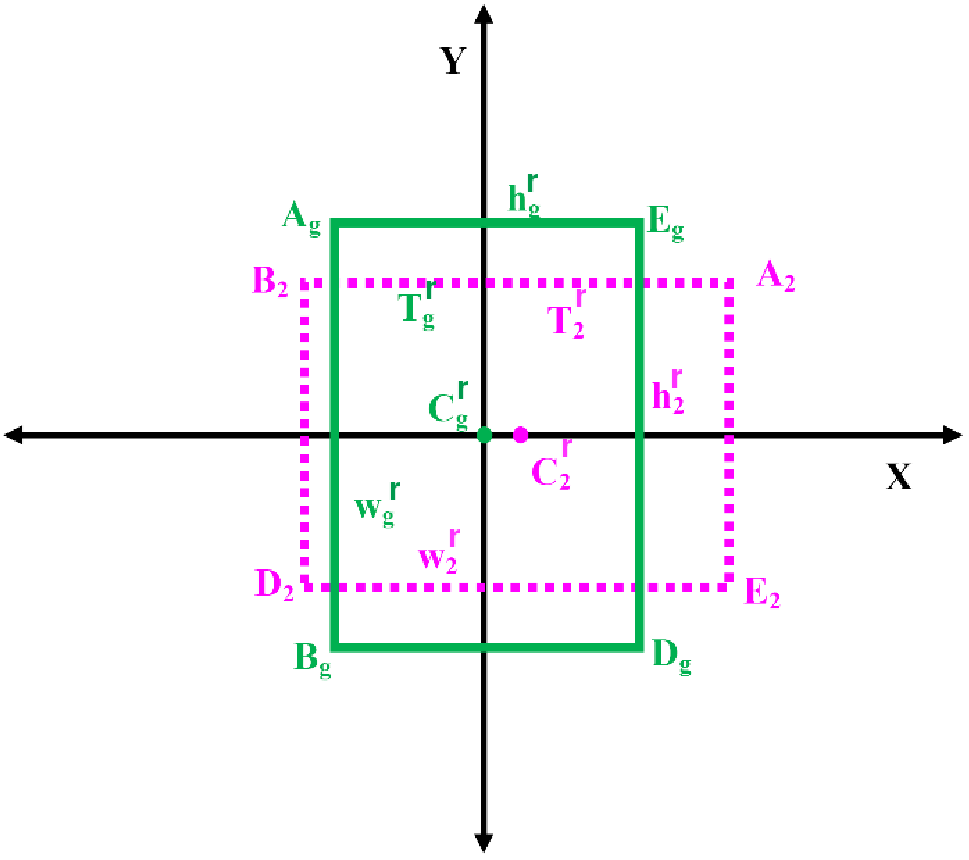,height=0.4\textwidth, width=0.5\textwidth}}
\centerline{b \hspace{0.2\textwidth} c }
\caption{\textbf{Example 1 -} area overlap between ground truth oriented tracker and fixed trackers obtained by (a) Alg$_{1}$ (b) Alg$_{2}$. \label{example1_orientation}}
\end{figure*}

Here, we consider one example where the object is changing its orientation from one frame to another, but the trackers $T_{1}$ and $T_{2}$ generated by two different tracking algorithms Alg$1$ and Alg$2$, respectively are fixed.
We assume that ground truth tracker ($T_{g}$) is not fixed i.e., it changes its orientation based on object orientation. Suppose, in $t^{th}$ frame, the ground truth tracker $T_{g}^{t}$ (green colored rectangle) with center $C_{g}^{t}$, height $h_{g}^{t}$ and width $w_{g}^{t}$, is look like as depicted in Figure~\ref{example1_orientation}(a). After few number of frames, say in $r^{th}$ ($t<r$) frame, the object changes its orientation (assume with $90^{o}$). And the corresponding ground truth tracker $T_{g}^{r}$ (green colored rectangle) with center $C_{g}^{r}$, height $h_{g}^{r}$ and width $w_{g}^{r}$ in the $r^{th}$ frame can be look like as in Figures~\ref{example1_orientation}(b) and (c). Two different tracking algorithms (Alg$1$ and Alg$2$) generate two fixed sized trackers $T_{1}^{r}$ (dotted red colored rectangle) with center $C_{1}^{r}$, height $h_{1}^{r}$ \& width $w_{1}^{r}$ and $T_{2}^{r}$ (dotted pink colored rectangle) with center $C_{2}^{r}$, height $h_{2}^{r}$ \& width $w_{2}^{r}$ in the $r^{th}$ frame are displayed in Figures~\ref{example1_orientation}(b) and (c), respectively. We also assume that $C_{g}^{r}=C_{1}^{r}$, $C_{g}^{r}<C_{2}^{r}$, $h_{1}^{r}=h_{2}^{r}=w_{g}^{2}$, $w_{1}^{r}=w_{2}^{r}=h_{g}^{r}$ and $C_{2}^{r}=C_{g}^{r}+\alpha^{r}$. We also assume that $w_{1}^{r}=w_{g}^{r}+w^{r}$, $h_{g}^{r}=h_{1}^{g}+h^{r}$, $w_{2}^{r}=w_{g}^{r}+w^{r}$ and $h_{g}^{r}=h_{2}^{g}+h^{r}$.

Now area overlap ($E_{A_{1}}^{r}$) between ground truth tracker $T_{g}^{r}$ and tracker $T_{1}^{r}$ generated by Alg$1$ at the $r^{th}$ frame, is calculated as
\begin{eqnarray}
\begin{array}{l}
E_{A_{1}}^{r}  = \frac{{T_g^{r}  \cap T_1^{r} }}{{T_g^{r}  \cup T_1^{r} }}\\
               =\frac{{h_{1}^{r} w_{g}^{r} }} {{ h_{1}^{r} w_{g}^{r} + h_{1}^{r} w^{r} + h^{r} w_{g}^{r} }}. \label{examp1_orientation_area1}
\end{array}
\end{eqnarray}

Again area overlap ($E_{A_{2}}^{r}$) between the ground truth tracker $T_{g}^{r}$ and tracker $T_{2}^{r}$ generated by Alg$2$ at the $r^{th}$ frame, is calculated as
\begin{eqnarray}
\begin{array}{l}
E_{A_{2}}^{t}  = \frac{{T_g^{r}  \cap T_2^{r} }}{{T_g^{r}  \cup T_2^{r} }} \\
               =\frac{{h_{2}^{r} w_{g}^{r} }} {{ h_{2}^{r} w_{g}^{r} + h_{2}^{r} w^{r} + h^{r} w_{g}^{r} }}\\
               =\frac{{h_{1}^{r} w_{g}^{r} }} {{ h_{1}^{r} w_{g}^{r} + h_{1}^{r} w^{r} + h^{r} w_{g}^{r} }}, \label{examp1_orientation_area2}
\end{array}
\end{eqnarray}
since $h_{1}^{r}=h_{2}^{r}$ and $w_{1}^{r}=w_{2}^{r}$.

Now center location errors $E_{C_{1}}^{r}$ and $E_{C_{2}}^{r}$ of the trackers $T_{1}^{r}$ and $T_{2}^{r}$ at the $r^{th}$ frame, respectively are calculated as
\begin{eqnarray}
\begin{array}{l}
 E_{C_{1}}^{r}  = \left\| {C_{g}^{r}  - C_{1}^{r} } \right\| \\
                = 0, \label{examp1_orientation_center1}
 \end{array}
\end{eqnarray}
since $C_{g}^{r}=C_{1}^{r}$.

\begin{eqnarray}
\begin{array}{l}
 E_{C_{2}}^{r}  = \left\| {C_{g}^{r}  - C_{2}^{r} } \right\| \\
                =\left\| {C_{g}^{r}  - (C_{g}^{r}+\alpha^{r}) } \right\| \\
                =\alpha^{r}, \label{examp1_orientation_center2}
\end{array}
\end{eqnarray}
since $C_{2}^{r}=C_{g}^{r}+\alpha^{r}$.

Eqs. (\ref{examp1_orientation_area1}) and (\ref{examp1_orientation_area2}) conclude that $E_{A_{2}}^{r}=E_{A_{1}}^{r}$, i.e., both trackers $T_{2}^{r}$ and $T_{1}^{r}$ at $t^{th}$ frame are equally good. On the other hand, Eqs. (\ref{examp1_orientation_center1}) and (\ref{examp1_orientation_center2}) conclude that $E_{C_{1}}^{r}< E_{C_{2}}^{r}$, i.e., the tracker $T_{1}^{r}$ has lesser center location error than the tracker $T_{2}^{r}$ at the $r^{th}$ frame. If we consider center location error as a performance measure then the tracker $T_{1}^{r}$ is better than the tracker $T_{2}^{r}$ at the $r^{th}$ frame. If we consider area overlap as a performance measure then both the trackers $T_{1}^{r}$ and $T_{2}^{r}$ are equally good at the $t^{th}$ frame. If we consider both these measures no decision can be made. Therefore, minimum center location error indicates maximum area overlap for the fixed sized tracker is not true for the oriented object and performance measures individually can make ambiguous decision for the oriented object.

\subsection{Example2:}

\begin{figure*}[ht!]
\centerline{
\psfig{figure=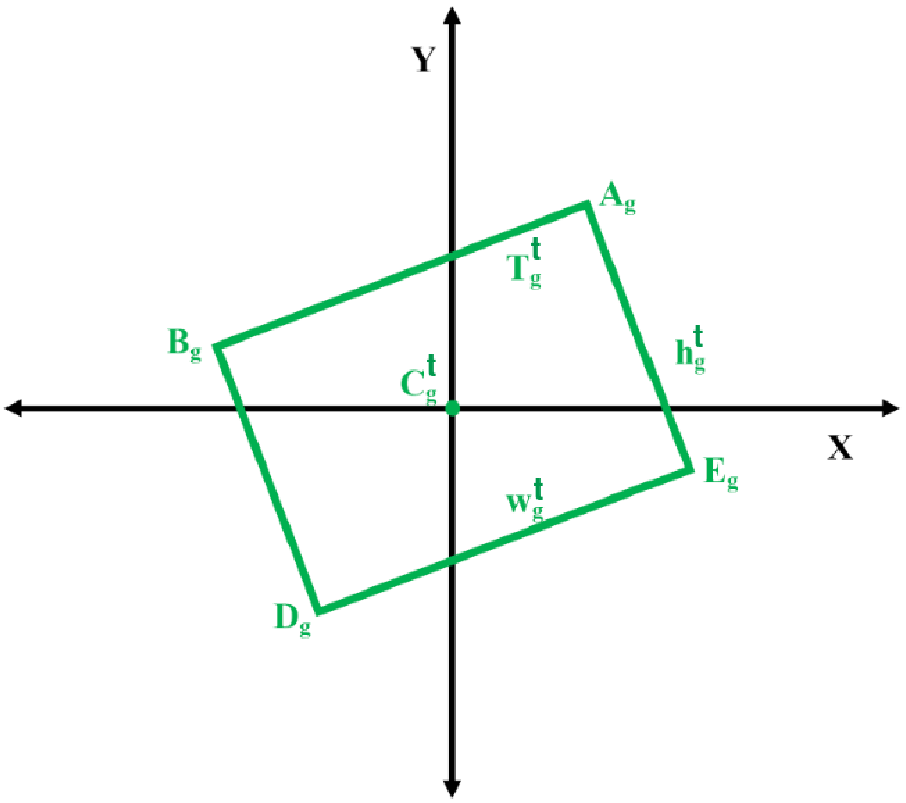,height=0.4\textwidth, width=0.5\textwidth}}
\centerline{a}
\vspace{0.0001\textwidth}
\centerline{
\psfig{figure=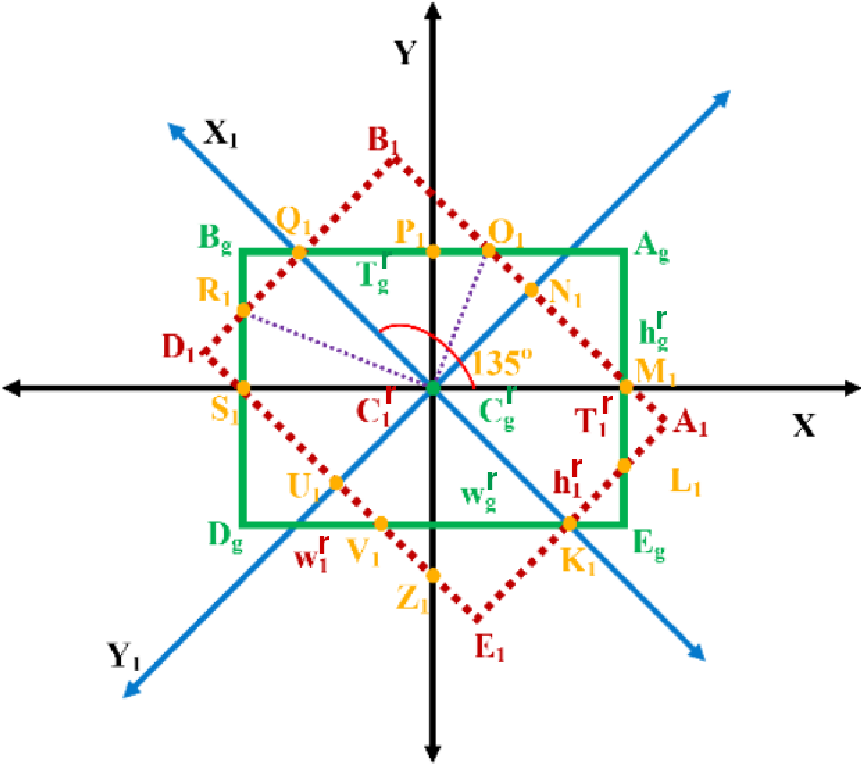,height=0.4\textwidth, width=0.5\textwidth}
\hspace{0.0001\textwidth}
\psfig{figure=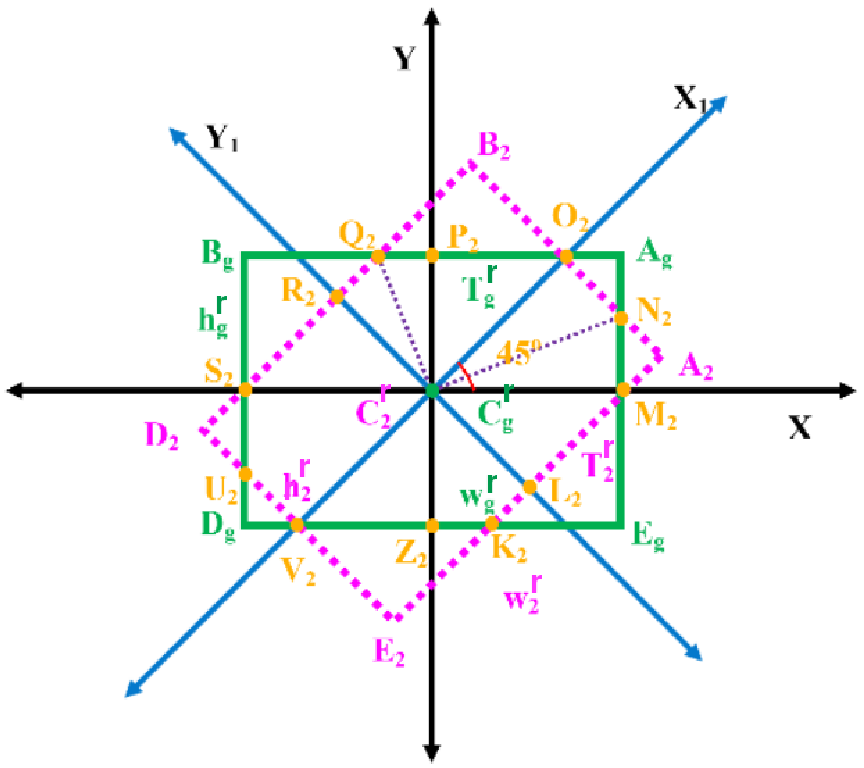,height=0.4\textwidth, width=0.5\textwidth}}
\centerline{b \hspace{0.2\textwidth} c }
\caption{\textbf{Example 2 -} area overlap between ground truth oriented tracker and oriented trackers obtained by (a) $Alg_{1}$ (b) $Alg_{2}$. \label{example2_orientation}}
\end{figure*}

Here, we consider another simple example (see Figure~\ref{example2_orientation}). Suppose $T_{g}^{t}$ (green colored rectangle) with center $C_{g}^{t}$, height $h_{g}^{t}$, width $w_{g}^{t}$ and $\theta_{r}^{t}=45^{o}$ be the ground truth tracker at the $t^{th}$ frame is displayed in Figure~\ref{example2_orientation}(a). After few number of frames,
the object changes its orientation and the corresponding ground truth tracker is also oriented based on object orientation. Let $T_{g}^{r}$ (green colored rectangle) with center $C_{g}^{r}$, height $h_{g}^{r}$, width $w_{g}^{r}$ and $\theta^{r}=0^{o}$ be ground truth tracker at the $r^{th}$ frame. Again let $T_{1}^{r}$ (red colored rectangle) with center $C_{1}^{r}$, height $h_{1}^{r}$, width $w_{1}^{r}$ \& orientation $\theta_{1}^{r}=135^{o}$ and $T_{2}^{r}$ (pink colored rectangle) with center $C_{2}^{r}$, height $h_{2}^{r}$, width $w_{2}^{r}$ \& orientation $\theta_{2}^{r}=45^{o}$ be two trackers generated by two different (oriented object) tracking algorithms Alg$1$ and Alg$2$ and are displayed in Figures~\ref{example2_orientation}(b) and (c), respectively.

Here, we also assume that the trackers $T_{g}^{r}$ and $T_{1}^{r}$ have same center location, width, height but different orientation (see Figure~\ref{example2_orientation}(b)). Therefore, we have $C_{g}^{r}=C_{1}^{r}$, $w_{g}^{r}=w_{1}^{r}$, $h_{g}^{r}=h_{1}^{r}$ and $\theta_{g}^{r}\neq \theta_{1}^{r}$. Now let $\theta_{1}^{r}=\theta_{g}^{r}+135^{o}$. We also assume that the trackers $T_{g}^{r}$ and $T_{2}^{r}$ have same center location, height, width but different orientation (see Figure~\ref{example2_orientation}(c)). Therefore, we have $C_{g}^{r}=C_{2}^{r}$, $h_{g}^{r}=h_{2}^{r}$, $w_{g}^{r}=w_{2}^{r}$ and $\theta_{g}^{r}\neq \theta_{2}^{r}$. Now let $\theta_{2}^{r}=\theta_{g}^{r}+45^{o}$.

This example concludes that two different oriented trackers have the same values for area overlap and center location error. If we consider the existing performance measures: area overlap and center location error, then both these Alg$1$ and Alg$2$ are equally good. But for oriented object tracking, both these algorithms Alg$1$ and Alg$2$ provides different interpretations.

\section{Problems of Existing Evaluation Measure for Scaled and Oriented Object Tracking} \label{peem_scaled_oriented_object_tracking}

\subsection{Example1:}

\begin{figure*}[ht!]
\centerline{
\psfig{figure=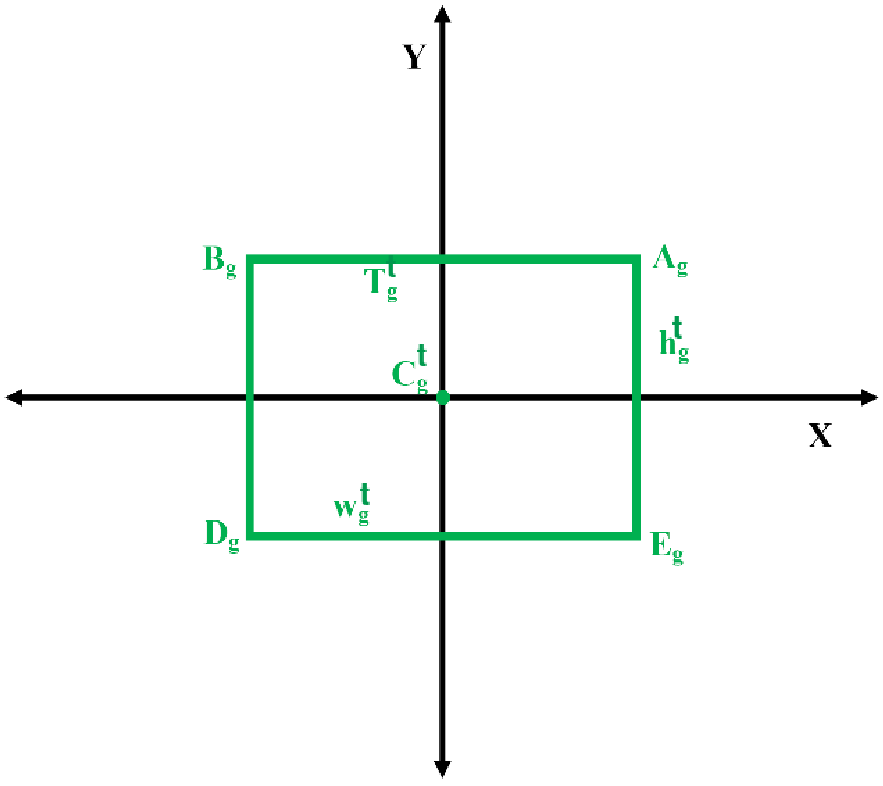,height=0.4\textwidth, width=0.5\textwidth}}
\centerline{a}
\vspace{0.0001\textwidth}
\centerline{
\psfig{figure=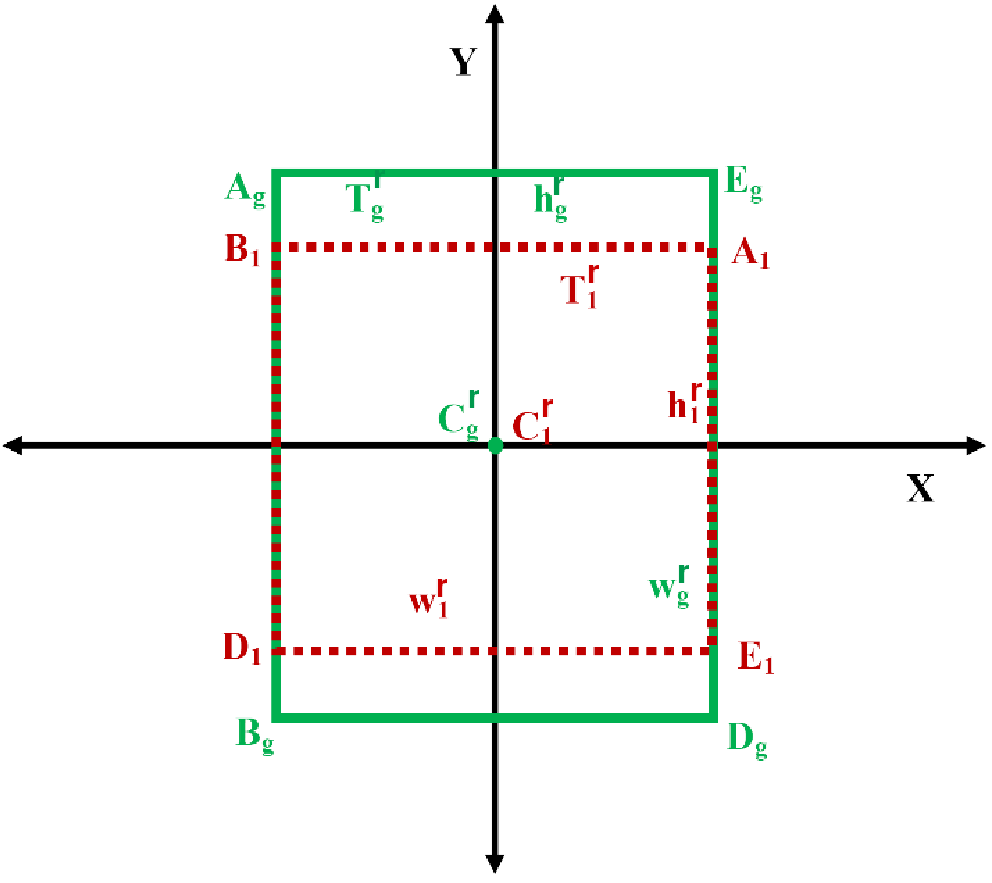,height=0.4\textwidth, width=0.5\textwidth}
\hspace{0.0001\textwidth}
\psfig{figure=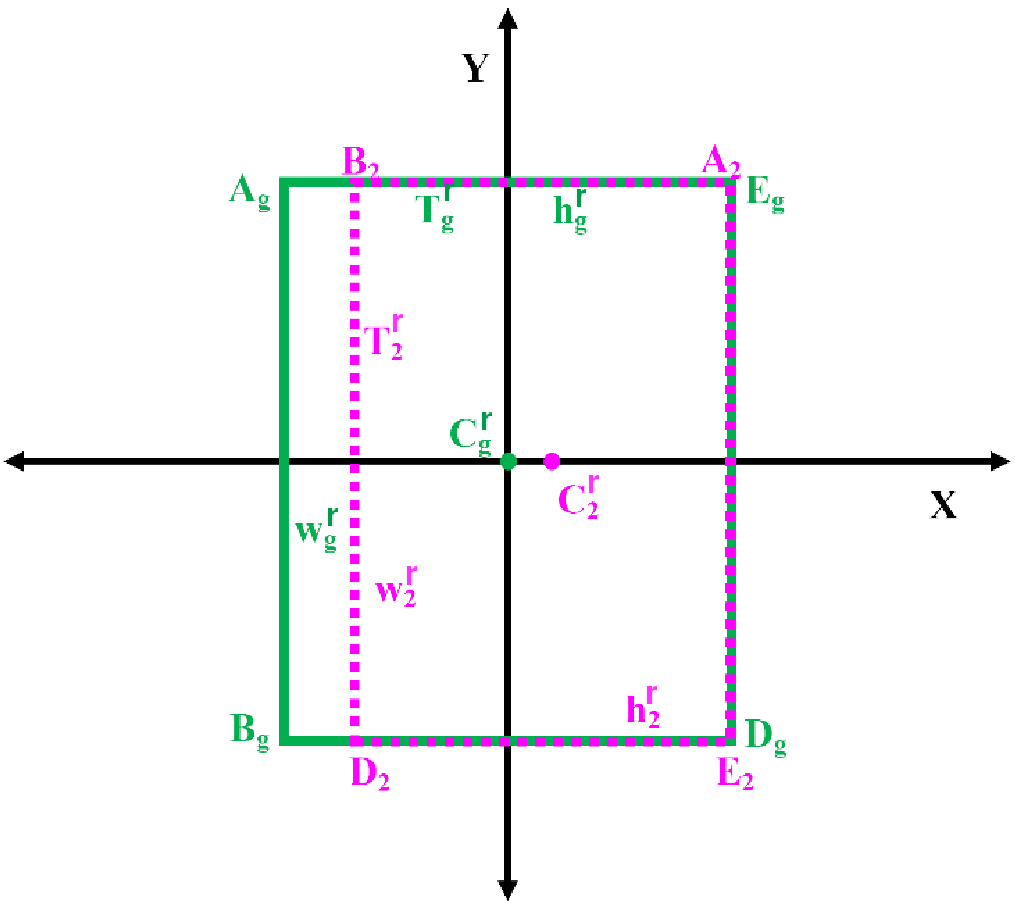,height=0.4\textwidth, width=0.5\textwidth}}
\centerline{b \hspace{0.2\textwidth} c }
\caption{\textbf{Example 1-} area overlap between ground truth scaled \& oriented tracker and scaled \& oriented trackers obtained by (a) Alg$_{1}$ and (b) Alg$_{2}$. \label{example1_scaled_orientation}}
\end{figure*}

Here, we consider one example where object is changing its scale and orientation together from one frame to another and the trackers $T_{1}$ and $T_{2}$ generated by two different tracking algorithms Alg$1$ and Alg$2$, respectively are not fixed (i.e., both are changing their scale and orientation together based on object's scale and orientation). We assume that ground truth tracker ($T_{g}$) is not fixed i.e., it changes its scale and orientation based on object's scale and orientation. We assume that at $t^{th}$ frame, the ground truth tracker (green colored rectangle) $T_{g}^{t}$ with center $C_{g}^{t}$, height $h_{g}^{t}$, width $w_{g}^{t}$ and orientation $\theta{g}^{t}=0^{o}$ is presented as in Figure~\ref{example1_scaled_orientation}(a). In the $r^{th}$ ($r>t$) frame, the object changes its scale and orientation jointly and the corresponding ground truth tracker (green colored rectangle) $T_{g}^{r}$ with center $C_{g}^{r}$, height ($h_{g}^{r}$), width ($w_{g}^{r}$) and $\theta_{g}^{r}=90^{o}$ is shown in both Figures~\ref{example1_scaled_orientation}(b) and (c). Here, it is also assume that $h_{g}^{r}>h_{g}^{t}$ and $w_{g}^{r}>w_{g}^{t}$. Therefore, we have $h_{g}^{r}=h_{g}^{t}+ h^{r}$, $w_{g}^{r}=w_{g}^{t}+w^{r}$ and $\theta_{g}^{r}=\theta_{g}^{t}+90^{o}$.

We assume that two different tracking algorithms generates two trackers $T_{1}^{r}$ (dotted red colored rectangle with center $C_{1}^{r}$, height $h_{1}^{r}$, width $w_{1}^{r}$ \& orientation $\theta_{1}^{r}=0^{o}$) and $T_{2}^{r}$ (dotted pink colored rectangle with center $C_{2}^{r}$, height $h_{2}^{r}$, width $w_{2}^{r}$ and orientation $\theta_{2}^{r}=90^{o}$) in Figure~\ref{example1_scaled_orientation}(b) and (c), respectively. Here, we also assume that $w_{1}^{r}=w_{g}^{r}$, $h_{1}^{r}=h_{g}^{r}-h^{r}$, $h_{2}^{r}=h_{g}^{r}$, $w_{2}^{r}=w_{g}^{r}-w^{r}$, $C_{1}^{r}=C_{g}^{r}$, $C_{2}^{r}=C_{g}^{r}+w^{r}$ and $h^{r}>w^{r}$.

Area overlap for both the trackers $T_{1}^{r}$ and $T_{2}^{r}$ with ground truth tracker $T_{g}^{r}$ are calculated as

\begin{eqnarray}
\begin{array}{l}
 E_{A_{1}}^{r}  = \frac{{T_{g}^{r}  \cap T_{1}^{r} }}{{T_{g}^{r}  \cup T_{1}^{r} }} \\
  = \frac{{w_{1}^{r} h_{1}^{r} }}{{w_{g}^{r} h_{g}^{r} }} = \frac{{w_{g}^{r} \left( {h_{g}^{r} - h^{r}} \right)}}{{w_{g}^{r} h_{g}^{r} }} = 1 - \frac{h^{r}}{{h_{g}^{r} }} \label{examp1_scale_orientation_area1}
 \end{array}
\end{eqnarray}
and
\begin{eqnarray}
\begin{array}{l}
 E_{A_{2}}^{r}  = \frac{{T_{g}^{r}  \cap T_{2}^{r} }}{{T_{g}^{r}  \cup T_{2}^{r} }} \\
  = \frac{{w_{2}^{r} h_{2}^{r} }}{{w_{g}^{r} h_{g}^{r} }} = \frac{{h_{g}^{r} \left( {w_{g}^{r}  - w^{r}} \right)}}{{w_{g}^{r} h_{g}^{r} }}
  = 1 - \frac{w^{r}}{{w_{g}^{r} }}.\label{examp1_scale_orientation_area2}
 \end{array}
\end{eqnarray}

Now center location errors $E_{{C}_{1}}^{r}$ and $E_{{C}_{2}}^{r}$ for both the trackers $T_{1}^{r}$ and $T_{2}^{r}$, respectively are calculated as

\begin{eqnarray}
\begin{array}{l}
E_{{C}_{1}}^{r}=\|C_{1}^{r}-C_{g}^{r} \| =0  \label{examp1_scale_orientation_center1}
\end{array}
\end{eqnarray}
and
\begin{eqnarray}
\begin{array}{l}
E_{{C}_{2}}^{r}=\|C_{2}^{r}-C_{g}^{r} \| = \|C_{g}^{r}+w^{r}-C_{g}^{r} \|=w^{r}.\label{examp1_scale_orientation_center2}
\end{array}
\end{eqnarray}

Now if $\frac{h^{r}}{h_{g}^{r}}=\frac{w^{r}}{w_{g}^{r}}$, then $E_{A_{1}}^{r}=E_{A_{2}}^{r}$. This example highlights that two different trackers have different center location error but same amount of area overlap if  $\frac{h^{r}}{h_{g}^{r}}=\frac{w^{r}}{w_{g}^{r}}$. Therefore, in this example also, the existing performance measures fail to evaluate quality of scaled and oriented object tracking algorithms. In the next section, we have developed three new auxiliary measures to quantify tracking algorithms in such complex environments.

\section{Proposed Performance Measures} \label{ppm}

Three new auxiliary evaluation measures namely error in estimated height of optimal tracker ($E_{h}^{t}$), error in estimated width of optimal tracker($E_{w}^{t}$) and error in estimated orientation angle of optimal tracker ($E_{\theta}^{t}$), respectively have been proposed to validate the tracking algorithms. For the $t^{th}$ frame, these are calculated as
\begin{eqnarray}
\begin{array}{l}
E_{h}^{t}= \|h_{g}^{t}-h^{t}\|, \\
E_{w}^{t}= \|w_{g}^{t}-w^{t}\|, \\
E_{\theta}^{t}= \|\theta_{g}^{t}-\theta^{t}\|, \label{proposed_evaluation_measure}
\end{array}
\end{eqnarray}
where $h_{g}^{t}$, $w_{g}^{t}$ and $\theta_{g}^{t}$ are the actual height, width and orientation angle, respectively, and $h^{t}$, $w^{t}$ and $\theta^{t}$ denote the estimated height, width and orientation angle for the $t^{th}$ frame, respectively. $\|.\|$ is the Euclidean norm. For a good tracking system, these three measures should be equal to zero.

Now, we define average error in estimated height of optimal tracker ($E_{h}^{avg}$), average error in estimated width of optimal tracker ($E_{w}^{avg}$) and average error in estimated orientation angle of optimal tracker ($E_{\theta}^{avg}$), to measure an overall performances of tracking algorithms.
These are calculated as
\begin{eqnarray}
\begin{array}{l}
E_{h}^{avg}=\frac{1}{N_{frames}} \sum_{t=1}^{N_{frames}} E_{h}^{t}, \\
E_{w}^{avg}=\frac{1}{N_{frames}} \sum_{t=1}^{N_{frames}} E_{w}^{t}, \\
E_{\theta}^{avg}=\frac{1}{N_{frames}} \sum_{t=1}^{N_{frames}} E_{\theta}^{t}, \label{avg_proposed_evaluation_measure}
\end{array}
\end{eqnarray}
where $N_{frames}$ is the total number of frames of a video sequence. For a good tracking system, $E_{h}^{avg}$, $E_{w}^{avg}$ and $E_{\theta}^{avg}$ should be equal to zero.

Another composite measure called as matching score ($E_{ms}^{t}$) at the $t^{th}$ frame between ground truth tracker ($T_{g}^{t}$) and tracker ($T^{t}$) generated by any tracking algorithm is proposed to evaluate performance of tracking algorithm under such complex environment. Three new proposed measures and two existing measures are combined to calculate matching score as
\begin{eqnarray}
\begin{array}{l}
E_{ms}^{t}=\frac{1}{5}\left[f_{c}^{t}+f_{h}^{t}+f_{w}^{t}+f_{\theta}^{t}+f_{A}^{t} \right],\label{ms_proposed_evaluation_measure}
\end{array}
\end{eqnarray}
where 
\begin{eqnarray}
\begin{array}{l}
f_{c}^{t} = \frac{1}{1+E_{c}^{t}}
\end{array} 
\end{eqnarray}  

\begin{eqnarray}
\begin{array}{l}
f_{h}^{t} = \frac{1}{1+E_{h}^{t}}
\end{array} 
\end{eqnarray}

\begin{eqnarray}
\begin{array}{l}
f_{w}^{t} = \frac{1}{1+E_{w}^{t}}
\end{array} 
\end{eqnarray}

\begin{eqnarray}
\begin{array}{l}
f_{\theta}^{t} = \frac{1}{1+E_{\theta}^{t}}
\end{array} 
\end{eqnarray}

\begin{eqnarray}
\begin{array}{l}
f_{A}^{t} = E_{A}^{t}
\end{array} 
\end{eqnarray}

with $E_{C}^{t}$, $E_{h}^{t}$, $E_{w}^{t}$, $E_{\theta}^{t}$ and $E_{A}^{t}$ are center location, height, width, orientation errors and area overlap at $t^{th}$ frame, respectively. 

Now we define average matching score ($E_{ms}^{avg}$) to measure overall performance of tracking algorithm, is calculated as
\begin{eqnarray}
\begin{array}{l}
E_{ms}^{avg}=\frac{1}{N_{frames}} \sum_{t=1}^{N_{frames}} E_{ms}^{t}, \label{ms_average_proposed_evaluation_measure}
\end{array}
\end{eqnarray}
where $N_{frames}$ is the total number of frames of a video sequence. For a good tracking system, $E_{ms}^{avg}$ should be equal to $1$.

\begin{table*}[ht!]
\addtolength{\tabcolsep}{-6.0pt}
\begin{center}
\begin{tabular}{|l|c|c|c|c|c|c|c|c|c|c|} \hline
\textbf{Sequence} &\multicolumn{10}{|c|}{\textbf{Area Under The Curve (AUC): Success Plot$\uparrow$}} \\ \cline{2-11}
   &\textbf{ANT} &\textbf{ASMS} &\textbf{KCF} &\textbf{LADCF} &\textbf{MEEM} &\textbf{RCO} &\textbf{SA-Siam-P} &\textbf{SA-Siam-R} &\textbf{SiamRPN} &\textbf{UPDT} \\ \hline
Bag          &0.666 &0.473 &0.565 &0.693 &0.326 &0.679 &0.682 &0.679 &\textbf{0.729} &0.674 \\
Basketball   &0.591 &0.629 &0.448 &0.686 &0.623 &0.682 &0.584 &0.633 &\textbf{0.711} &0.659 \\ 
Blanket      &0.605 &0.682 &0.637 &0.604 &0.549 &\textbf{0.693} &0.604 &0.535 &0.484 &0.558 \\
Bmx          &0.203 &0.376 &0.489 &0.308 &0.352 &\textbf{0.526} &0.422 &0.451 &0.333 &0.406 \\
Book         &0.304 &0.426 &0.339 &0.541 &0.419 &\textbf{0.602} &0.457 &0.553 &0.574 &0.509 \\
Butterfly    &0.303 &\textbf{0.601} &0.428 &0.498 &0.356 &0.568 &0.473 &0.534 &0.565 &0.526 \\
Crabs1       &0.414 &0.411 &0.298 &0.377 &0.386 &0.440 &0.369 &0.442 &\textbf{0.552} &0.485 \\
Dinosaur     &0.619 &0.399 &0.352 &0.438 &0.383 &0.452 &\textbf{0.634} &0.537 &0.588 &0.608 \\
Fernando     &0.428 &0.424 &0.388 &0.445 &0.454 &0.457 &0.482 &0.442 &\textbf{0.573} &0.449 \\
Fish2        &0.399 &0.483 &0.327 &0.396 &0.343 &0.435 &0.475 &\textbf{0.568} &0.523 &0.428 \\
Girl         &0.476 &\textbf{0.689} &0.506 &0.589 &0.468 &0.513 &0.598 &0.621 &0.626 &0.396 \\
gymnastics2  &0.377 &0.452 &0.529 &0.436 &0.426 &0.512 &0.450 &0.472 &\textbf{0.565} &0.471 \\
hand         &0.456 &0.471 &0.489 &0.406 &0.404 &0.490 &0.527 &0.481 &\textbf{0.583} &0.485 \\
helicopter   &0.631 &0.611 &0.552 &0.440 &0.420 &0.538 &0.665 &0.515 &\textbf{0.691} &0.573 \\
Iceskater2   &0.468 &0.572 &0.419 &0.543 &0.538 &0.569 &0.511 &0.576 &\textbf{0.630} &0.597 \\
Matrix       &0.259 &0.463 &0.457 &0.538 &0.490 &\textbf{0.674} &0.453 &0.575 &0.521 &0.415 \\
Motocross2   &0.345 &0.593 &0.300 &0.488 &0.209 &\textbf{0.603} &\textbf{0.603} &0.489 &0.390 &0.525 \\
Shaking      &0.238 &0.509 &0.707 &0.688 &0.553 &0.716 &0.727 &0.677 &\textbf{0.782} &0.707 \\
Soccer1      &0.367 &0.443 &0.648 &0.627 &0.518 &\textbf{0.694} &0.431 &0.533 &0.618 &0.676 \\
Traffic      &0.630 &0.458 &0.766 &0.658 &0.727 &0.739 &0.713 &0.713 &\textbf{0.767} &0.733 \\ \hline
Average over &      &      &      &      &      &      &      &      &      &      \\      
60 sequences &0.439 &0.508 &0.482 &0.520 &0.447 &0.579 &0.543 &0.551 &\textbf{0.590} &0.544\\ \hline 
\end{tabular}
\end{center}
\caption{Shows {\sc auc} of the success plots of the existing measures. The bold value indicates the best tracker. \label{table_auc_success}}
\end{table*}

\begin{table*}[ht!]
\addtolength{\tabcolsep}{-6.0pt}
\begin{center}
\begin{tabular}{|l|c|c|c|c|c|c|c|c|c|c|} \hline
\textbf{Sequence} &\multicolumn{10}{|c|}{\textbf{Area Under The Curve (AUC): Success Plot of Matching Score $\uparrow$}} \\ \cline{2-11}
   &\textbf{ANT} &\textbf{ASMS} &\textbf{KCF} &\textbf{LADCF} &\textbf{MEEM} &\textbf{RCO} &\textbf{SA-Siam-P} &\textbf{SA-Siam-R} &\textbf{SiamRPN} &\textbf{UPDT} \\ \hline
Bag          &0.273 &0.186 &0.204 &0.297 &0.142 &0.295 &0.294 &0.277 &\textbf{0.376} &0.295 \\
Basketball   &0.269 &0.315 &0.178 &0.329 &0.308 &0.329 &0.261 &0.307 &\textbf{0.374} &0.327\\
Blanket      &0.358 &\textbf{0.406} &0.328 &0.337 &0.273 &0.374 &0.309 &0.277 &0.244 &0.318 \\
Bmx          &0.071 &0.138 &\textbf{0.176} &0.100 &0.125 &0.173 &0.154 &0.151 &0.110 &0.144 \\
Book         &0.144 &0.234 &0.184 &0.296 &0.220 &0.318 &0.246 &\textbf{0.326} &0.322 &0.287 \\
Butterfly    &0.124 &\textbf{0.241} &0.172 &0.193 &0.148 &0.220 &0.188 &0.209 &\textbf{0.241} &0.209 \\
Crabs1       &0.232 &0.198 &0.169 &0.178 &0.190 &0.192 &0.159 &0.209 &\textbf{0.263} &0.209 \\
Dinosaur     &0.254 &0.173 &0.182 &0.190 &0.183 &0.217 &\textbf{0.292} &0.239 &0.261 &0.275  \\
Fernando     &0.177 &0.156 &0.149 &0.166 &0.193 &0.184 &0.171 &0.169 &\textbf{0.219} &0.162 \\
Fish2        &0.206 &0.243 &0.195 &0.236 &0.236 &0.239 &0.233 &\textbf{0.299} &\textbf{0.299} &0.210 \\
Girl         &0.211 &0.323 &0.216 &0.250 &0.210 &0.214 &0.247 &0.258 &\textbf{0.284} &0.195 \\
gymnastics2  &0.160 &0.236 &0.302 &0.222 &0.231 &\textbf{0.286} &0.244 &0.258 &0.259 &0.282 \\
hand         &0.251 &0.275 &0.276 &0.199 &0.212 &0.250 &0.288 &0.259 &\textbf{0.338} &0.279 \\
helicopter   &0.328 &0.302 &0.254 &0.222 &0.197 &0.277 &\textbf{0.389} &0.235 &0.369 &0.319 \\
Iceskater2   &0.166 &0.221 &0.176 &0.217 &0.236 &0.247 &0.211 &0.242 &\textbf{0.274} &0.260 \\
Matrix       &0.161 &0.251 &0.254 &0.285 &0.251 &\textbf{0.353} &0.224 &0.295 &0.268 &0.209 \\
Motocross2   &0.103 &0.188 &0.101 &0.146 &0.091 &0.189 &\textbf{0.203} &0.157 &0.116 &0.163 \\
Shaking      &0.096 &0.301 &0.374 &0.357 &0.220 &0.321 &0.364 &0.293 &\textbf{0.417} &0.360 \\
Soccer1      &0.186 &0.229 &0.349 &0.308 &0.236 &\textbf{0.369} &0.209 &0.265 &0.314 &0.340 \\
Traffic      &0.392 &0.296 &0.508 &0.389 &0.434 &0.471 &0.469 &0.467 &\textbf{0.519} &0.504 \\ \hline
Average over &      &      &      &      &      &      &      &      &      &      \\      
60 sequences &0.208 &0.246 &0.237 &0.246 &0.217 &0.276 &0.258 &0.259 &\textbf{0.294} &0.268\\ \hline 
\end{tabular}
\end{center}
\caption{Shows {\sc auc} of the success plots of new measure - matching score. The bold value indicates the best tracker. \label{table_auc_new_success}}
\end{table*}

To measure the performance of an algorithm on a sequence of frames, we count the number of successful frames whose matching score ($E_{ms}^{t}$) is larger than a given threshold value. The matching score plot shows the ratios of successful frames at the thresholds varied from $0$ to $1$. The matching rate value at a specific threshold (e.g., $0.5$) may not always be a good representative for evaluating the trackers' performance. Instead of that, the area under the curve ({\sc auc}) of each matching score plot is considered to evaluate the performance of the tracking algorithms.

\section{Experiments} \label{result_analysis}

\subsection{Dataset} \label{dataset}

We use entire {\sc vot-2018}\footnote{\tiny{\url{https://www.votchallenge.net/vot2018/dataset.html}}}~\cite{kristan2018sixth} dataset for our experiments. {\sc vot-2018} dataset contains 60 video sequences. The target in the sequences is annotated by a rotated bounding box and all sequences are per-frame annotated by the following visual attributes: (i) occlusion, (ii) illumination change, (iii) motion change, (iv) size change, and (v) camera motion. Frames that do not correspond to any of the five attributes are denoted as (vi) unassigned. All these complex factors make object tracking difficult on this dataset. If a tracker fails to track object in a frame, we reset the tracker to 0 (as oppose to {\sc vot} protocol). We only consider one-pass evaluation ({\sc ope}) to draw the plots to compare performances of all the trackers with respect to both existing and the proposed measures.

\subsection{Baselines} \label{baselines}

We use various state-of-the-art trackers on {\sc vot-2018} dataset such as 
{\sc ant}$^{1}$ [9], {\sc kcf}\footnote{\tiny{\url{https://github.com/vojirt/kcf}}}~\cite{henriques2014high},
{\sc asms}\footnote{\tiny{\url{https://github.com/vojirt/asms}}}~\cite{vojir2014robust}, 
{\sc updt}~\cite{bhat2018unveiling}, 
{\sc ladcf}\footnote{\tiny{\url{https://github.com/XU-TIANYANG/LADCF.git}}}~\cite{xu2019learning},
{\sc meem}\footnote{\tiny{\url{http://www.cs.bu.edu/groups/ivc/software/MEEM/}}}~\cite{zhang2014meem}, 
{\sc rco}$^{1}$ [9], {\sc sa-s}iam-{\sc r}$^{1}$ [9], {\sc sa-s}iam-{\sc p}$^{1}$ [9], and {\sc s}iam{\sc rpn}~\cite{li2018high} for our experiments. Among them, {\sc sa-s}iam-{\sc r}, {\sc sa-s}iam-{\sc p}, and {\sc s}iam{\sc rpn} apply {\sc s}iamese networks, {\sc ladcf}, {\sc rco}, {\sc kcf}, and {\sc updt} are developed based on discriminative correlation filters, {\sc asms} applies mean shift, {\sc ant} uses optical flow, and {\sc meem} is based on support vector machine.

\subsection{Quantitative Results Analysis} \label{quantitative_result_analysis}

\begin{table*}[ht!]
\addtolength{\tabcolsep}{-6.0pt}
\begin{center}
\begin{tabular}{|l|c|c|c|c|c|c|c|c|c|c|} \hline
\textbf{Sequence} &\multicolumn{10}{|c|}{\textbf{Precision Value at Threshold 20 Pixels $\uparrow$}} \\ \cline{2-11}
   &\textbf{ANT} &\textbf{ASMS} &\textbf{KCF} &\textbf{LADCF} &\textbf{MEEM} &\textbf{RCO} &\textbf{SA-Siam-P} &\textbf{SA-Siam-R} &\textbf{SiamRPN} &\textbf{UPDT} \\ \hline
Bag          &0.607 &0.474 &0.526 &0.995 &0.036 &0.959 &\textbf{0.995} &\textbf{0.995} &\textbf{0.995} &\textbf{0.995} \\
Basketball   &0.934 &0.941 &0.538 &0.989 &0.804 &\textbf{0.992} &0.895 &0.989 &0.934 &0.990 \\
Blanket      &0.933 &0.978 &0.987 &\textbf{0.996} &0.987 &0.987 &0.991 &0.987 &0.973 &0.889 \\
Bmx          &0.263 &0.237 &\textbf{0.316} &0.237 &0.263 &0.263 &0.145 &0.211 &0.197 &0.158 \\
Book         &0.360 &0.600 &0.383 &\textbf{0.909} &0.611 &\textbf{0.909} &0.646 &0.783 &0.817 &0.806 \\
Butterfly    &0.291 &0.907 &0.318 &0.523 &0.325 &0.649 &0.358 &0.589 &\textbf{0.709} &0.669 \\ 
Crabs1       &0.381 &0.713 &0.250 &0.363 &0.350 &0.456 &0.588 &0.594 &0.719 &\textbf{0.806} \\
Dinosaur     &0.715 &0.248 &0.267 &0.684 &0.340 &0.494 &0.809 &0.638 &0.813 &\textbf{0.985}\\
Fernando     &0.507 &0.223 &0.562 &0.627 &0.647 &\textbf{0.678} &0.619 &0.363 &0.634 &0.606 \\
Fish2        &0.571 &0.845 &0.258 &0.761 &0.371 &0.803 &0.800 &\textbf{0.874} &0.819 &0.855 \\
Girl         &0.916 &\textbf{0.921} &0.313 &0.729 &0.849 &0.378 &0.281 &0.335 &0.595 &0.176 \\
gymnastics2  &0.329 &0.683 &\textbf{0.736} &0.688 &0.729 &0.738 &0.692 &0.696 &0.671 &0.725 \\
hand         &\textbf{0.978} &0.719 &0.697 &0.873 &0.655 &0.742 &0.828 &0.757 &0.865 &0.933 \\
helicopter   &0.637 &0.863 &0.595 &0.472 &0.301 &0.623 &0.643 &0.836 &\textbf{0.879} &0.675 \\
Iceskater2   &0.402 &0.454 &0.266 &0.521 &0.646 &\textbf{0.897} &0.636 &0.761 &0.714 &0.829 \\
Matrix       &0.480 &0.770 &0.650 &0.850 &0.850 &\textbf{0.990} &0.730 &0.830 &0.760 &0.910 \\
Motocross2   &0.098 &0.148 &0.00 &0.066 &0.033 &0.098 &0.098 &0.098 &\textbf{0.213} &0.197 \\
Shaking      &0.660 &0.488 &0.858 &0.939 &0.923 &\textbf{0.997} &0.995 &\textbf{0.997} &\textbf{0.997} &0.995 \\
Soccer1      &0.513 &0.515 &0.865 &0.918 &0.837 &0.954 &0.546 &0.819 &0.859 &\textbf{0.959} \\
Traffic      &0.953 &0.613 &0.993 &\textbf{0.995} &\textbf{0.995} &\textbf{0.995} &\textbf{0.995} &\textbf{0.995} &\textbf{0.995} &\textbf{0.995} \\ \hline
Average over &      &      &      &      &      &      &      &      &      &      \\      
60 sequences &0.576 &0.617 &0.519 &0.707 &0.578 &0.730 &0.664 &0.707 &\textbf{0.758} &\textbf{0.758} \\ \hline 
\end{tabular}
\end{center}
\caption{Shows precision value at threshold (20 pixels). The bold value indicates the best tracker. \label{table_precision_value}}
\end{table*}

\begin{table*}[ht!]
\addtolength{\tabcolsep}{-6.0pt}
\begin{center}
\begin{tabular}{|l|c|c|c|c|c|c|c|c|c|c|} \hline
\textbf{Sequence} &\multicolumn{10}{|c|}{\textbf{Success Rate at Threshold 0.5 $\uparrow$}} \\ \cline{2-11}
   &\textbf{ANT} &\textbf{ASMS} &\textbf{KCF} &\textbf{LADCF} &\textbf{MEEM} &\textbf{RCO} &\textbf{SA-Siam-P} &\textbf{SA-Siam-R} &\textbf{SiamRPN} &\textbf{UPDT} \\ \hline
Bag          &0.888 &0.444 &0.796 &0.949 &0.077 &0.888 &0.868 &0.893 &\textbf{0.995} &0.949 \\
Basketball   &0.805 &0.849 &0.366 &0.934 &0.906 &0.966 &0.775 &0.847 &\textbf{0.942} &0.899 \\
Blanket      &0.644 &0.867 &0.920 &0.787 &0.760 &\textbf{0.942} &0.827 &0.667 &0.418 &0.778 \\
Bmx          &0.118 &0.224 &0.487 &0.197 &0.263 &\textbf{0.579} &0.355 &0.316 &0.197 &0.342 \\
Book         &0.251 &0.400 &0.320 &0.697 &0.463 &\textbf{0.789} &0.568 &0.703 &0.731 &0.669 \\
Butterfly    &0.159 &\textbf{0.795} &0.258 &0.490 &0.166 &0.755 &0.424 &0.702 &0.609 &0.616 \\ 
Crabs1       &0.325 &0.313 &0.150 &0.200 &0.286 &0.363 &0.150 &0.344 &\textbf{0.706} &0.544 \\
Dinosaur     &\textbf{0.899} &0.328 &0.288 &0.420 &0.316 &0.515 &0.776 &0.571 &0.782 &0.822\\
Fernando     &0.318 &0.277 &0.329 &0.449 &0.445 &0.455 &0.466 &0.390 &\textbf{0.688} &0.404 \\
Fish2        &0.419 &0.555 &0.216 &0.239 &0.323 &0.281 &0.481 &\textbf{0.665} &0.623 &0.365 \\
Girl         &0.432 &0.895 &0.510 &0.844 &0.382 &0.560 &0.867 &\textbf{0.911} &0.811 &0.153 \\
gymnastics2  &0.336 &0.438 &0.596 &0.421 &0.404 &0.471 &0.429 &0.429 &\textbf{0.629} &0.433 \\
hand         &0.367 &0.494 &0.543 &0.255 &0.337 &0.566 &0.581 &0.487 &\textbf{0.764} &0.487 \\
helicopter   &0.794 &0.794 &0.524 &0.329 &0.463 &0.523 &0.773 &0.486 &\textbf{0.870} &0.605 \\
Iceskater2   &0.414 &0.743 &0.393 &0.648 &0.648 &0.685 &0.569 &0.704 &\textbf{0.785} &0.765 \\
Matrix       &0.150 &0.520 &0.580 &0.620 &0.560 &\textbf{0.880} &0.500 &0.740 &0.580 &0.240 \\
Motocross2   &0.098 &0.836 &0.246 &0.459 &0.0819 &0.853 &\textbf{0.869} &0.574 &0.164 &0.623 \\
Shaking      &0.022 &0.551 &0.858 &\textbf{0.997} &0.699 &\textbf{0.997} &0.984 &0.981 &0.975 &0.981 \\
Soccer1      &0.253 &0.457 &0.801 &0.816 &0.625 &\textbf{0.883} &0.431 &0.561 &0.793 &0.875 \\
Traffic      &0.895 &0.576 &0.974 &0.921 &\textbf{0.989} &0.974 &0.958 &0.927 &0.969 &0.974 \\ \hline
Average over &      &      &      &      &      &      &      &      &      &      \\      
60 sequences &0.429 &0.568 &0.508 &0.584 &0.459 &0.696 &0.632 &0.645 &\textbf{0.702} &0.626 \\ \hline 
\end{tabular}
\end{center}
\caption{Shows success rate at threshold (0.5) in the success plot of the existing measure. The bold value indicates the best tracker. \label{table_success_value}}
\end{table*}

\begin{table*}[ht!]
\addtolength{\tabcolsep}{-6.0pt}
\begin{center}
\begin{tabular}{|l|c|c|c|c|c|c|c|c|c|c|} \hline
\textbf{Sequence} &\multicolumn{10}{|c|}{\textbf{New Success Rate at Threshold 0.5 $\uparrow$}} \\ \cline{2-11}
   &\textbf{ANT} &\textbf{ASMS} &\textbf{KCF} &\textbf{LADCF} &\textbf{MEEM} &\textbf{RCO} &\textbf{SA-Siam-P} &\textbf{SA-Siam-R} &\textbf{SiamRPN} &\textbf{UPDT} \\ \hline
Bag          &0.066 &0.015 &0.005 &0.077 &0.005 &0.102 &0.097 &0.03 &\textbf{0.229} &0.066 \\
Basketball   &0.052 &0.095 &0.023 &0.123 &0.066 &0.126 &0.047 &0.081 &\textbf{0.215} &0.105 \\
Blanket      &0.280 &\textbf{0.284} &0.080 &0.142 &0.031 &0.156 &0.093 &0.084 &0.084 &0.129 \\
Bmx          &0.000 &0.026 &0.013 &0.000 &0.013 &0.000 &\textbf{0.039} &0.026 &0.000 &0.000 \\
Book         &0.017 &0.057 &0.034 &0.128 &0.069 &0.137 &0.103 &\textbf{0.240} &0.189 &0.120 \\
Butterfly    &0.007 &0.013 &0.007 &0.007 &0.000 &0.019 &0.000 &0.007 &\textbf{0.053} &0.007 \\ 
Crabs1       &0.069 &0.019 &0.031 &0.013 &0.025 &0.000 &0.006 &0.075 &\textbf{0.081} &0.025\\
Dinosaur     &0.043 &0.034 &0.065 &0.015 &0.046 &0.028 &\textbf{0.095} &0.064 &0.046 &0.039\\
Fernando     &\textbf{0.051} &0.000 &0.003 &0.000 &0.014 &0.000 &0.000 &0.010 &0.024 &0.000 \\
Fish2        &0.081 &0.048 &0.100 &0.100 &0.145 &0.061 &0.065 &0.139 &\textbf{0.171} &0.029 \\
Girl         &0.028 &\textbf{0.125} &0.035 &0.027 &0.053 &0.013 &0.033 &0.039 &0.097 &0.021 \\
gymnastics2  &0.000 &0.113 &0.213 &0.104 &0.121 &0.158 &0.138 &0.150 &0.100 &\textbf{0.171} \\
hand         &0.059 &0.105 &0.154 &0.034 &0.059 &0.086 &0.112 &0.112 &\textbf{0.184} &0.075 \\
helicopter   &0.189 &0.138 &0.095 &0.041 &0.102 &0.079 &0.312 &0.105 &\textbf{0.259} &0.162 \\
Iceskater2   &0.002 &0.013 &0.007 &0.013 &0.044 &0.044 &0.035 &0.029 &\textbf{0.076} &0.037\\
Matrix       &0.020 &0.120 &0.150 &0.160 &0.110 &\textbf{0.190} &0.020 &0.160 &0.120 &0.040 \\
Motocross2   &0.000 &0.000 &\textbf{0.016} &0.000 &\textbf{0.016} &0.000 &0.000 &0.000 &0.000 &0.000 \\
Shaking      &0.000 &0.148 &0.230 &0.167 &0.025 &0.093 &0.184 &0.060 &\textbf{0.326} &0.205 \\
Soccer1      &0.038 &0.089 &0.176 &0.117 &0.051 &\textbf{0.229} &0.048 &0.079 &0.133 &0.153 \\
Traffic      &0.230 &0.126 &0.439 &0.209 &0.304 &0.408 &0.387 &0.408 &\textbf{0.492} &0.419\\ \hline
Average over &      &      &      &      &      &      &      &      &      &      \\      
60 sequences &0.062 &0.078 &0.094 &0.074 &0.065 &0.097 &0.091 &0.095 &\textbf{0.144} &0.090 \\ \hline 
\end{tabular}
\end{center}
\caption{Shows the success rate at threshold (0.5) in the success plots of the proposed measure --- matching score. The bold value indicates the best tracker. \label{table_new_success_value}}
\end{table*}

Tables~\ref{table_auc_success} and~\ref{table_auc_new_success} show {\sc auc} of success plots of the existing measure and the proposed measure, respectively. The bold value indicates the best tracker among the considered state-of-the-art trackers. Tables~\ref{table_precision_value},~\ref{table_success_value}, and~\ref{table_new_success_value} present precision value at threshold 20 pixels, success rate at threshold 0.5 , and new success rate at threshold 0.5. Bold values in the tables indicate the best trackers. According to {\sc auc} of success plots of existing measures, SiamRPM performs the best on 10 sequences among randomly selected 20 sequences. In case of average of 60 sequences, SiamRPM also performs the best. SiamRPM obtains the best results for 11 sequences with respect to {\sc auc} of success plots of the proposed measure --- matching score. SiamRPM also performs the best with respect to {\sc auc} in the average success plot of 60 sequences.

\subsection{Qualitative Results Analysis} \label{qualitative_result_analysis}

\begin{figure*}[htp!]
\centerline{ 
\tcbox[sharp corners, size = tight, boxrule=0.2mm, colframe=black, colback=white]{
\psfig{figure=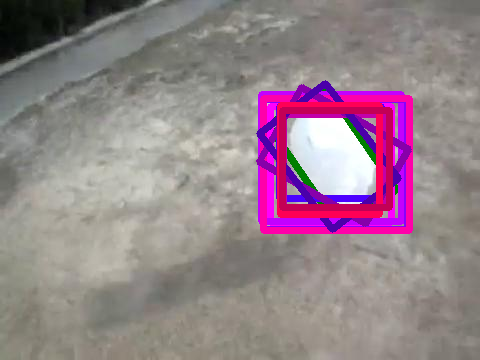, width=0.23\textwidth,height=0.2\textwidth}}
\hspace{0.001\textwidth}
\tcbox[sharp corners, size = tight, boxrule=0.2mm, colframe=black, colback=white]{
\psfig{figure=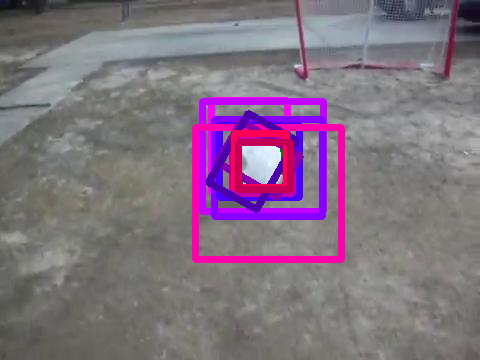, width=0.23\textwidth,height=0.2\textwidth}}
\hspace{0.001\textwidth}
\tcbox[sharp corners, size = tight, boxrule=0.2mm, colframe=black, colback=white]{
\psfig{figure=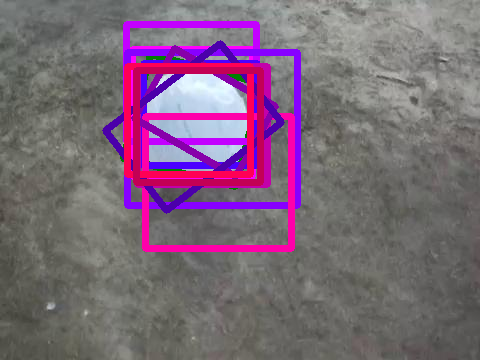, width=0.23\textwidth,height=0.2\textwidth}}
\hspace{0.001\textwidth}
\tcbox[sharp corners, size = tight, boxrule=0.2mm, colframe=black, colback=white]{
\psfig{figure=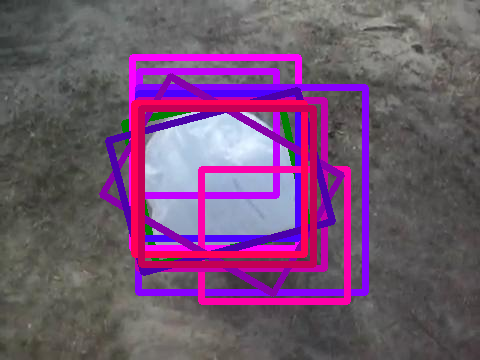, width=0.23\textwidth,height=0.2\textwidth}}}
\vspace{-0.03\textwidth}
\centerline{\hspace{0.03\textwidth} Bag sequence}
\vspace{0.001\textwidth}
\centerline{ 
\tcbox[sharp corners, size = tight, boxrule=0.2mm, colframe=black, colback=white]{
\psfig{figure=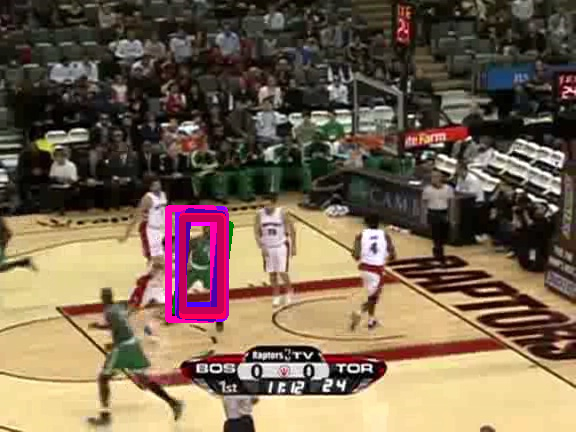, width=0.23\textwidth,height=0.2\textwidth}}
\hspace{0.001\textwidth}
\tcbox[sharp corners, size = tight, boxrule=0.2mm, colframe=black, colback=white]{
\psfig{figure=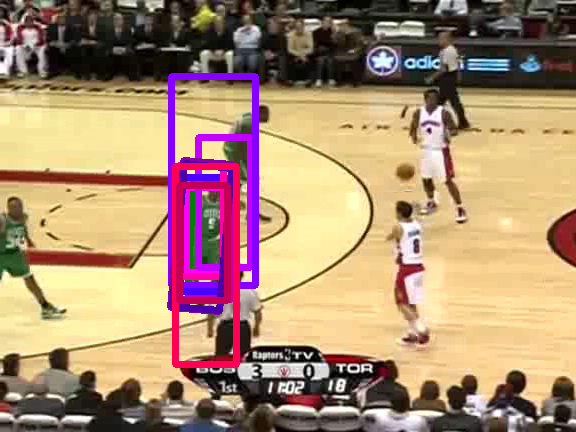, width=0.23\textwidth,height=0.2\textwidth}}
\hspace{0.001\textwidth}
\tcbox[sharp corners, size = tight, boxrule=0.2mm, colframe=black, colback=white]{
\psfig{figure=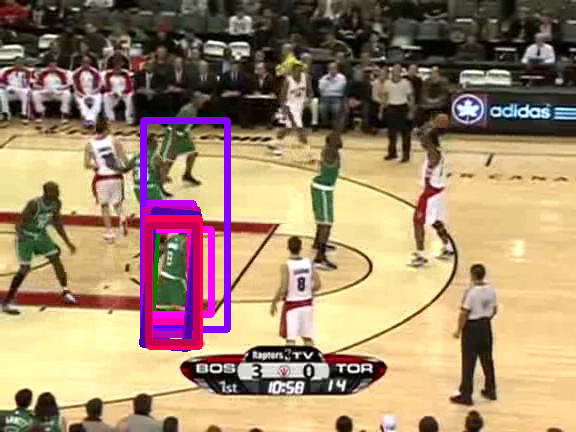, width=0.23\textwidth,height=0.2\textwidth}}
\hspace{0.001\textwidth}
\tcbox[sharp corners, size = tight, boxrule=0.2mm, colframe=black, colback=white]{
\psfig{figure=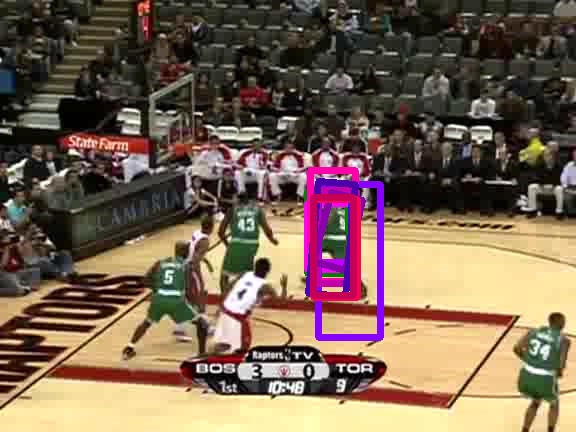, width=0.23\textwidth,height=0.2\textwidth}}}
\vspace{-0.03\textwidth}
\centerline{\hspace{0.03\textwidth} Basketball sequence}
\vspace{0.001\textwidth}
\centerline{ 
\tcbox[sharp corners, size = tight, boxrule=0.2mm, colframe=black, colback=white]{
\psfig{figure=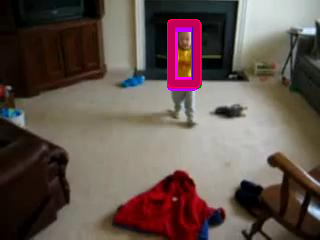, width=0.23\textwidth,height=0.2\textwidth}}
\hspace{0.001\textwidth}
\tcbox[sharp corners, size = tight, boxrule=0.2mm, colframe=black, colback=white]{
\psfig{figure=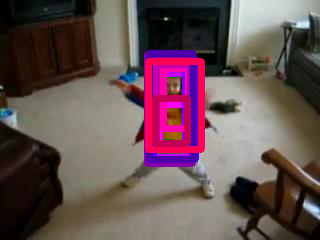, width=0.23\textwidth,height=0.2\textwidth}}
\hspace{0.001\textwidth}
\tcbox[sharp corners, size = tight, boxrule=0.2mm, colframe=black, colback=white]{
\psfig{figure=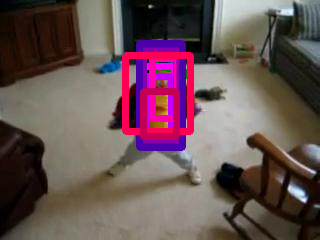, width=0.23\textwidth,height=0.2\textwidth}}
\hspace{0.001\textwidth}
\tcbox[sharp corners, size = tight, boxrule=0.2mm, colframe=black, colback=white]{
\psfig{figure=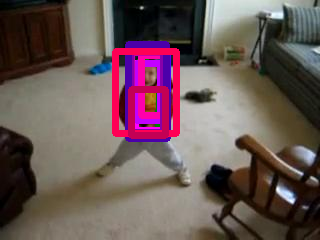, width=0.23\textwidth,height=0.2\textwidth}}}
\vspace{-0.03\textwidth}
\centerline{\hspace{0.03\textwidth} Blanket sequence}
\vspace{0.001\textwidth}
\centerline{ 
\tcbox[sharp corners, size = tight, boxrule=0.2mm, colframe=black, colback=white]{
\psfig{figure=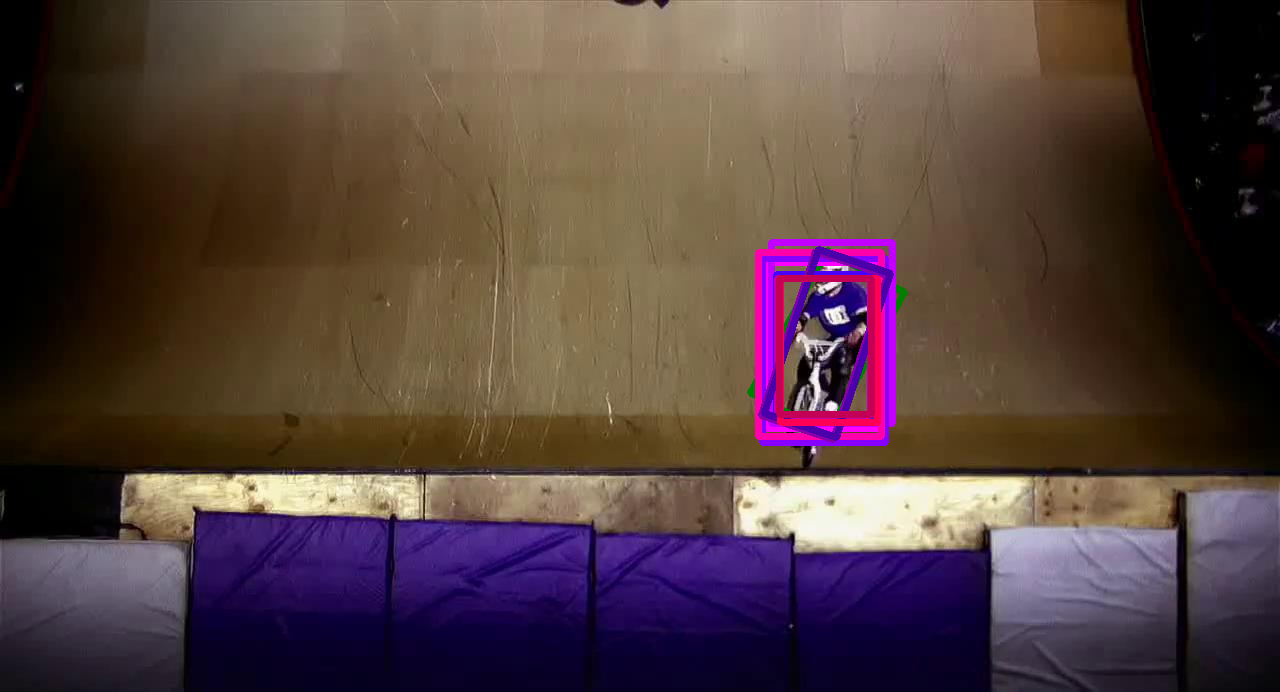, width=0.23\textwidth,height=0.2\textwidth}}
\hspace{0.001\textwidth}
\tcbox[sharp corners, size = tight, boxrule=0.2mm, colframe=black, colback=white]{
\psfig{figure=result_images/bmx_00000003.png, width=0.23\textwidth,height=0.2\textwidth}}
\hspace{0.001\textwidth}
\tcbox[sharp corners, size = tight, boxrule=0.2mm, colframe=black, colback=white]{
\psfig{figure=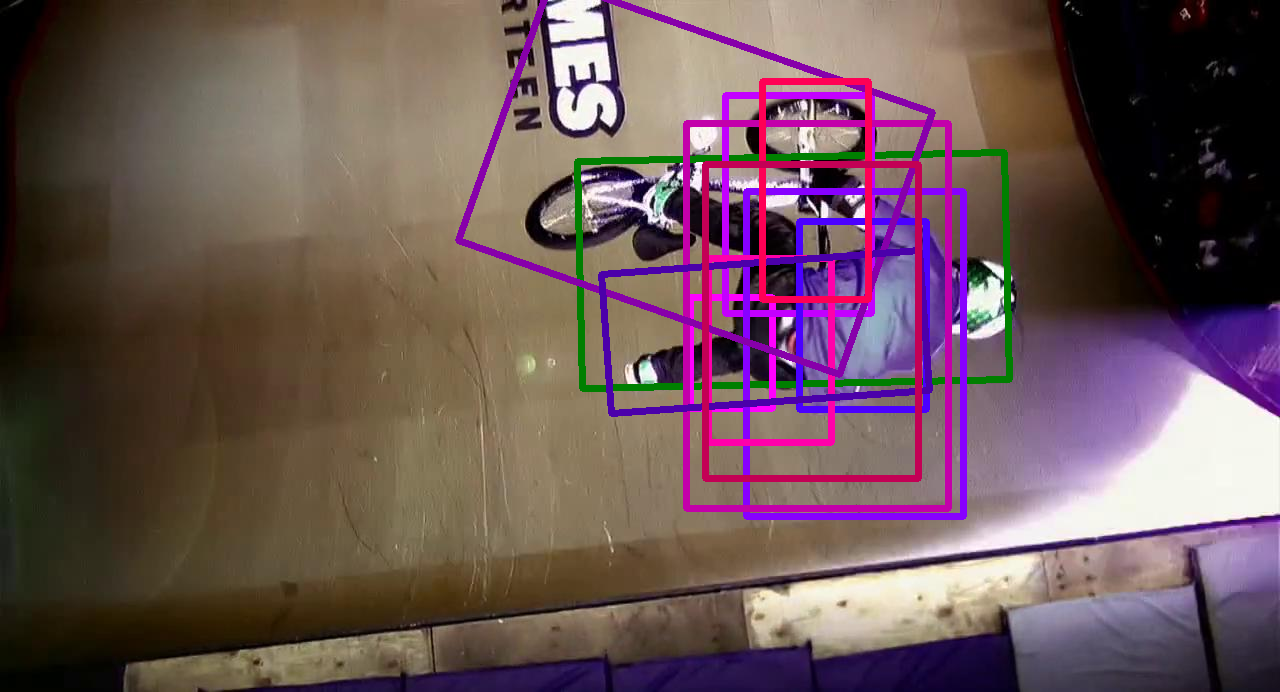, width=0.23\textwidth,height=0.2\textwidth}}
\hspace{0.001\textwidth}
\tcbox[sharp corners, size = tight, boxrule=0.2mm, colframe=black, colback=white]{
\psfig{figure=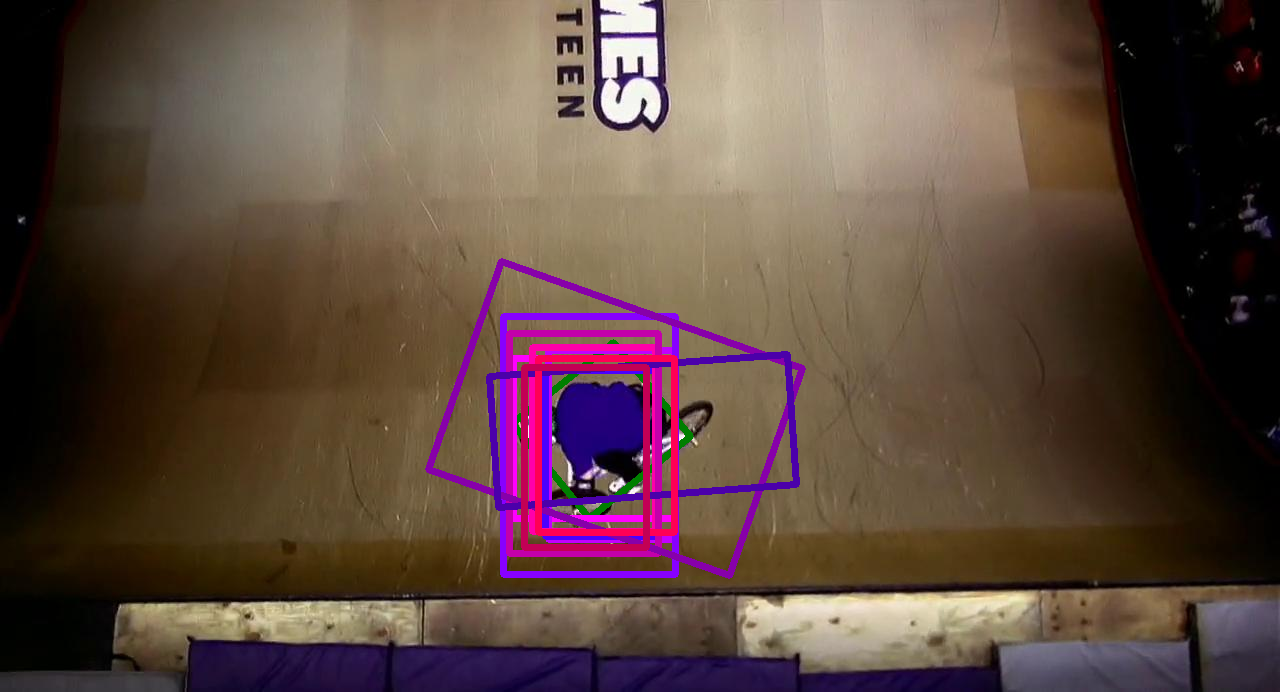, width=0.23\textwidth,height=0.2\textwidth}}}
\vspace{-0.03\textwidth}
\centerline{\hspace{0.03\textwidth} Bmx sequence}
\vspace{0.001\textwidth}
\centerline{ 
\tcbox[sharp corners, size = tight, boxrule=0.2mm, colframe=black, colback=white]{
\psfig{figure=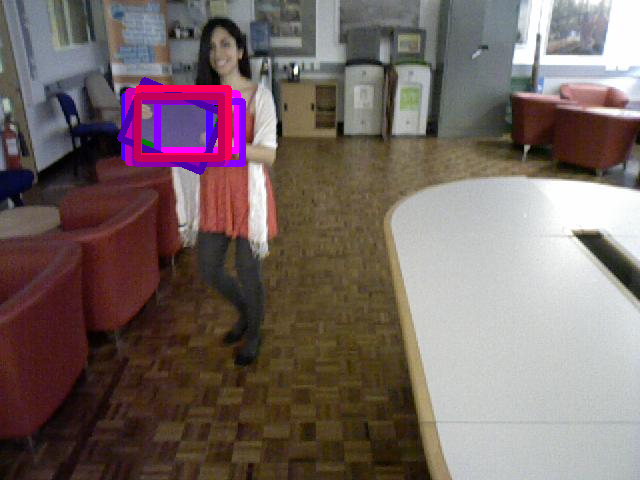, width=0.23\textwidth,height=0.2\textwidth}}
\hspace{0.001\textwidth}
\tcbox[sharp corners, size = tight, boxrule=0.2mm, colframe=black, colback=white]{
\psfig{figure=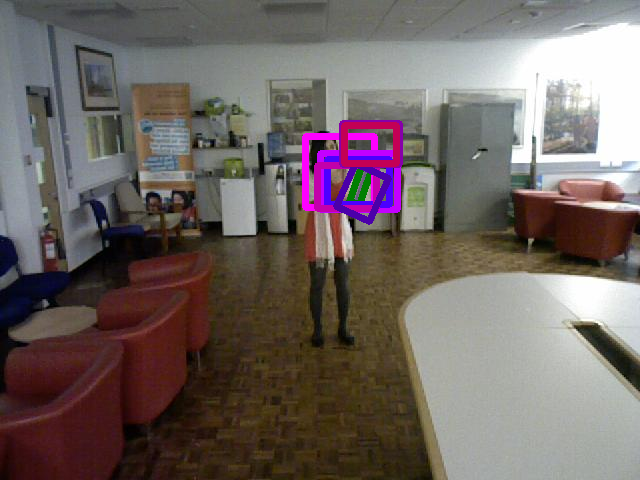, width=0.23\textwidth,height=0.2\textwidth}}
\hspace{0.001\textwidth}
\tcbox[sharp corners, size = tight, boxrule=0.2mm, colframe=black, colback=white]{
\psfig{figure=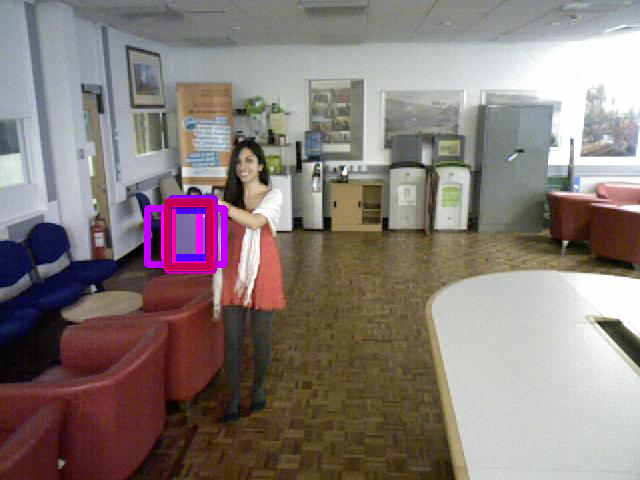, width=0.23\textwidth,height=0.2\textwidth}}
\hspace{0.001\textwidth}
\tcbox[sharp corners, size = tight, boxrule=0.2mm, colframe=black, colback=white]{
\psfig{figure=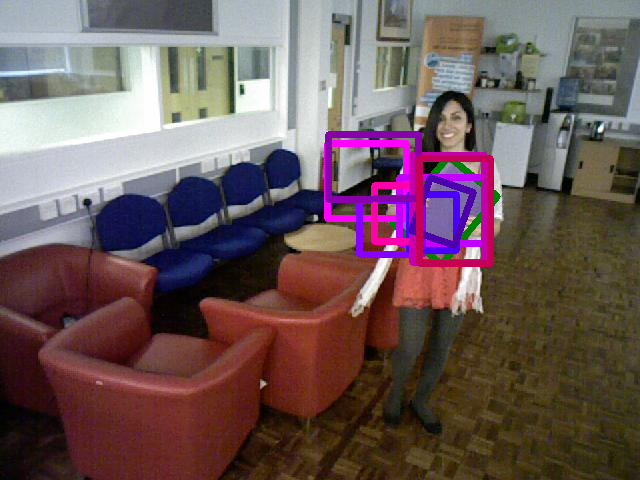, width=0.23\textwidth,height=0.2\textwidth}}}
\vspace{-0.03\textwidth}
\centerline{\hspace{0.03\textwidth} Book sequence}
\vspace{0.001\textwidth}
\centerline{ 
\tcbox[sharp corners, size = tight, boxrule=0.2mm, colframe=black, colback=white]{
\psfig{figure=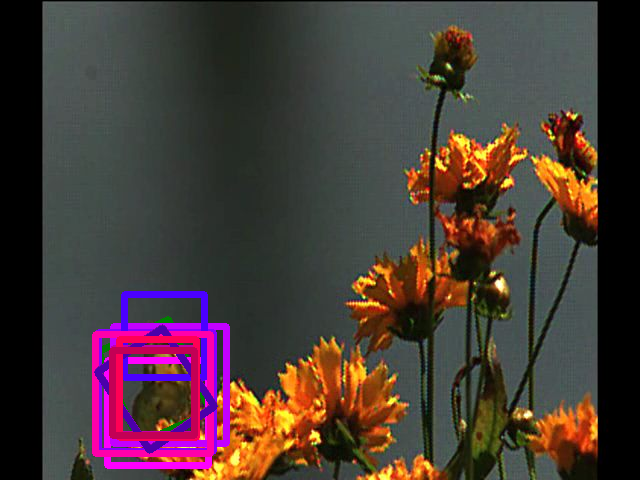, width=0.23\textwidth,height=0.2\textwidth}}
\hspace{0.001\textwidth}
\tcbox[sharp corners, size = tight, boxrule=0.2mm, colframe=black, colback=white]{
\psfig{figure=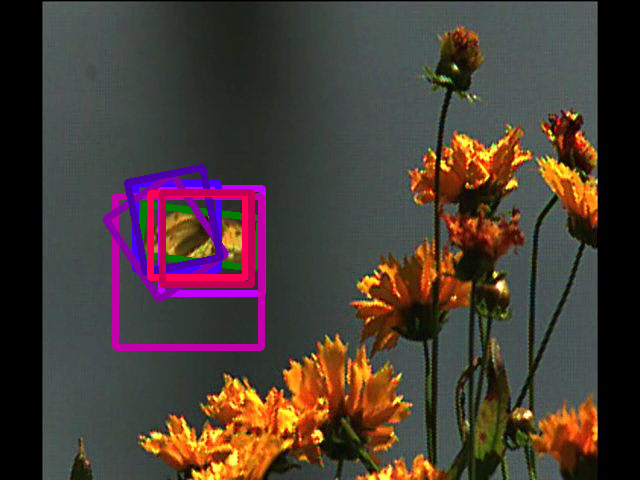, width=0.23\textwidth,height=0.2\textwidth}}
\hspace{0.001\textwidth}
\tcbox[sharp corners, size = tight, boxrule=0.2mm, colframe=black, colback=white]{
\psfig{figure=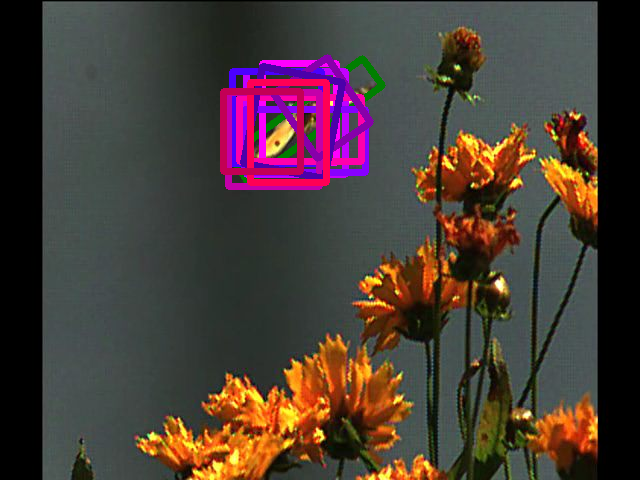, width=0.23\textwidth,height=0.2\textwidth}}
\hspace{0.001\textwidth}
\tcbox[sharp corners, size = tight, boxrule=0.2mm, colframe=black, colback=white]{
\psfig{figure=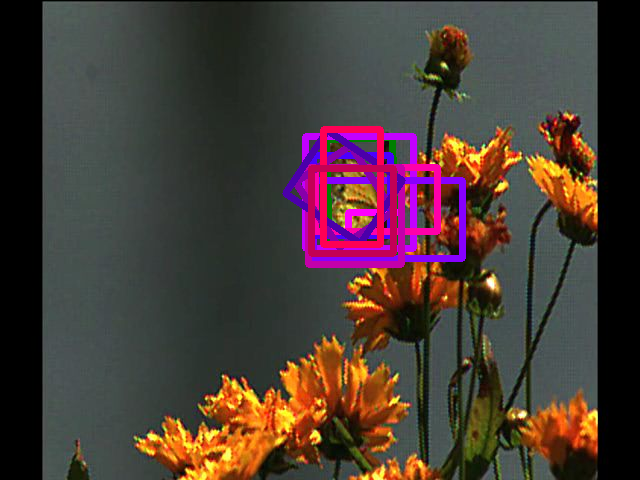, width=0.23\textwidth,height=0.2\textwidth}}}
\vspace{-0.03\textwidth}
\centerline{\hspace{0.03\textwidth} Butterfly sequence}
\caption{Illustrates the tracking results of ten state-of-the-art trackers on Bag, Basketball, Blanket, Bmx, Book, and Butterfly sequences. Each colored rectangle corresponds to each tracker's output. \label{figure_results1}}
\end{figure*}

\begin{figure*}[htp!]
\centerline{ 
\tcbox[sharp corners, size = tight, boxrule=0.2mm, colframe=black, colback=white]{
\psfig{figure=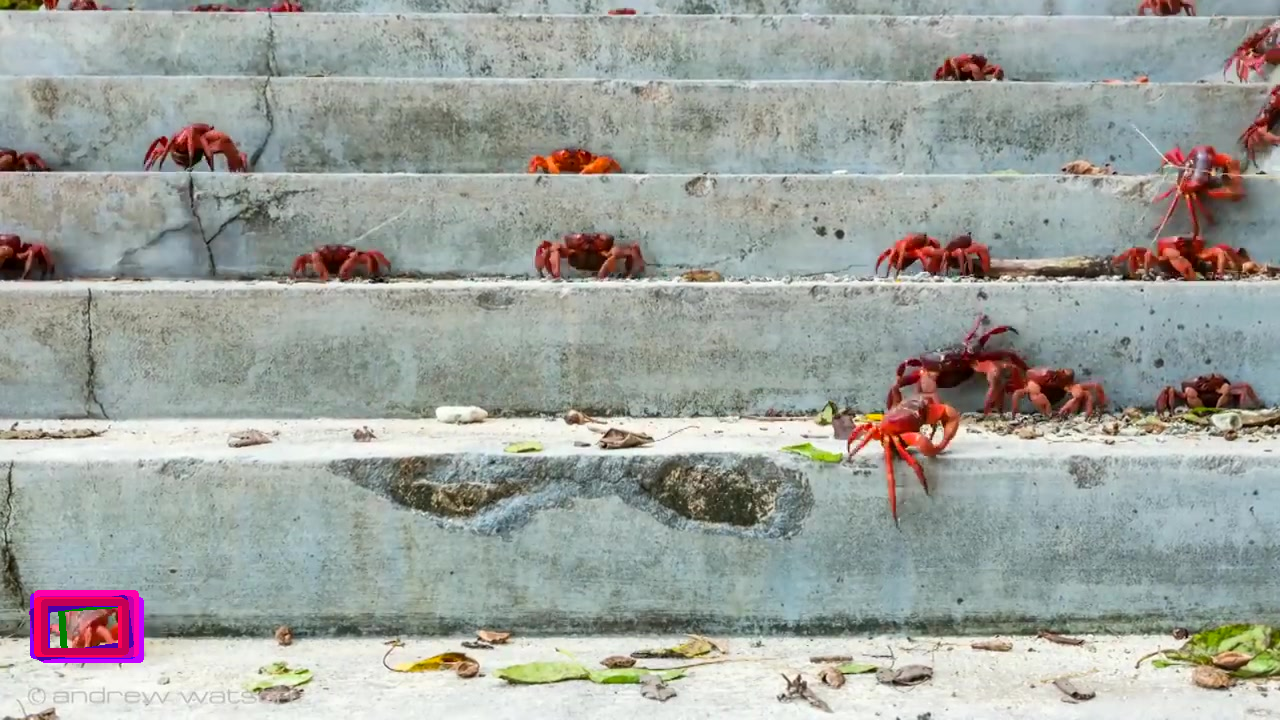, width=0.23\textwidth,height=0.2\textwidth}}
\hspace{0.001\textwidth}
\tcbox[sharp corners, size = tight, boxrule=0.2mm, colframe=black, colback=white]{
\psfig{figure=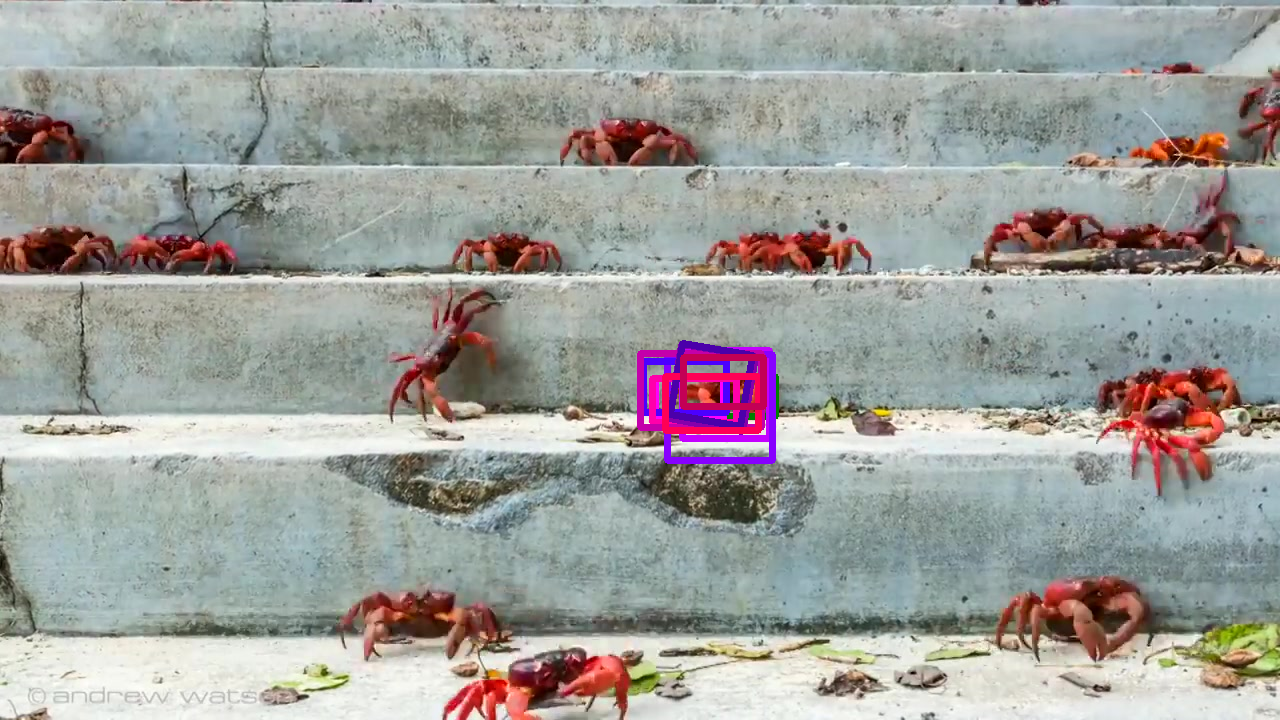, width=0.23\textwidth,height=0.2\textwidth}}
\hspace{0.001\textwidth}
\tcbox[sharp corners, size = tight, boxrule=0.2mm, colframe=black, colback=white]{
\psfig{figure=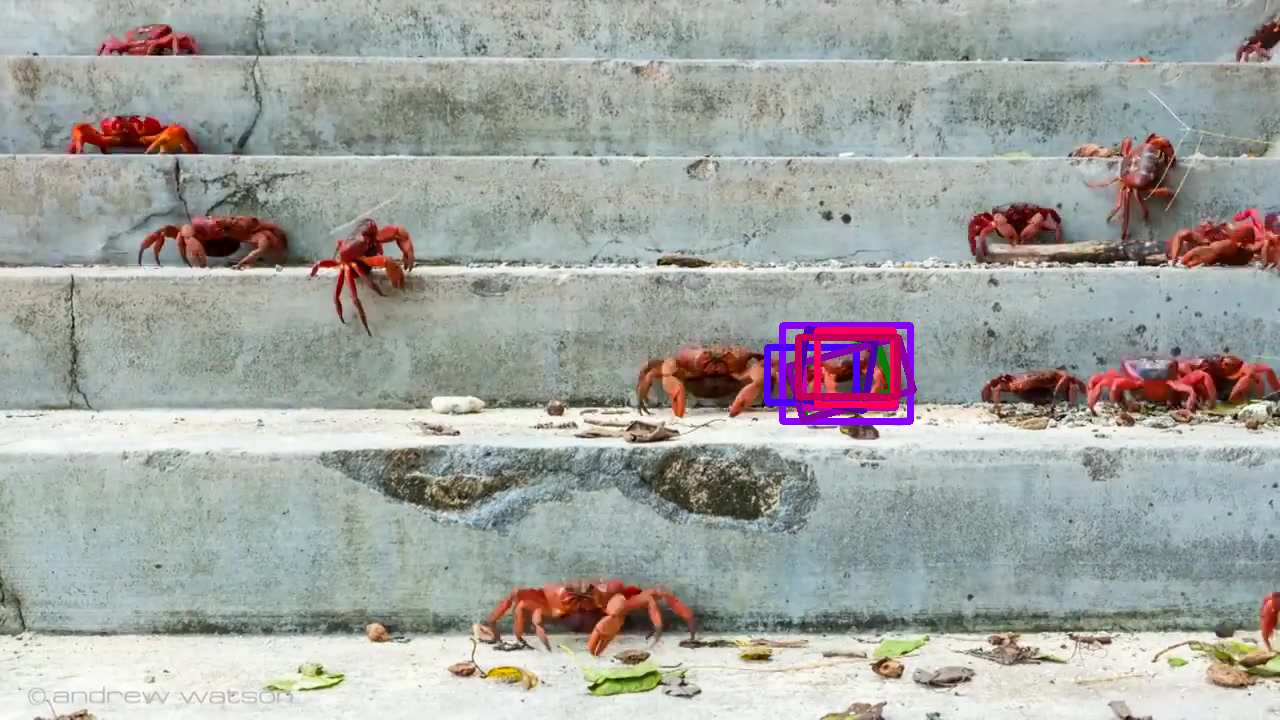, width=0.23\textwidth,height=0.2\textwidth}}
\hspace{0.001\textwidth}
\tcbox[sharp corners, size = tight, boxrule=0.2mm, colframe=black, colback=white]{
\psfig{figure=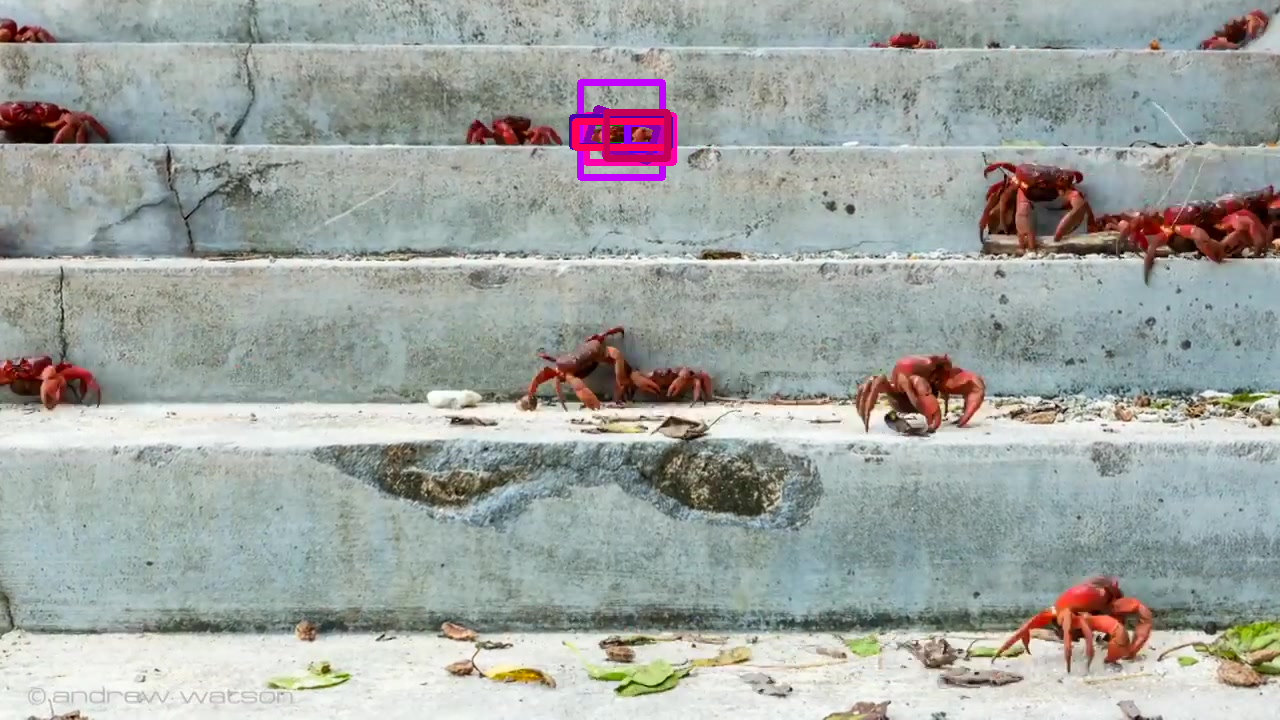, width=0.23\textwidth,height=0.2\textwidth}}}
\vspace{-0.03\textwidth}
\centerline{\hspace{0.03\textwidth} Crabs1 sequence}
\vspace{0.001\textwidth}
\centerline{ 
\tcbox[sharp corners, size = tight, boxrule=0.2mm, colframe=black, colback=white]{
\psfig{figure=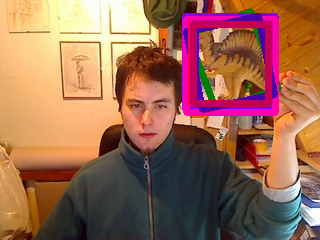, width=0.23\textwidth,height=0.2\textwidth}}
\hspace{0.001\textwidth}
\tcbox[sharp corners, size = tight, boxrule=0.2mm, colframe=black, colback=white]{
\psfig{figure=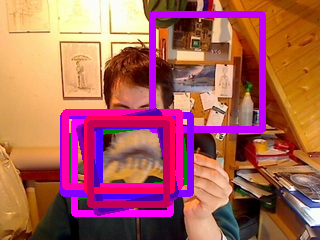, width=0.23\textwidth,height=0.2\textwidth}}
\hspace{0.001\textwidth}
\tcbox[sharp corners, size = tight, boxrule=0.2mm, colframe=black, colback=white]{
\psfig{figure=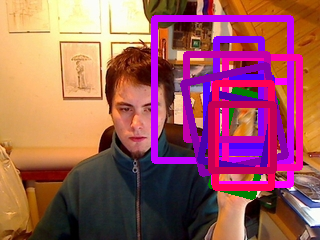, width=0.23\textwidth,height=0.2\textwidth}}
\hspace{0.001\textwidth}
\tcbox[sharp corners, size = tight, boxrule=0.2mm, colframe=black, colback=white]{
\psfig{figure=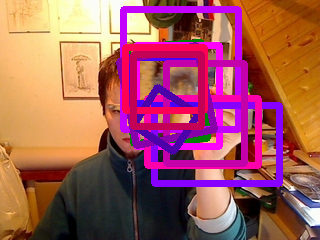, width=0.23\textwidth,height=0.2\textwidth}}}
\vspace{-0.03\textwidth}
\centerline{\hspace{0.03\textwidth} Dinosaur sequence}
\vspace{0.001\textwidth}
\centerline{ 
\tcbox[sharp corners, size = tight, boxrule=0.2mm, colframe=black, colback=white]{
\psfig{figure=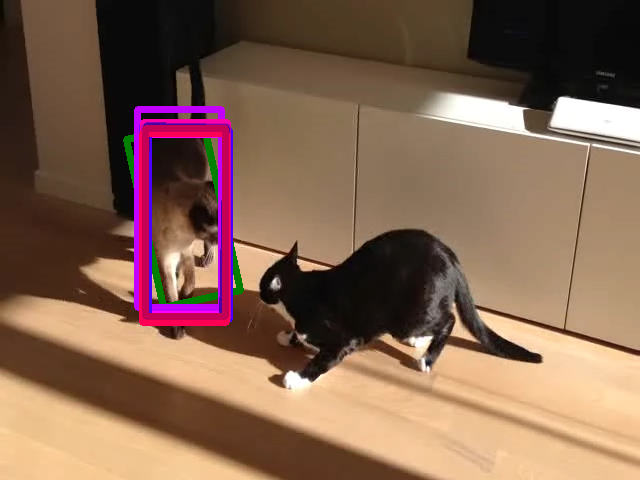, width=0.23\textwidth,height=0.2\textwidth}}
\hspace{0.001\textwidth}
\tcbox[sharp corners, size = tight, boxrule=0.2mm, colframe=black, colback=white]{
\psfig{figure=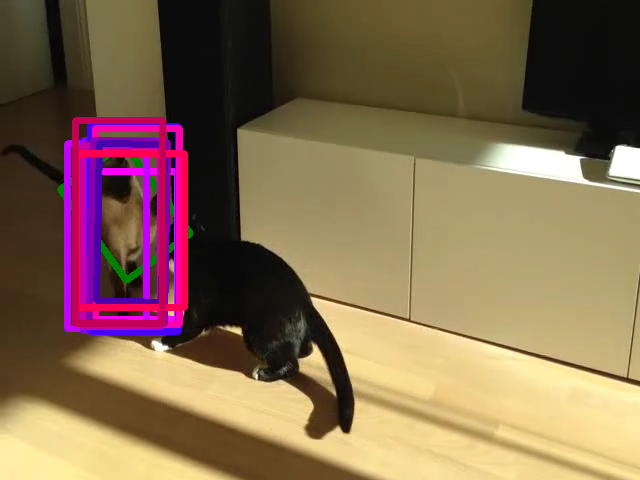, width=0.23\textwidth,height=0.2\textwidth}}
\hspace{0.001\textwidth}
\tcbox[sharp corners, size = tight, boxrule=0.2mm, colframe=black, colback=white]{
\psfig{figure=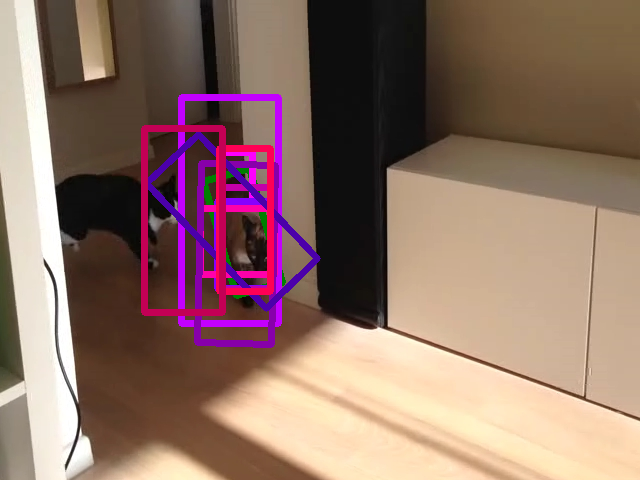, width=0.23\textwidth,height=0.2\textwidth}}
\hspace{0.001\textwidth}
\tcbox[sharp corners, size = tight, boxrule=0.2mm, colframe=black, colback=white]{
\psfig{figure=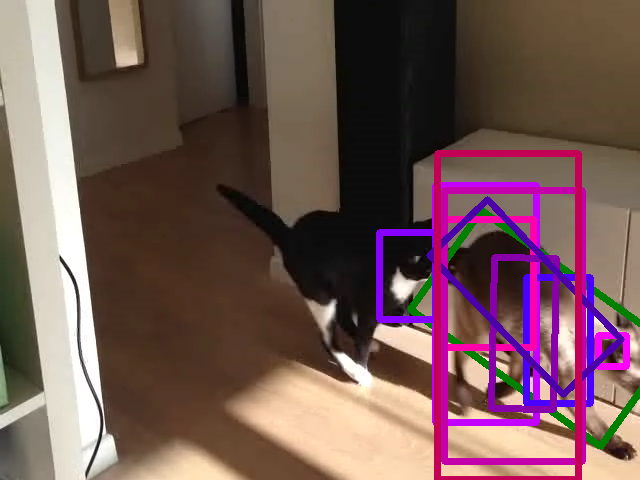, width=0.23\textwidth,height=0.2\textwidth}}}
\vspace{-0.03\textwidth}
\centerline{\hspace{0.03\textwidth} Fernando sequence}
\vspace{0.001\textwidth}
\centerline{ 
\tcbox[sharp corners, size = tight, boxrule=0.2mm, colframe=black, colback=white]{
\psfig{figure=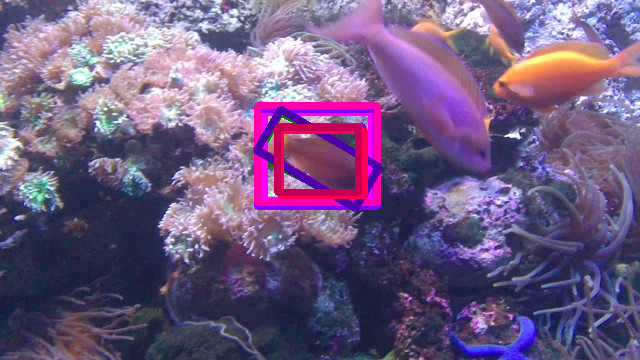, width=0.23\textwidth,height=0.2\textwidth}}
\hspace{0.001\textwidth}
\tcbox[sharp corners, size = tight, boxrule=0.2mm, colframe=black, colback=white]{
\psfig{figure=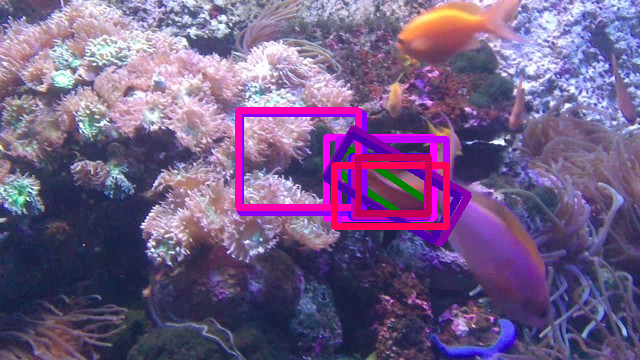, width=0.23\textwidth,height=0.2\textwidth}}
\hspace{0.001\textwidth}
\tcbox[sharp corners, size = tight, boxrule=0.2mm, colframe=black, colback=white]{
\psfig{figure=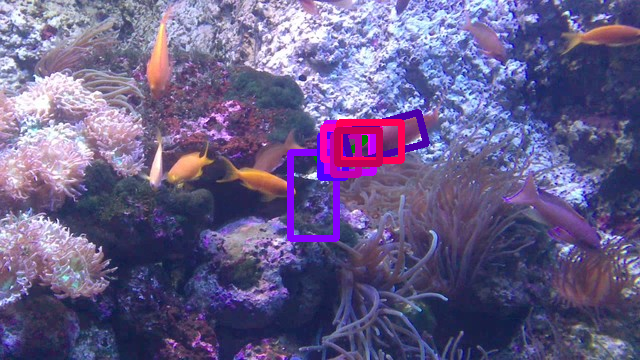, width=0.23\textwidth,height=0.2\textwidth}}
\hspace{0.001\textwidth}
\tcbox[sharp corners, size = tight, boxrule=0.2mm, colframe=black, colback=white]{
\psfig{figure=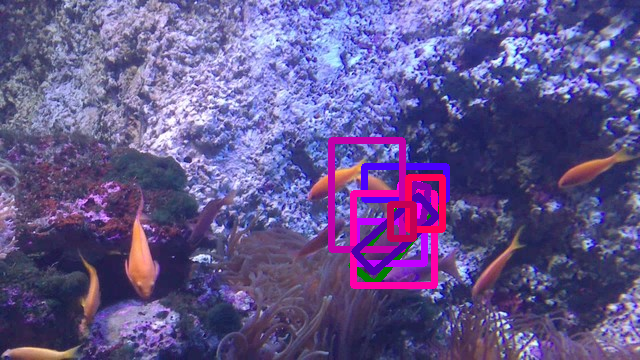, width=0.23\textwidth,height=0.2\textwidth}}}
\vspace{-0.03\textwidth}
\centerline{\hspace{0.03\textwidth} Fish2 sequence}
\vspace{0.001\textwidth}
\centerline{ 
\tcbox[sharp corners, size = tight, boxrule=0.2mm, colframe=black, colback=white]{
\psfig{figure=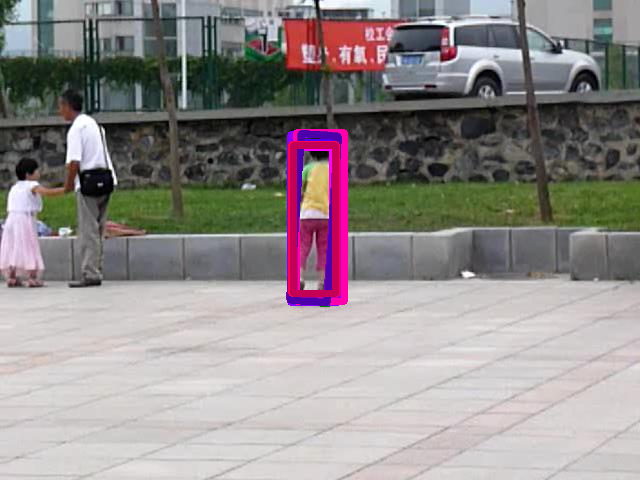, width=0.23\textwidth,height=0.2\textwidth}}
\hspace{0.001\textwidth}
\tcbox[sharp corners, size = tight, boxrule=0.2mm, colframe=black, colback=white]{
\psfig{figure=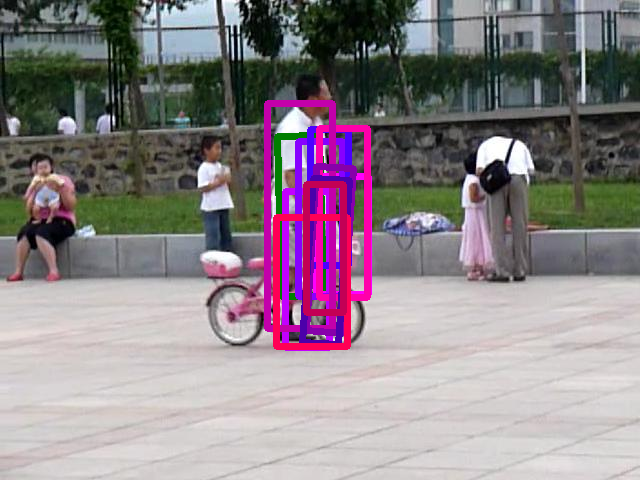, width=0.23\textwidth,height=0.2\textwidth}}
\hspace{0.001\textwidth}
\tcbox[sharp corners, size = tight, boxrule=0.2mm, colframe=black, colback=white]{
\psfig{figure=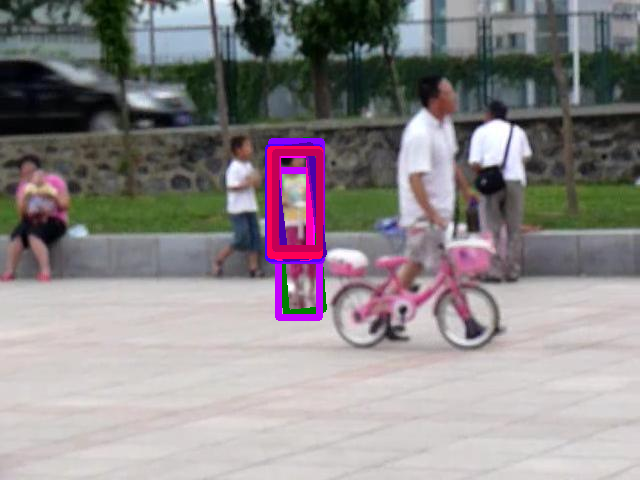, width=0.23\textwidth,height=0.2\textwidth}}
\hspace{0.001\textwidth}
\tcbox[sharp corners, size = tight, boxrule=0.2mm, colframe=black, colback=white]{
\psfig{figure=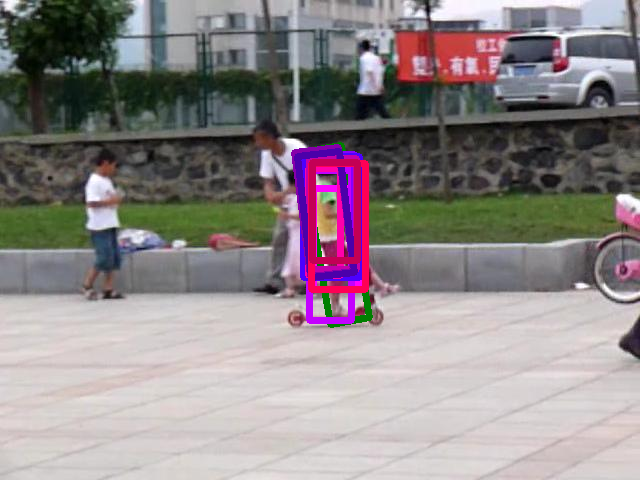, width=0.23\textwidth,height=0.2\textwidth}}}
\vspace{-0.03\textwidth}
\centerline{\hspace{0.03\textwidth} Girl sequence}
\vspace{0.001\textwidth}
\centerline{ 
\tcbox[sharp corners, size = tight, boxrule=0.2mm, colframe=black, colback=white]{
\psfig{figure=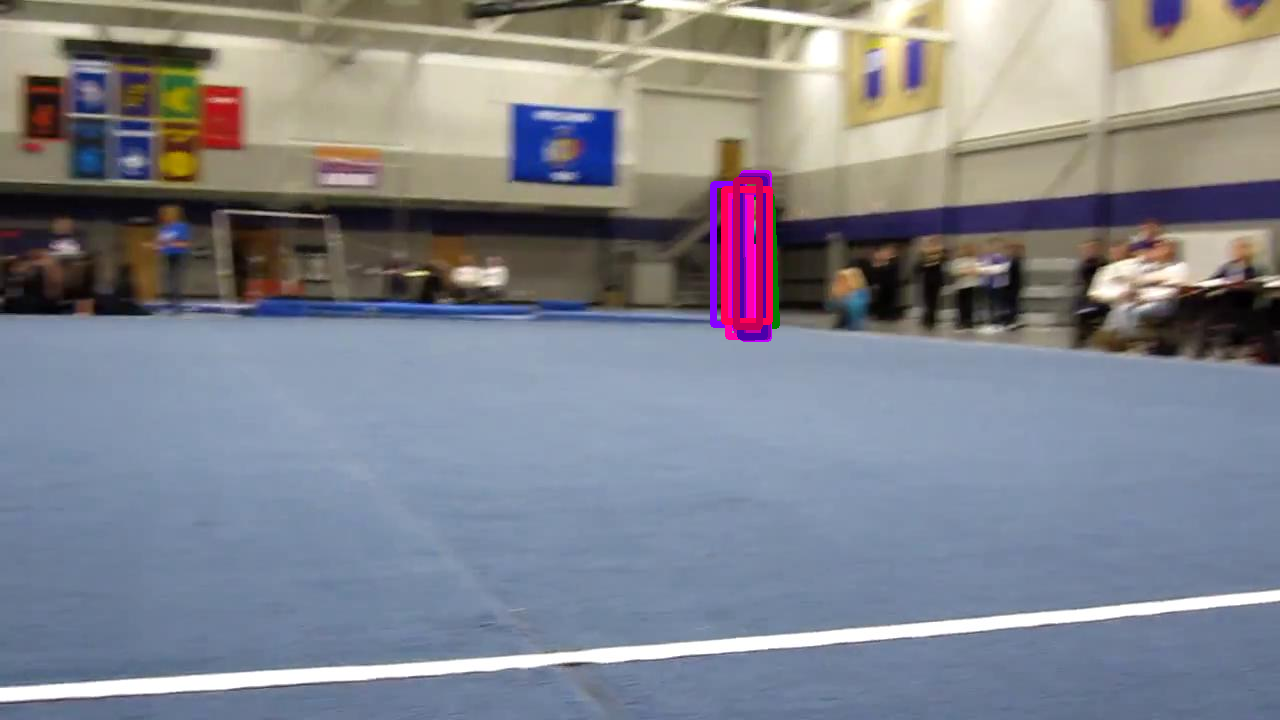, width=0.23\textwidth,height=0.2\textwidth}}
\hspace{0.001\textwidth}
\tcbox[sharp corners, size = tight, boxrule=0.2mm, colframe=black, colback=white]{
\psfig{figure=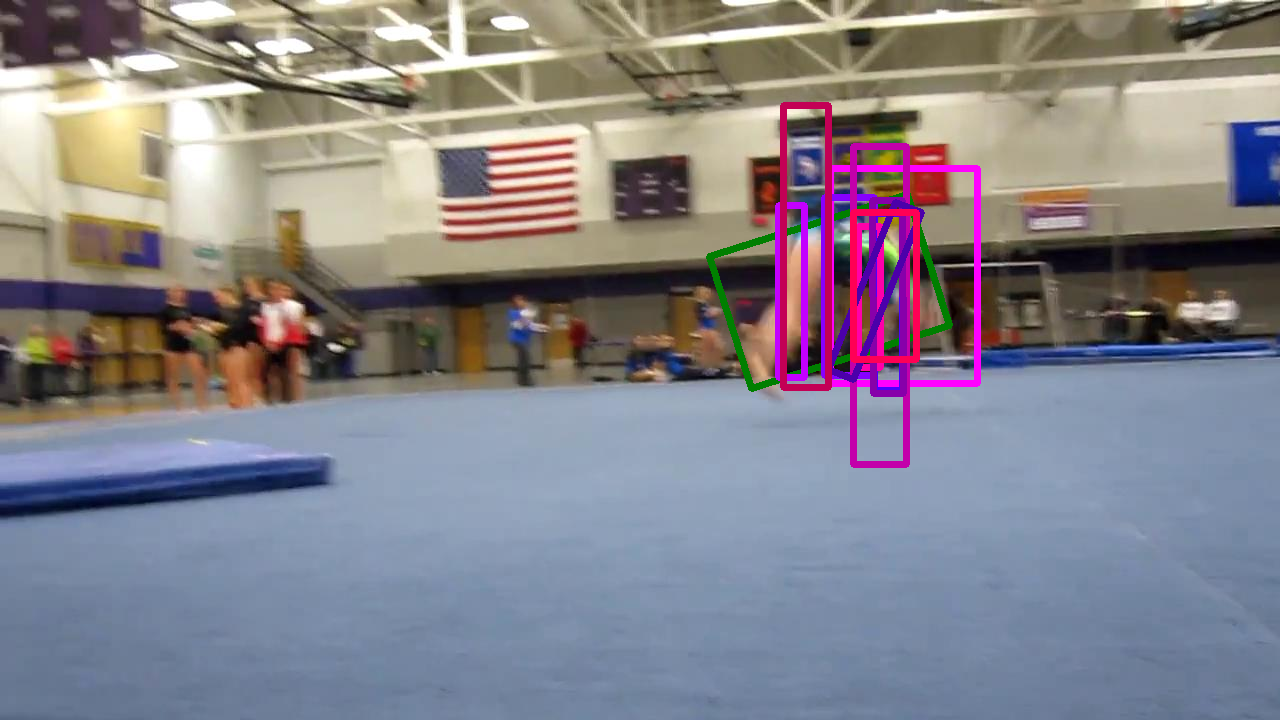, width=0.23\textwidth,height=0.2\textwidth}}
\hspace{0.001\textwidth}
\tcbox[sharp corners, size = tight, boxrule=0.2mm, colframe=black, colback=white]{
\psfig{figure=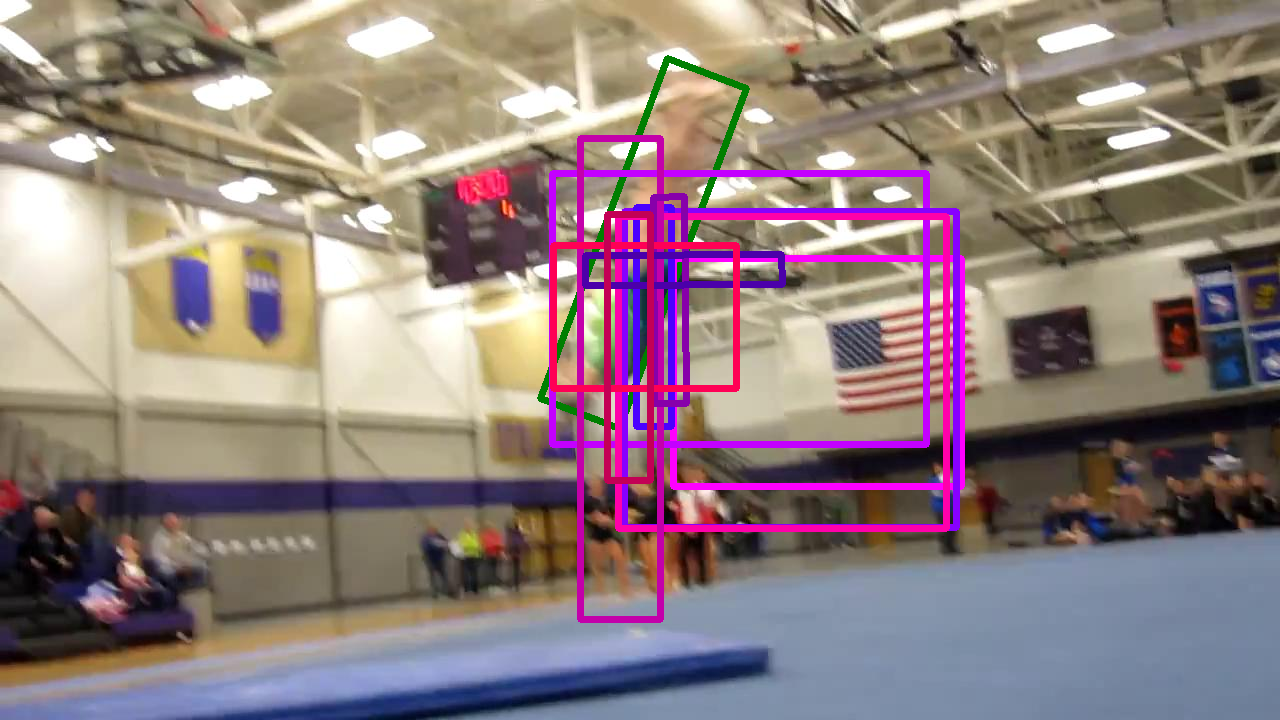, width=0.23\textwidth,height=0.2\textwidth}}
\hspace{0.001\textwidth}
\tcbox[sharp corners, size = tight, boxrule=0.2mm, colframe=black, colback=white]{
\psfig{figure=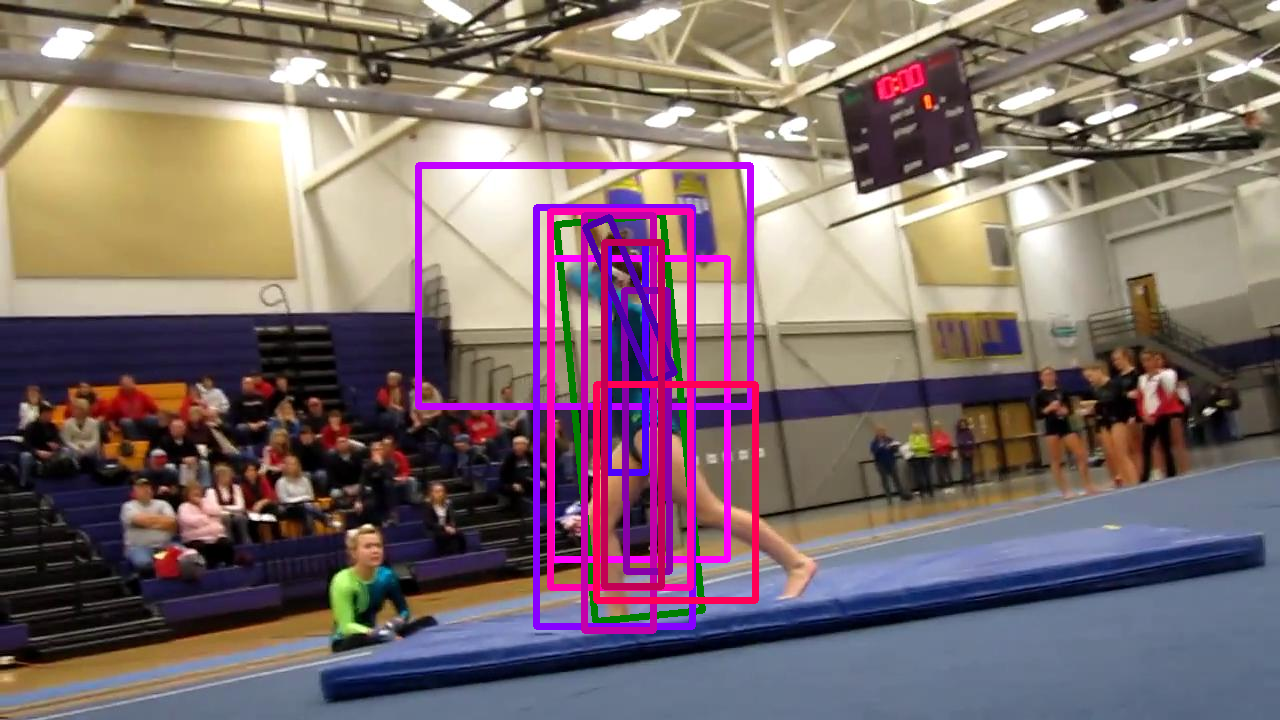, width=0.23\textwidth,height=0.2\textwidth}}}
\vspace{-0.03\textwidth}
\centerline{\hspace{0.03\textwidth} Gymnastics2 sequence}
\caption{Illustrates the tracking results of ten state-of-the-art trackers on Crabs1, Dinosaur, Fernando, Fish2, Girl, and Gymnastics2 sequences. Each colored rectangle corresponds to each tracker's output. \label{figure_results2}}
\end{figure*}

\begin{figure*}[htp!]
\centerline{ 
\tcbox[sharp corners, size = tight, boxrule=0.2mm, colframe=black, colback=white]{
\psfig{figure=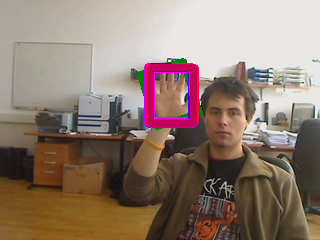, width=0.23\textwidth,height=0.2\textwidth}}
\hspace{0.001\textwidth}
\tcbox[sharp corners, size = tight, boxrule=0.2mm, colframe=black, colback=white]{
\psfig{figure=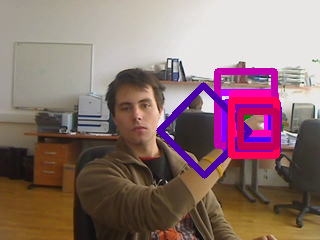, width=0.23\textwidth,height=0.2\textwidth}}
\hspace{0.001\textwidth}
\tcbox[sharp corners, size = tight, boxrule=0.2mm, colframe=black, colback=white]{
\psfig{figure=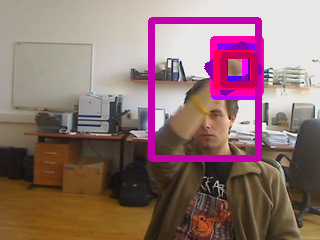, width=0.23\textwidth,height=0.2\textwidth}}
\hspace{0.001\textwidth}
\tcbox[sharp corners, size = tight, boxrule=0.2mm, colframe=black, colback=white]{
\psfig{figure=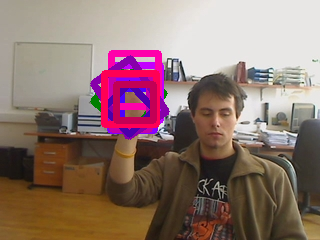, width=0.23\textwidth,height=0.2\textwidth}}}
\vspace{-0.03\textwidth}
\centerline{\hspace{0.03\textwidth} Hand sequence}
\vspace{0.001\textwidth}
\centerline{ 
\tcbox[sharp corners, size = tight, boxrule=0.2mm, colframe=black, colback=white]{
\psfig{figure=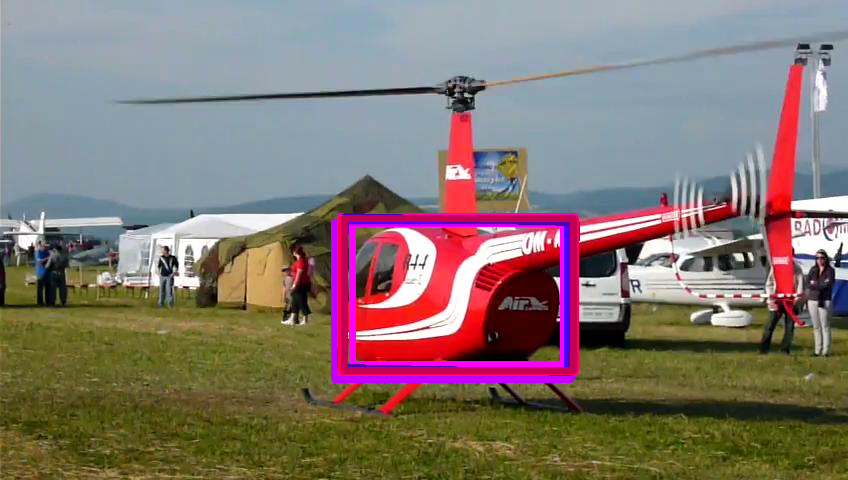, width=0.23\textwidth,height=0.2\textwidth}}
\hspace{0.001\textwidth}
\tcbox[sharp corners, size = tight, boxrule=0.2mm, colframe=black, colback=white]{
\psfig{figure=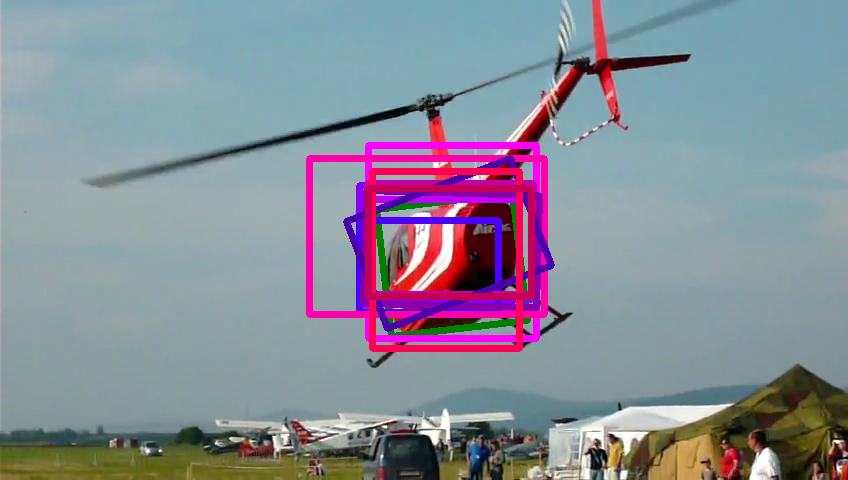, width=0.23\textwidth,height=0.2\textwidth}}
\hspace{0.001\textwidth}
\tcbox[sharp corners, size = tight, boxrule=0.2mm, colframe=black, colback=white]{
\psfig{figure=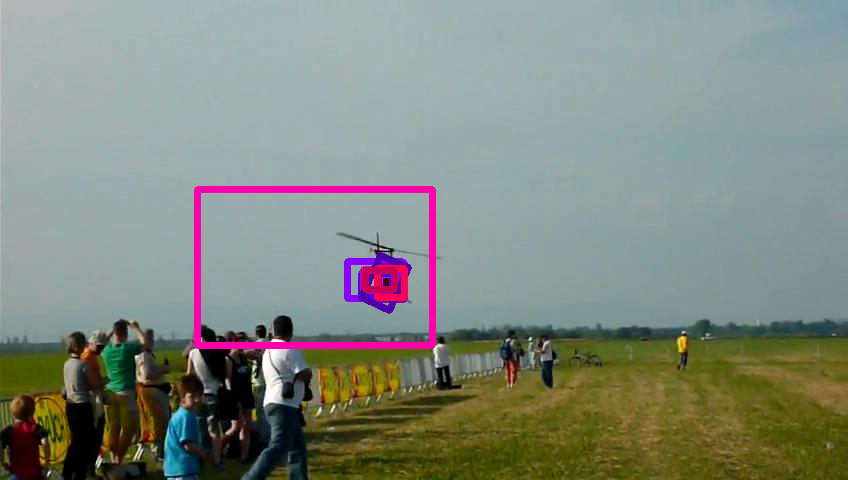, width=0.23\textwidth,height=0.2\textwidth}}
\hspace{0.001\textwidth}
\tcbox[sharp corners, size = tight, boxrule=0.2mm, colframe=black, colback=white]{
\psfig{figure=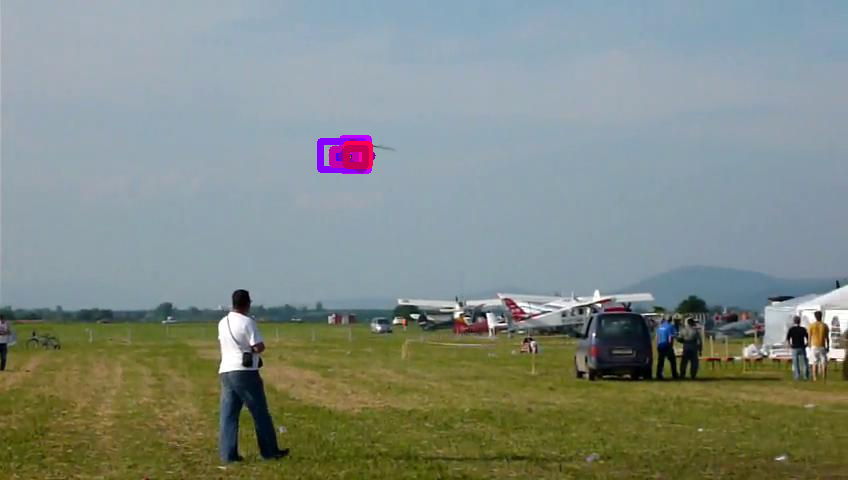, width=0.23\textwidth,height=0.2\textwidth}}}
\vspace{-0.03\textwidth}
\centerline{\hspace{0.03\textwidth} Helicopter sequence}
\vspace{0.001\textwidth}
\centerline{ 
\tcbox[sharp corners, size = tight, boxrule=0.2mm, colframe=black, colback=white]{
\psfig{figure=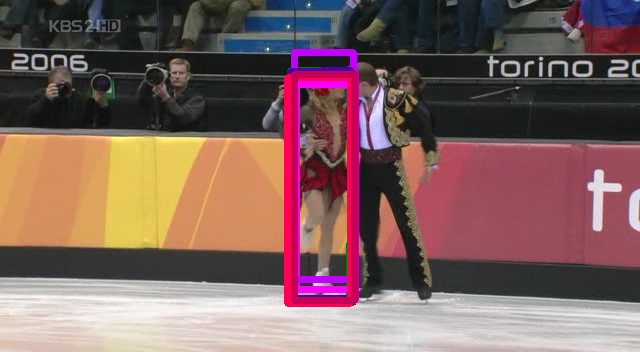, width=0.23\textwidth,height=0.2\textwidth}}
\hspace{0.001\textwidth}
\tcbox[sharp corners, size = tight, boxrule=0.2mm, colframe=black, colback=white]{
\psfig{figure=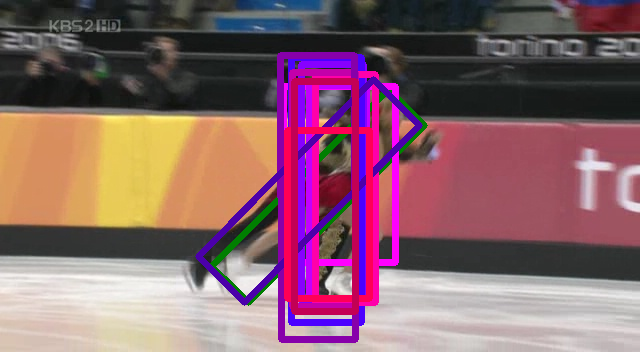, width=0.23\textwidth,height=0.2\textwidth}}
\hspace{0.001\textwidth}
\tcbox[sharp corners, size = tight, boxrule=0.2mm, colframe=black, colback=white]{
\psfig{figure=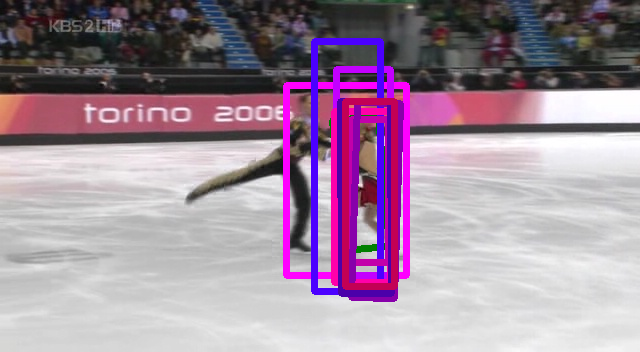, width=0.23\textwidth,height=0.2\textwidth}}
\hspace{0.001\textwidth}
\tcbox[sharp corners, size = tight, boxrule=0.2mm, colframe=black, colback=white]{
\psfig{figure=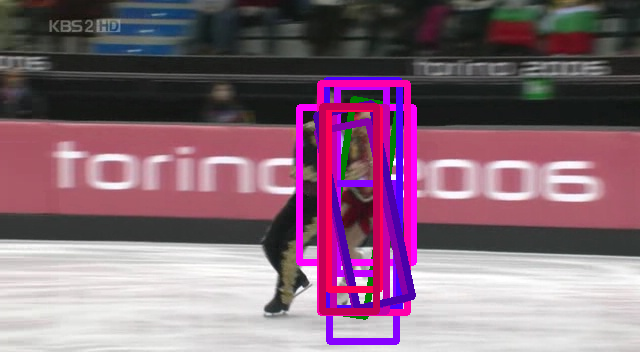, width=0.23\textwidth,height=0.2\textwidth}}}
\vspace{-0.03\textwidth}
\centerline{\hspace{0.03\textwidth} Iceskater2 sequence}
\vspace{0.001\textwidth}
\centerline{ 
\tcbox[sharp corners, size = tight, boxrule=0.2mm, colframe=black, colback=white]{
\psfig{figure=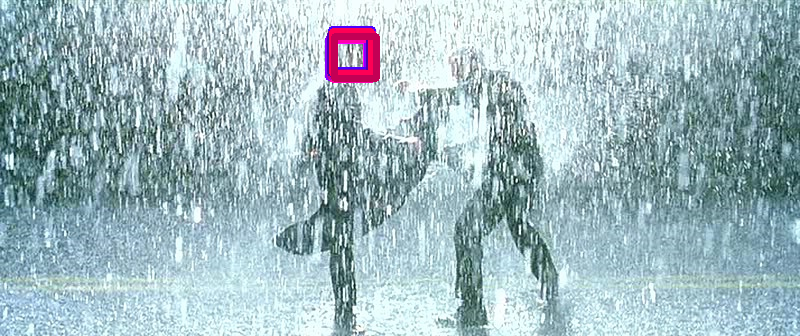, width=0.23\textwidth,height=0.2\textwidth}}
\hspace{0.001\textwidth}
\tcbox[sharp corners, size = tight, boxrule=0.2mm, colframe=black, colback=white]{
\psfig{figure=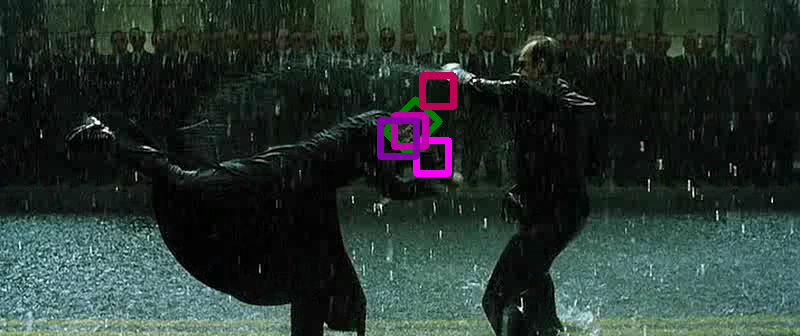, width=0.23\textwidth,height=0.2\textwidth}}
\hspace{0.001\textwidth}
\tcbox[sharp corners, size = tight, boxrule=0.2mm, colframe=black, colback=white]{
\psfig{figure=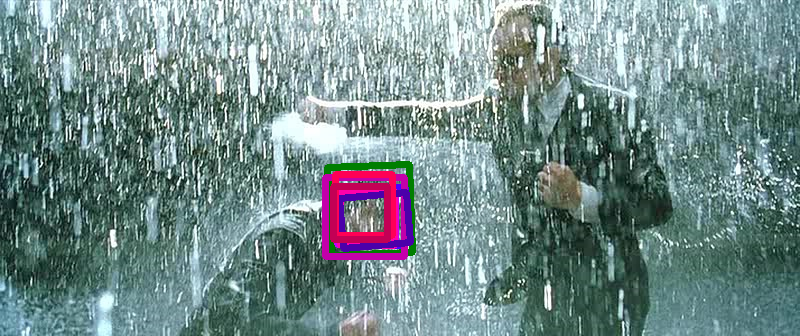, width=0.23\textwidth,height=0.2\textwidth}}
\hspace{0.001\textwidth}
\tcbox[sharp corners, size = tight, boxrule=0.2mm, colframe=black, colback=white]{
\psfig{figure=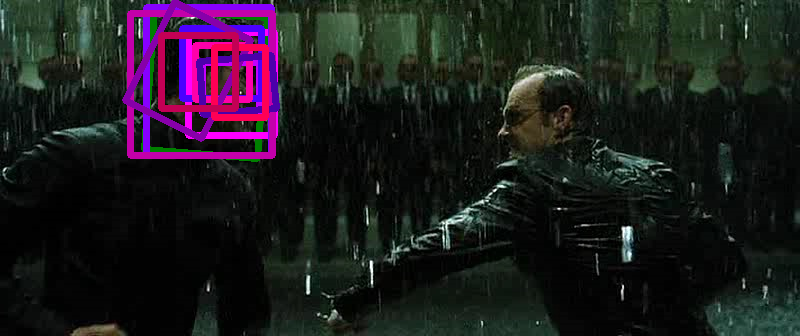, width=0.23\textwidth,height=0.2\textwidth}}}
\vspace{-0.03\textwidth}
\centerline{\hspace{0.03\textwidth} Matrix sequence}
\vspace{0.001\textwidth}
\centerline{ 
\tcbox[sharp corners, size = tight, boxrule=0.2mm, colframe=black, colback=white]{
\psfig{figure=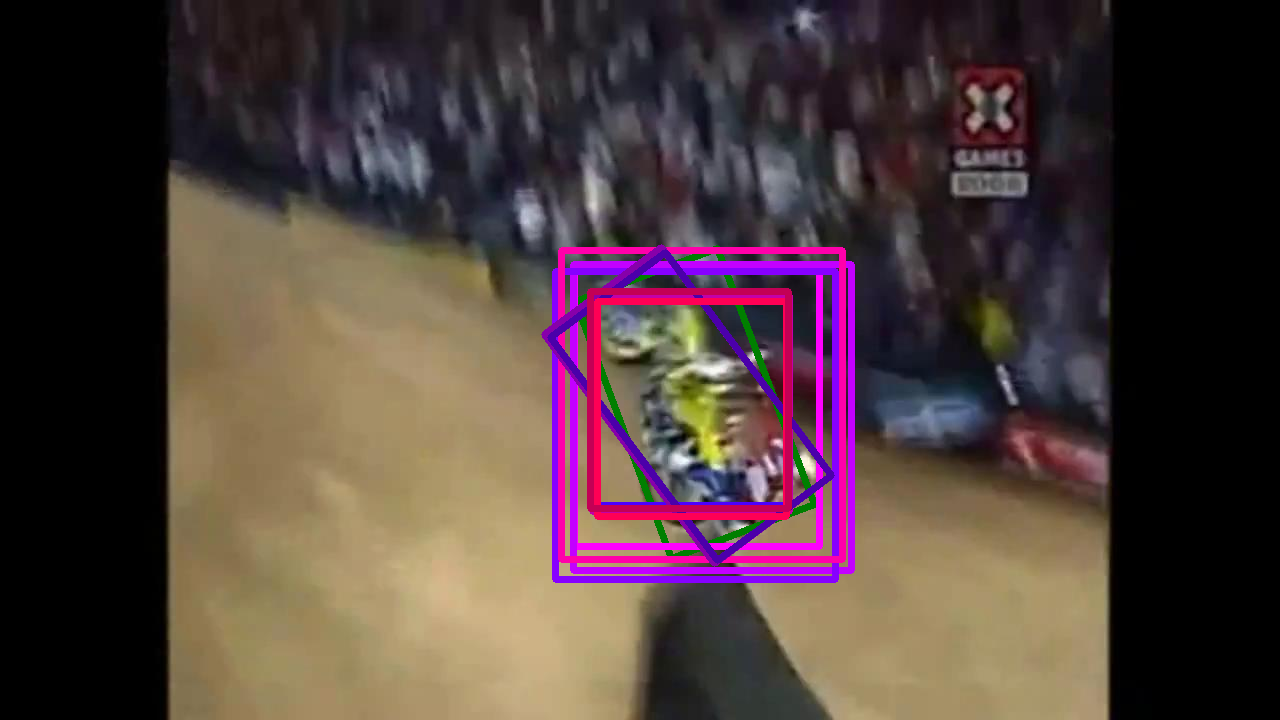, width=0.23\textwidth,height=0.2\textwidth}}
\hspace{0.001\textwidth}
\tcbox[sharp corners, size = tight, boxrule=0.2mm, colframe=black, colback=white]{
\psfig{figure=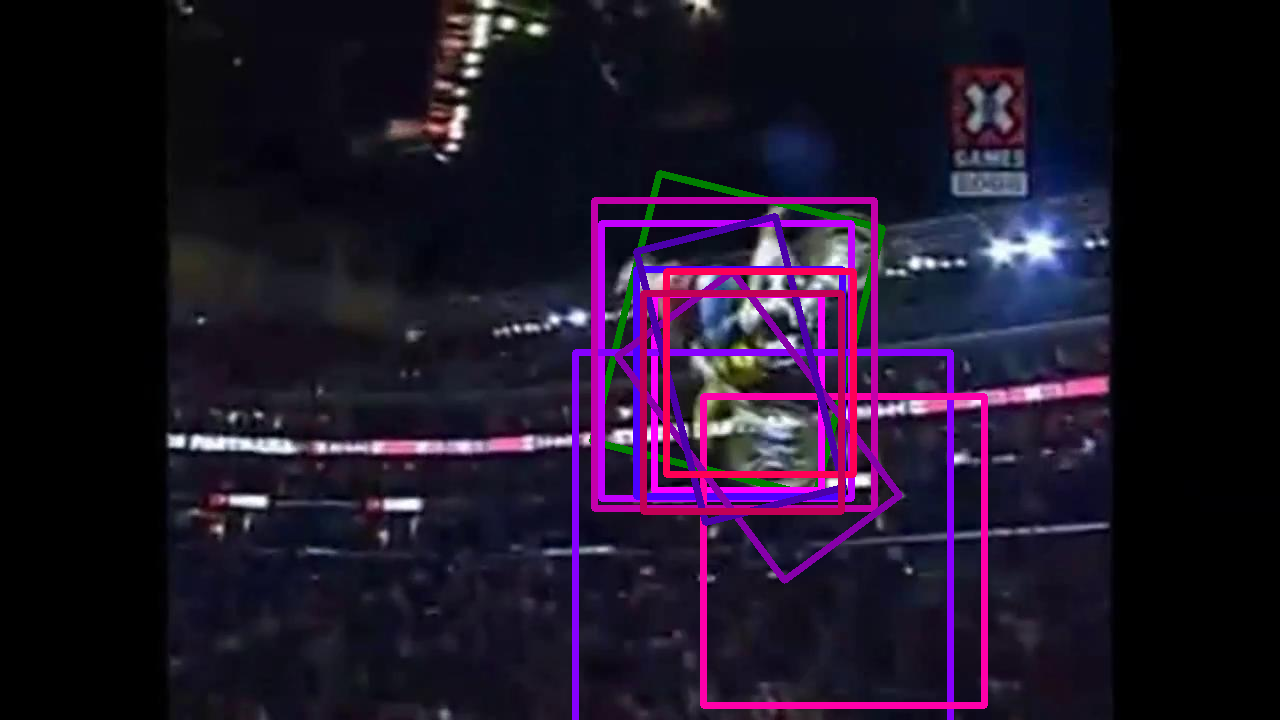, width=0.23\textwidth,height=0.2\textwidth}}
\hspace{0.001\textwidth}
\tcbox[sharp corners, size = tight, boxrule=0.2mm, colframe=black, colback=white]{
\psfig{figure=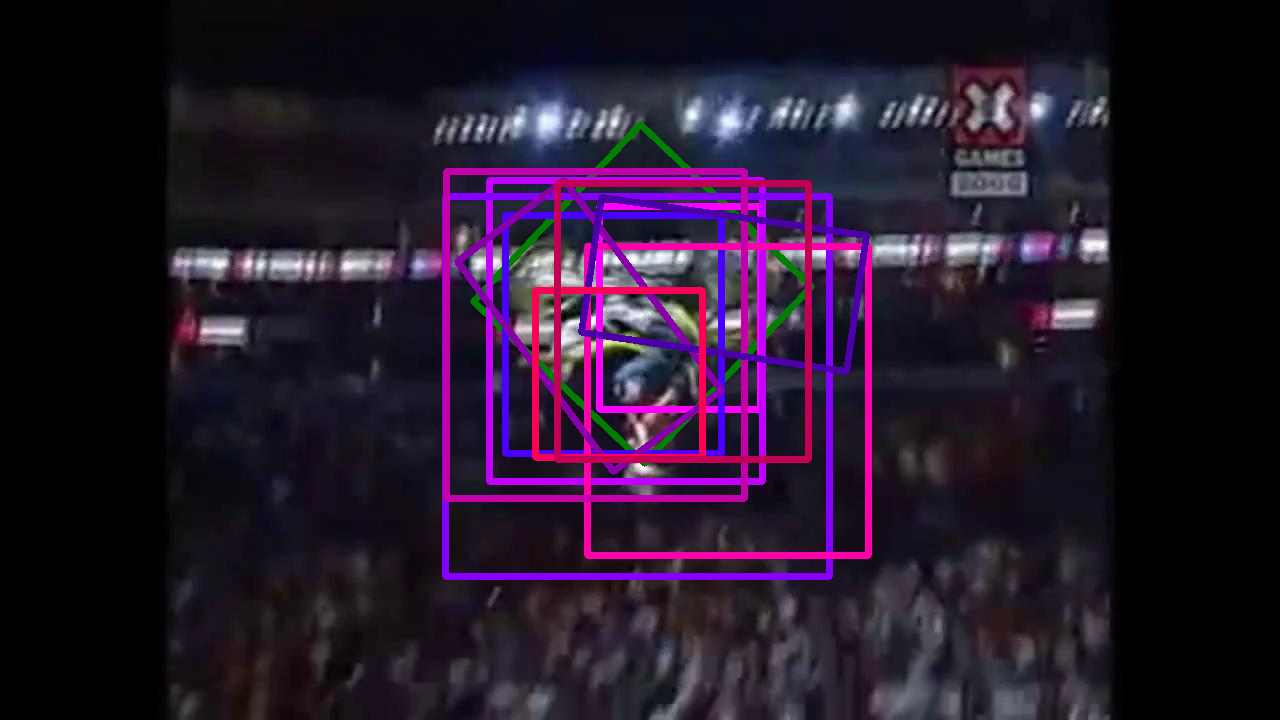, width=0.23\textwidth,height=0.2\textwidth}}
\hspace{0.001\textwidth}
\tcbox[sharp corners, size = tight, boxrule=0.2mm, colframe=black, colback=white]{
\psfig{figure=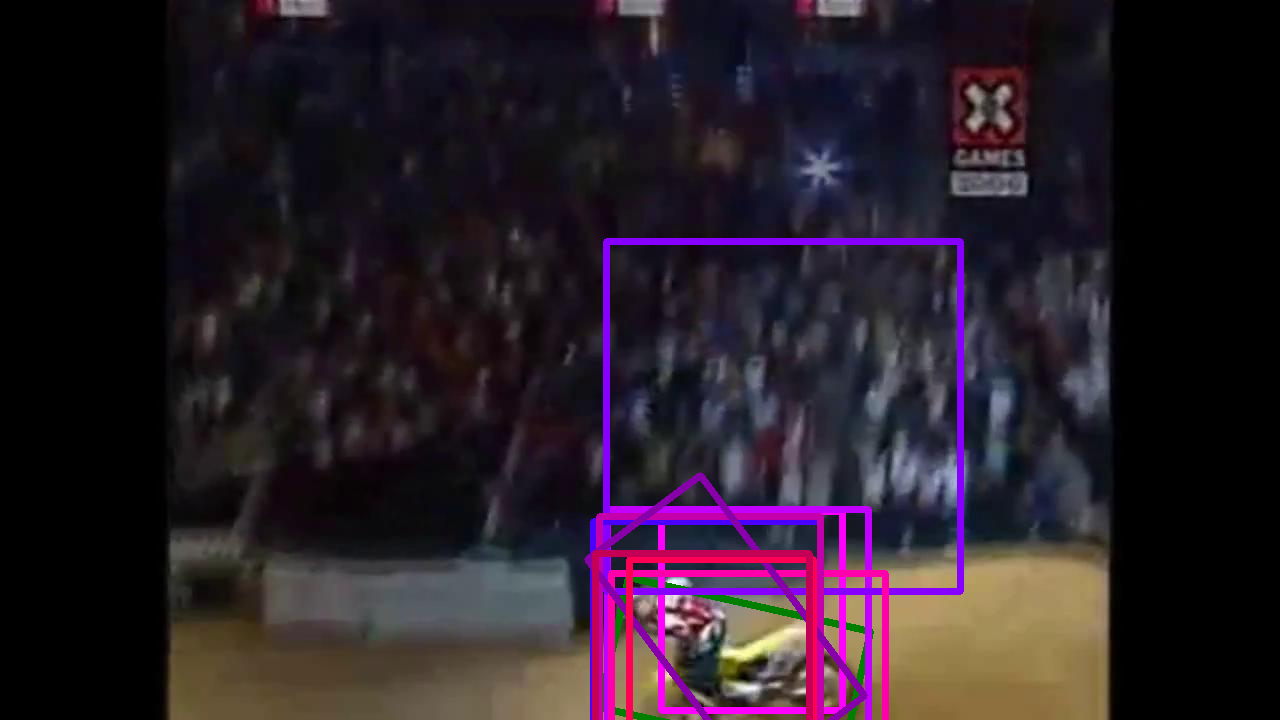, width=0.23\textwidth,height=0.2\textwidth}}}
\vspace{-0.03\textwidth}
\centerline{\hspace{0.03\textwidth} Motocross2 sequence}
\vspace{0.001\textwidth}
\centerline{ 
\tcbox[sharp corners, size = tight, boxrule=0.2mm, colframe=black, colback=white]{
\psfig{figure=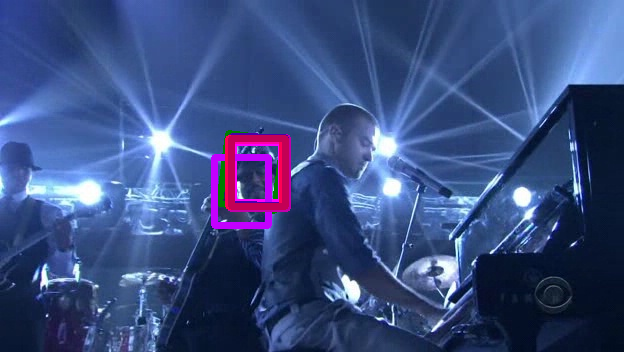, width=0.23\textwidth,height=0.2\textwidth}}
\hspace{0.001\textwidth}
\tcbox[sharp corners, size = tight, boxrule=0.2mm, colframe=black, colback=white]{
\psfig{figure=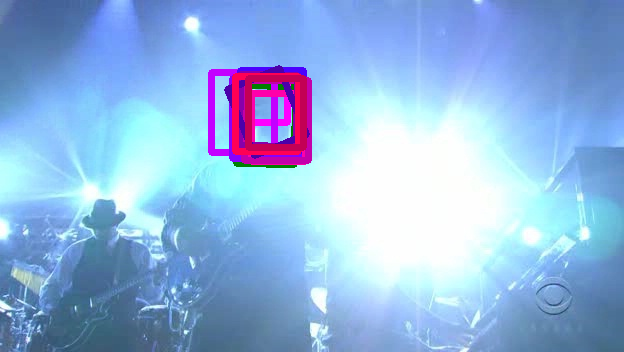, width=0.23\textwidth,height=0.2\textwidth}}
\hspace{0.001\textwidth}
\tcbox[sharp corners, size = tight, boxrule=0.2mm, colframe=black, colback=white]{
\psfig{figure=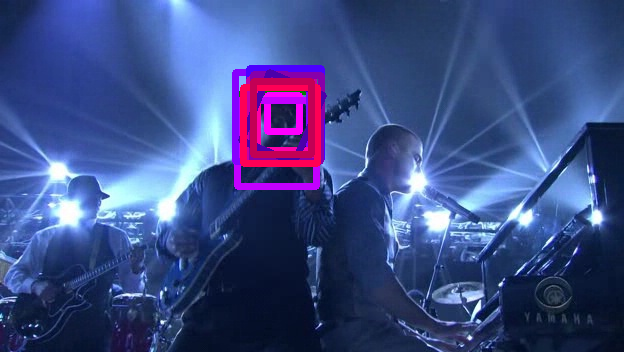, width=0.23\textwidth,height=0.2\textwidth}}
\hspace{0.001\textwidth}
\tcbox[sharp corners, size = tight, boxrule=0.2mm, colframe=black, colback=white]{
\psfig{figure=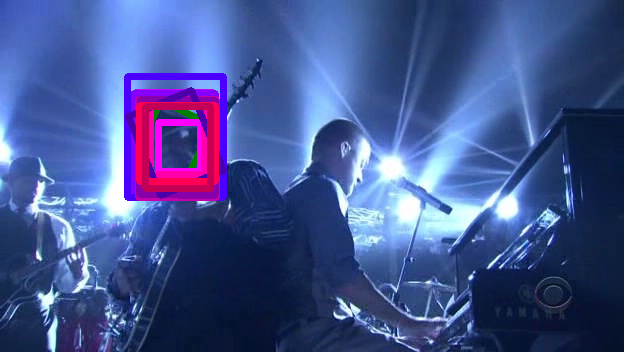, width=0.23\textwidth,height=0.2\textwidth}}}
\vspace{-0.03\textwidth}
\centerline{\hspace{0.03\textwidth} Shaking sequence}
\caption{Illustrates the tracking results of ten state-of-the-art trackers on Hand, Helicopter, Iceskater2, Matrix, Motocross2, and Shaking sequences. Each colored rectangle corresponds to each tracker's output. \label{figure_results3}}
\end{figure*}

\begin{figure*}[htp!]
\centerline{ 
\tcbox[sharp corners, size = tight, boxrule=0.2mm, colframe=black, colback=white]{
\psfig{figure=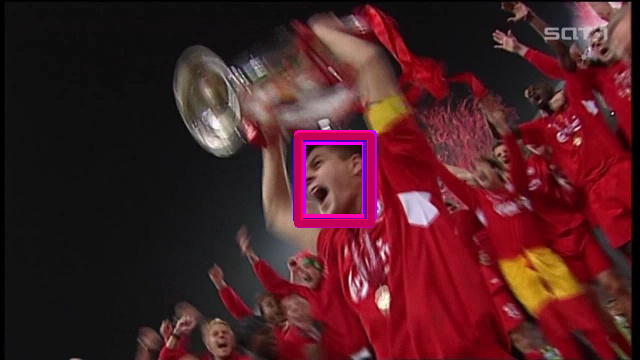, width=0.23\textwidth,height=0.2\textwidth}}
\hspace{0.001\textwidth}
\tcbox[sharp corners, size = tight, boxrule=0.2mm, colframe=black, colback=white]{
\psfig{figure=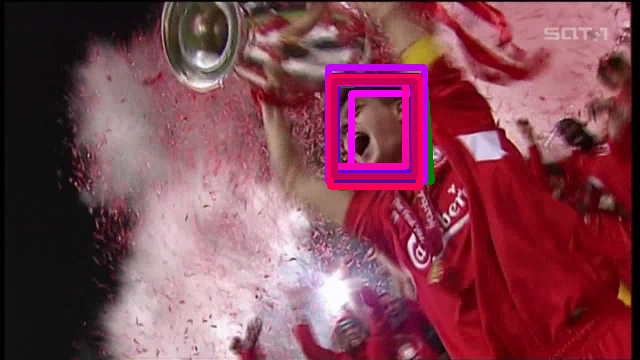, width=0.23\textwidth,height=0.2\textwidth}}
\hspace{0.001\textwidth}
\tcbox[sharp corners, size = tight, boxrule=0.2mm, colframe=black, colback=white]{
\psfig{figure=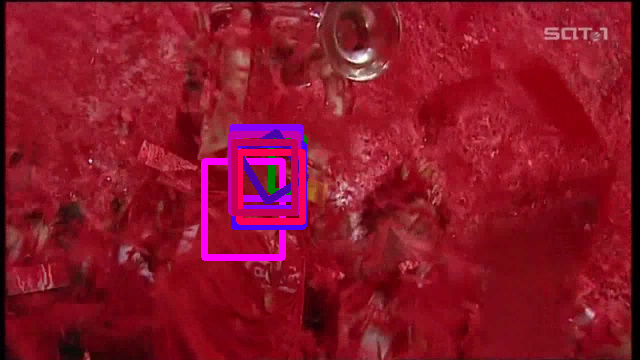, width=0.23\textwidth,height=0.2\textwidth}}
\hspace{0.001\textwidth}
\tcbox[sharp corners, size = tight, boxrule=0.2mm, colframe=black, colback=white]{
\psfig{figure=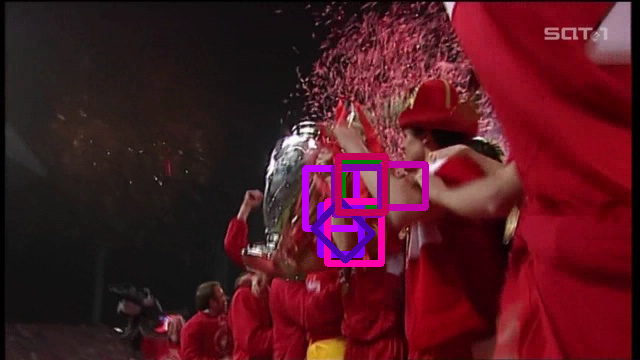, width=0.23\textwidth,height=0.2\textwidth}}}
\vspace{-0.03\textwidth}
\centerline{\hspace{0.03\textwidth} Soccer1 sequence}
\vspace{0.001\textwidth}
\centerline{ 
\tcbox[sharp corners, size = tight, boxrule=0.2mm, colframe=black, colback=white]{
\psfig{figure=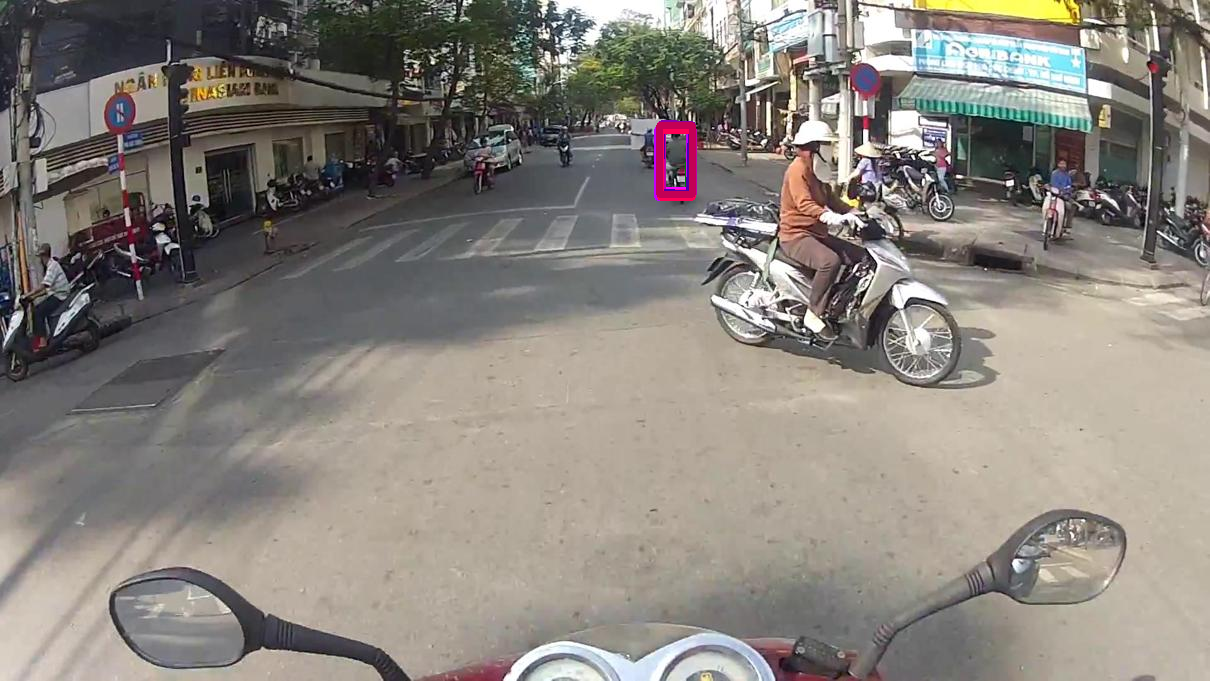, width=0.23\textwidth,height=0.2\textwidth}}
\hspace{0.001\textwidth}
\tcbox[sharp corners, size = tight, boxrule=0.2mm, colframe=black, colback=white]{
\psfig{figure=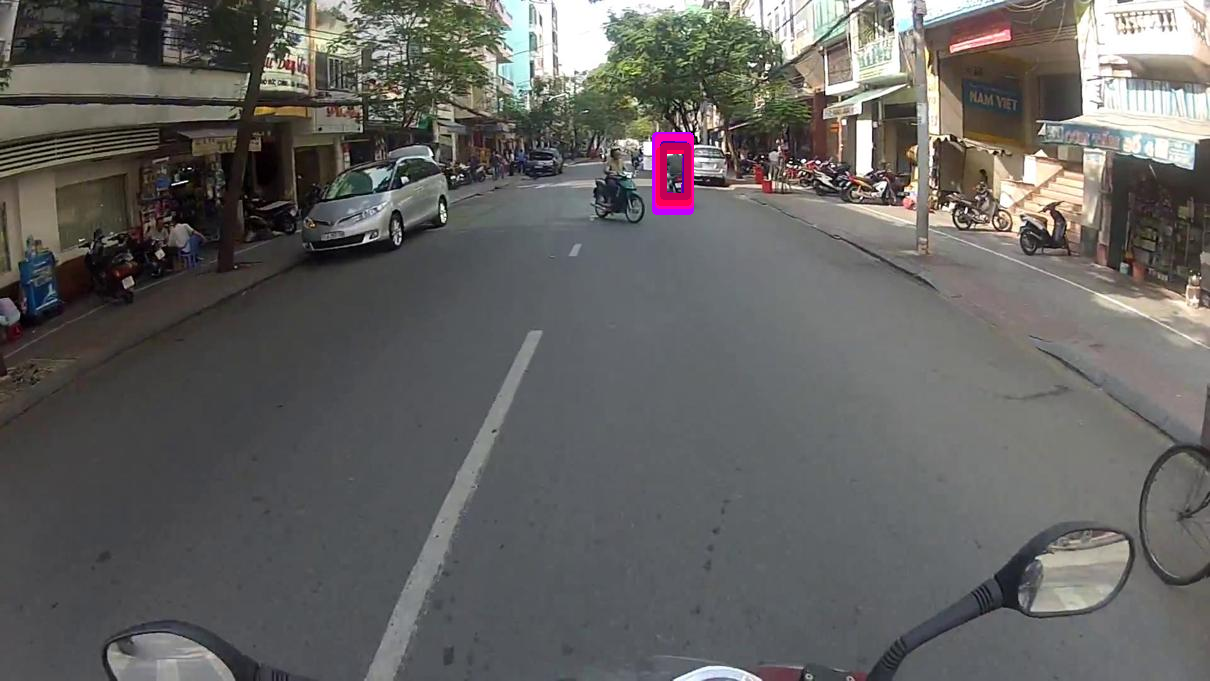, width=0.23\textwidth,height=0.2\textwidth}}
\hspace{0.001\textwidth}
\tcbox[sharp corners, size = tight, boxrule=0.2mm, colframe=black, colback=white]{
\psfig{figure=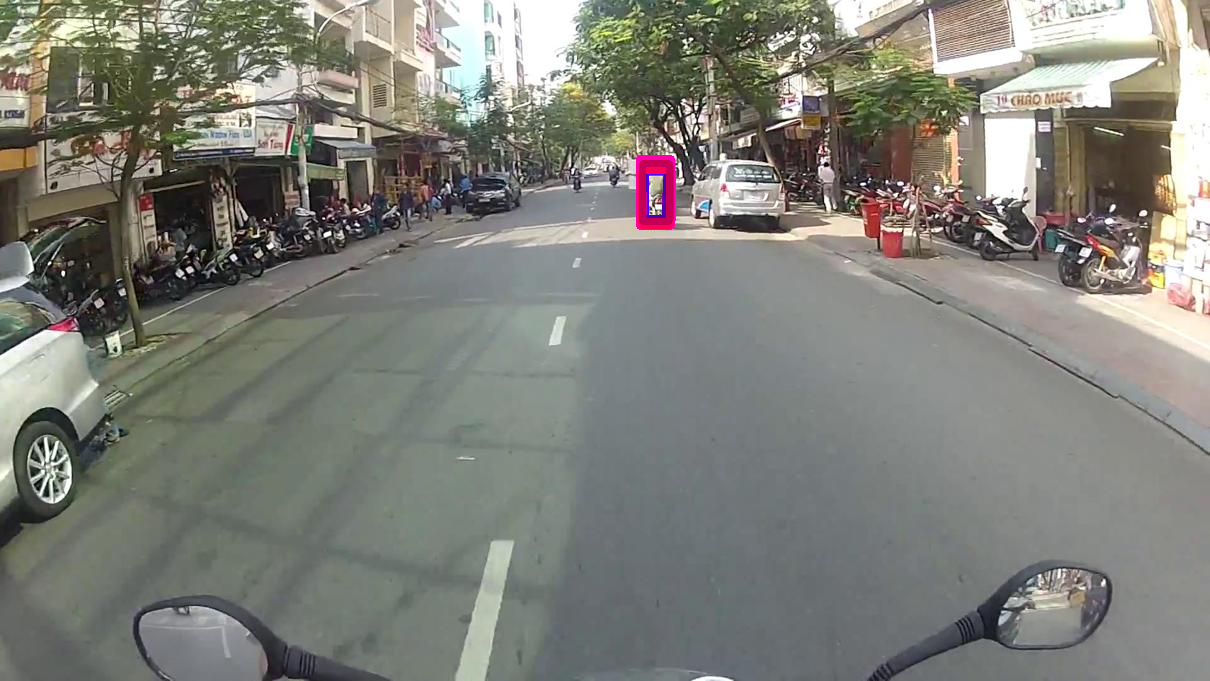, width=0.23\textwidth,height=0.2\textwidth}}
\hspace{0.001\textwidth}
\tcbox[sharp corners, size = tight, boxrule=0.2mm, colframe=black, colback=white]{
\psfig{figure=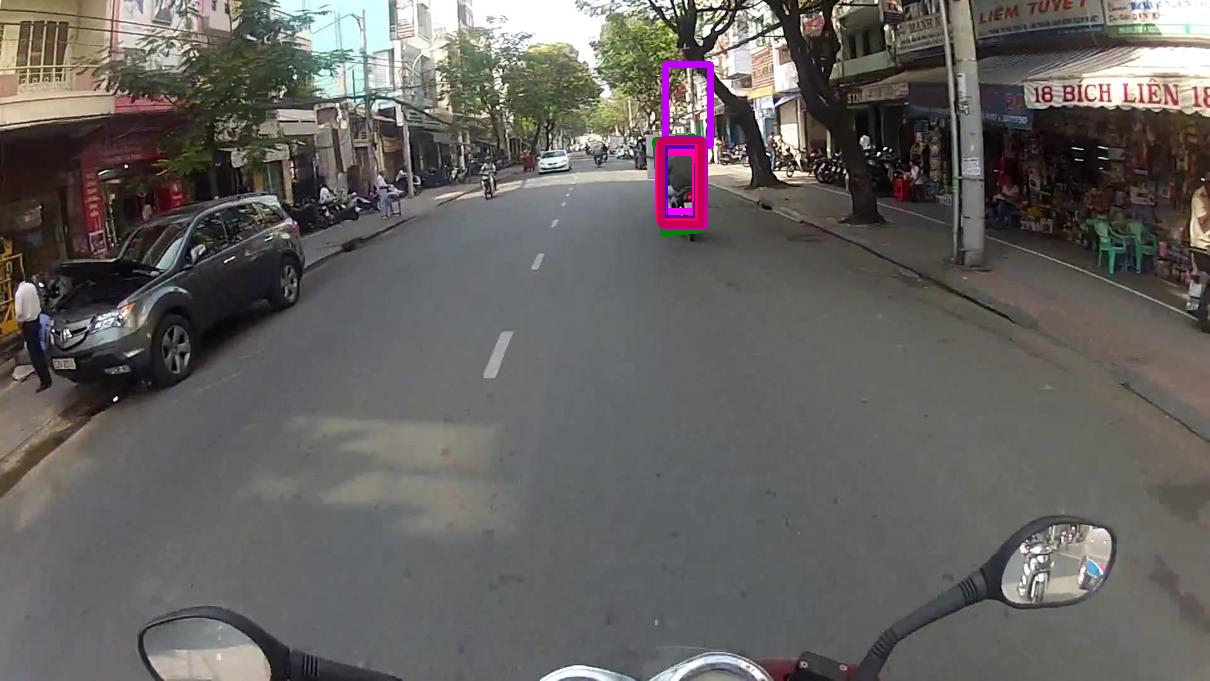, width=0.23\textwidth,height=0.2\textwidth}}}
\vspace{-0.03\textwidth}
\centerline{\hspace{0.03\textwidth} Traffic sequence}
\vspace{0.001\textwidth}
\centerline{ 
\tcbox[sharp corners, size = tight, boxrule=0.2mm, colframe=black, colback=white]{
\psfig{figure=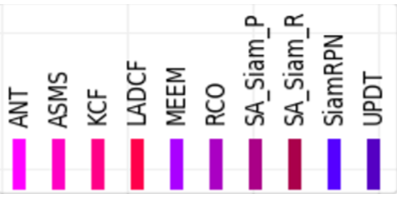, width=0.4\textwidth,height=0.15\textwidth}}}

\caption{Illustrates the tracking results of ten state-of-the-art trackers on Soccer1, and Traffic sequences. Each colored rectangle corresponds to each tracker's output. \label{figure_results4}}
\end{figure*}

Figures~\ref{figure_results1},~\ref{figure_results2},~\ref{figure_results3}, and~\ref{figure_results4} show the tracking results (selected few frames) of {\sc vot-2018} dataset obtained using the considered baselines --- {\sc ant}, {\sc asms}, {\sc kcf}, {\sc ladcf}, {\sc meem}, {\sc rco}, {\sc sa-s}iam-{\sc p}, {\sc sa-s}iam-{\sc r}, {\sc s}iam{\sc rpn}, and {\sc updt}. Object in Bag, Bmx, Book, Butterfly, Dinosaur, Fernando, Fish2, Gymnastics2, Hand, Iceskater2, and Motocross2 changes scale and orientation. Basketball, Fish2, Girl, and Soccer1 sequences contain occlusions. Matrix and Shaking sequences under go illumination variation. Object changes its scale in Blanket, Crabs1, and Helicopter sequences. For scaled and oriented object in the sequences such as Butterfly, Dinosaur, Fernando, Fish2, Gymnastics2, Hand, Iceskater2, and Motocross2, the trackers --- {\sc sa-s}iam-{\sc p}, {\sc sa-s}iam-{\sc r}, {\sc s}iam{\sc rpn}, and {\sc updt} performs better than other trackers. {\sc s}iam{\sc rpn}, {\sc ant}, {\sc rco}, and {\sc sa-s}iam-{\sc r} obtain robust tracking results on Basketball, Girl, Soccer1, and Fish2 sequences, respectively. 

\subsection{Analysis of Precision and Success Plots} \label{analysis_plots}

The success plot shows the ratio of successful frames at threshold varied from 0 to 1. Sometimes, the success plots and the precision plots make a different conclusion because they measure the tracker's other characteristics based on various metrics. However, {\sc auc} score of the success plot interprets the overall performance, which is more accurate than the center location error at the precision threshold value plot~\cite{benchmark_wu2013}. We analyze the tracker performance using {\sc auc} score of success plot, precision plot, and new success plot. The tracker has a maximum {\sc auc} score of the success plot indicates the best tracker. Since {\sc vot-2018} dataset contains 60 video sequences, it is challenging to analyze the state-of-the-art trackers' performances on every sequence. For this, we randomly select few (20) video sequences to analyze the existing trackers' performances based on {\sc auc} of success, precision, and new success plots. It is also challenging to analyze the trackers' average performances over {\sc vot-2018} dataset, using three plots to correspond to each video sequence of this dataset. For this reason, we use average success, precision, and new success plots over all sequences of this dataset. We analyze the performances of the trackers based on average {\sc auc} corresponds to these three plots. 

\begin{figure*}[htp!]
\centerline{Precision Plot \hspace{0.23\textwidth} Success Plot \hspace{0.21\textwidth} New Success Plot}
\centerline{ 
\tcbox[sharp corners, size = tight, boxrule=0.2mm, colframe=black, colback=white]{
\psfig{figure=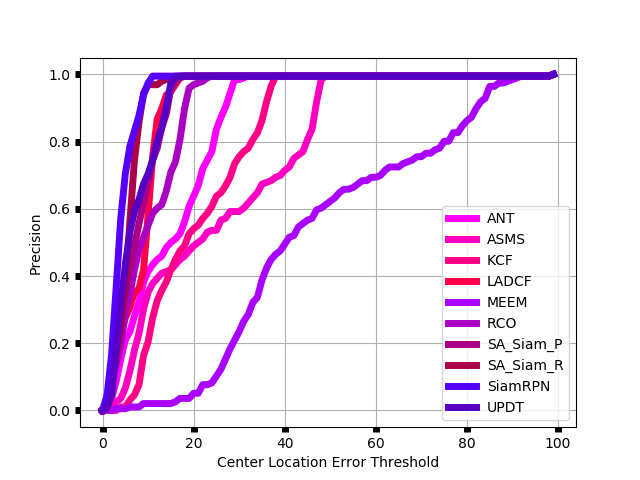, width=0.32\textwidth,height=0.2\textwidth}}
\hspace{0.001\textwidth}
\tcbox[sharp corners, size = tight, boxrule=0.2mm, colframe=black, colback=white]{
\psfig{figure=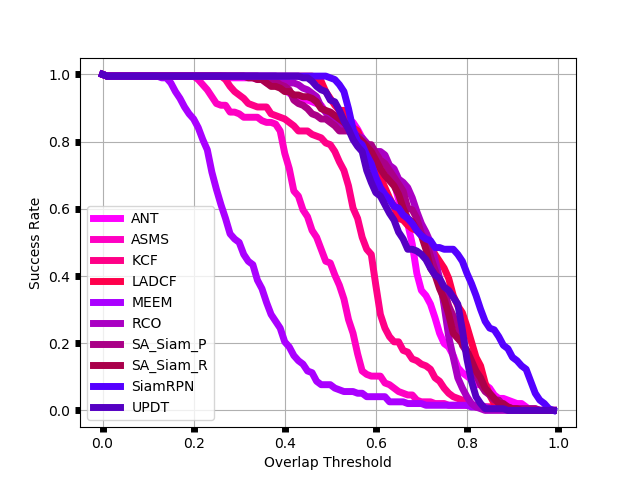, width=0.32\textwidth,height=0.2\textwidth}}
\hspace{0.001\textwidth}
\tcbox[sharp corners, size = tight, boxrule=0.2mm, colframe=black, colback=white]{
\psfig{figure=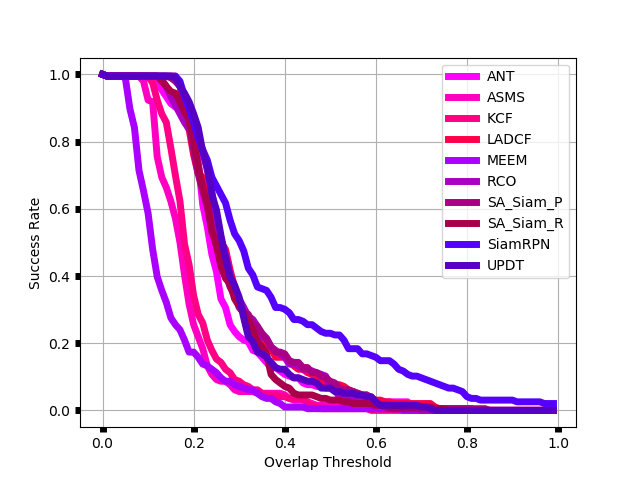, width=0.32\textwidth,height=0.2\textwidth}}}
\centerline{\hspace{0.03\textwidth} Bag sequence}
\vspace{0.001\textwidth}
\centerline{ 
\tcbox[sharp corners, size = tight, boxrule=0.2mm, colframe=black, colback=white]{
\psfig{figure=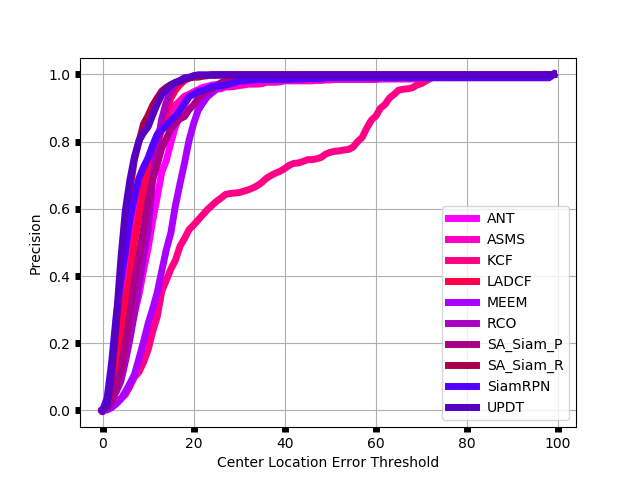, width=0.32\textwidth,height=0.2\textwidth}}
\hspace{0.001\textwidth}
\tcbox[sharp corners, size = tight, boxrule=0.2mm, colframe=black, colback=white]{
\psfig{figure=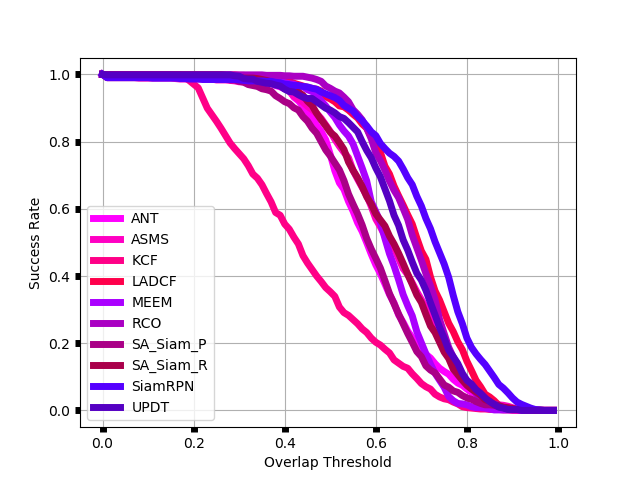, width=0.32\textwidth,height=0.2\textwidth}}
\hspace{0.001\textwidth}
\tcbox[sharp corners, size = tight, boxrule=0.2mm, colframe=black, colback=white]{
\psfig{figure=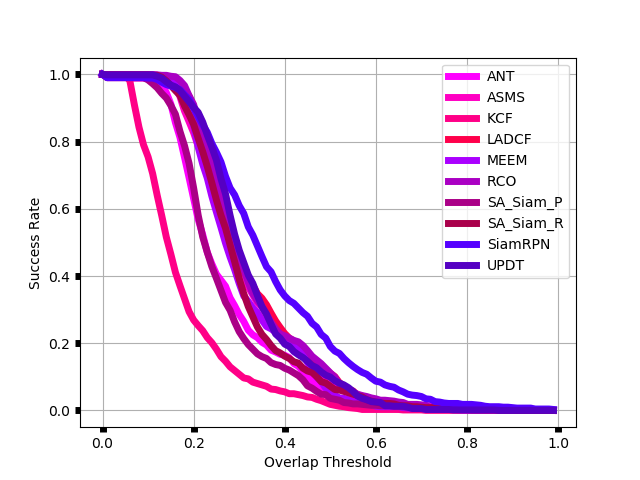, width=0.32\textwidth,height=0.2\textwidth}}}
\centerline{\hspace{0.03\textwidth} Basketball sequence}
\vspace{0.001\textwidth}
\centerline{ 
\tcbox[sharp corners, size = tight, boxrule=0.2mm, colframe=black, colback=white]{
\psfig{figure=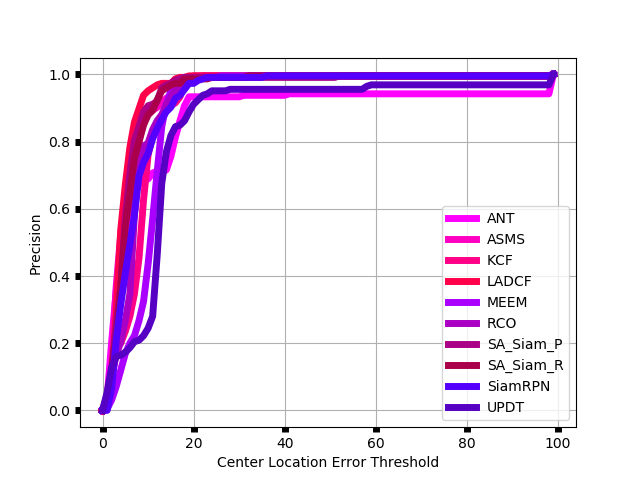, width=0.32\textwidth,height=0.2\textwidth}}
\hspace{0.001\textwidth}
\tcbox[sharp corners, size = tight, boxrule=0.2mm, colframe=black, colback=white]{
\psfig{figure=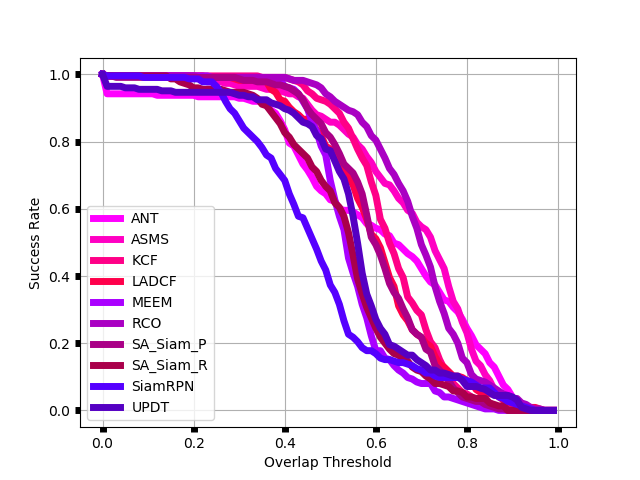, width=0.32\textwidth,height=0.2\textwidth}}
\hspace{0.001\textwidth}
\tcbox[sharp corners, size = tight, boxrule=0.2mm, colframe=black, colback=white]{
\psfig{figure=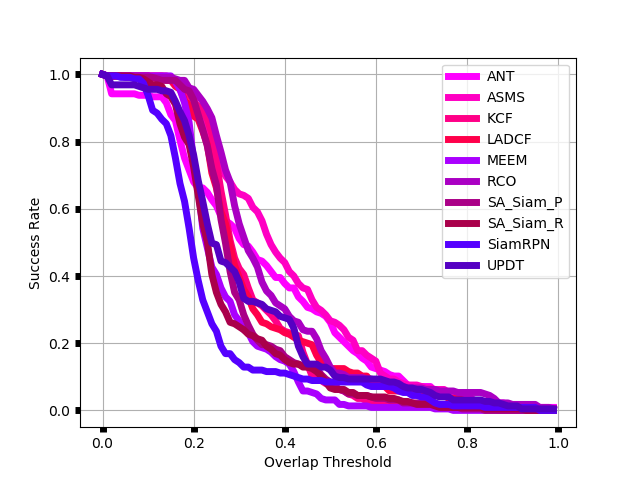, width=0.32\textwidth,height=0.2\textwidth}}}
\centerline{\hspace{0.03\textwidth} Blanket sequence}
\vspace{0.001\textwidth}
\centerline{ 
\tcbox[sharp corners, size = tight, boxrule=0.2mm, colframe=black, colback=white]{
\psfig{figure=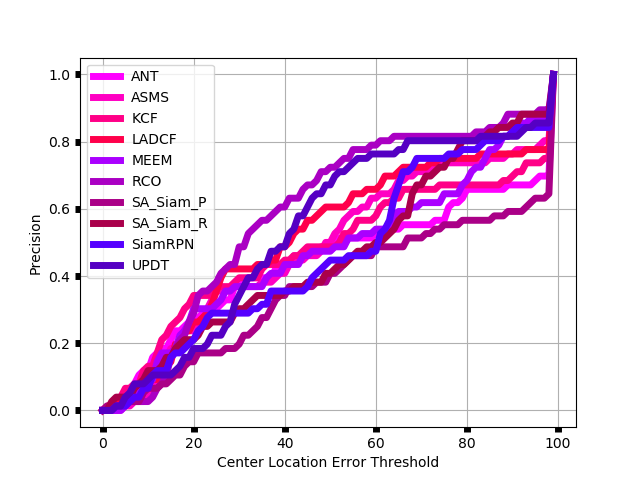, width=0.32\textwidth,height=0.2\textwidth}}
\hspace{0.001\textwidth}
\tcbox[sharp corners, size = tight, boxrule=0.2mm, colframe=black, colback=white]{
\psfig{figure=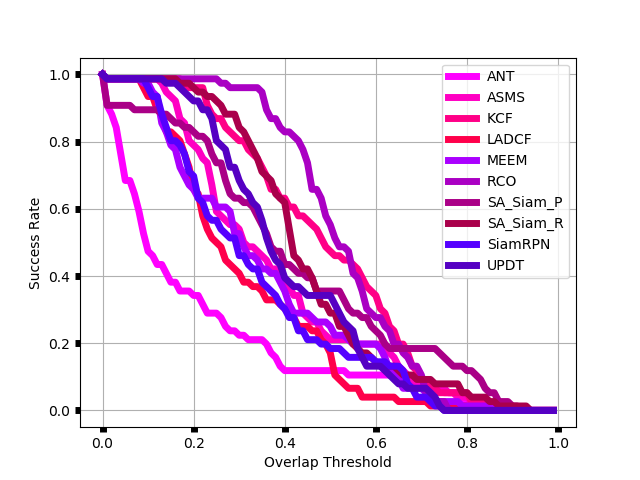, width=0.32\textwidth,height=0.2\textwidth}}
\hspace{0.001\textwidth}
\tcbox[sharp corners, size = tight, boxrule=0.2mm, colframe=black, colback=white]{
\psfig{figure=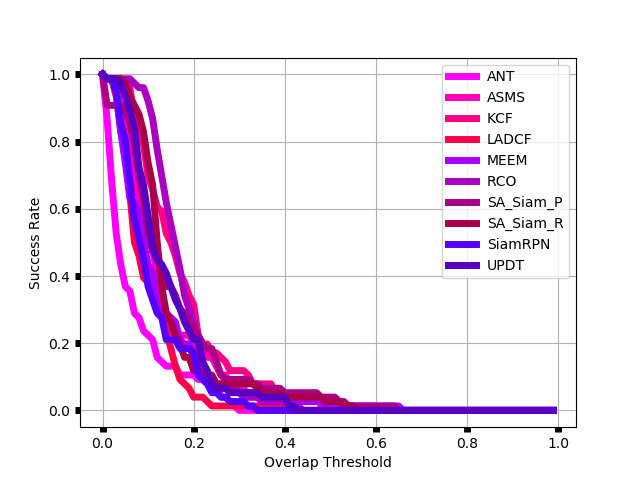, width=0.32\textwidth,height=0.2\textwidth}}}
\centerline{\hspace{0.03\textwidth} Bmx sequence}
\vspace{0.001\textwidth}
\centerline{ 
\tcbox[sharp corners, size = tight, boxrule=0.2mm, colframe=black, colback=white]{
\psfig{figure=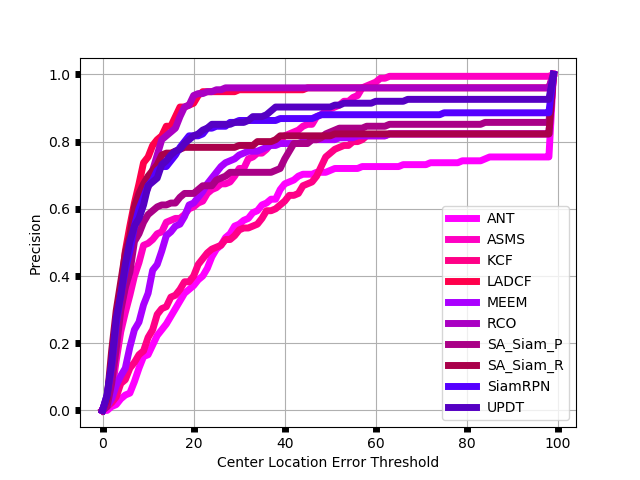, width=0.32\textwidth,height=0.2\textwidth}}
\hspace{0.001\textwidth}
\tcbox[sharp corners, size = tight, boxrule=0.2mm, colframe=black, colback=white]{
\psfig{figure=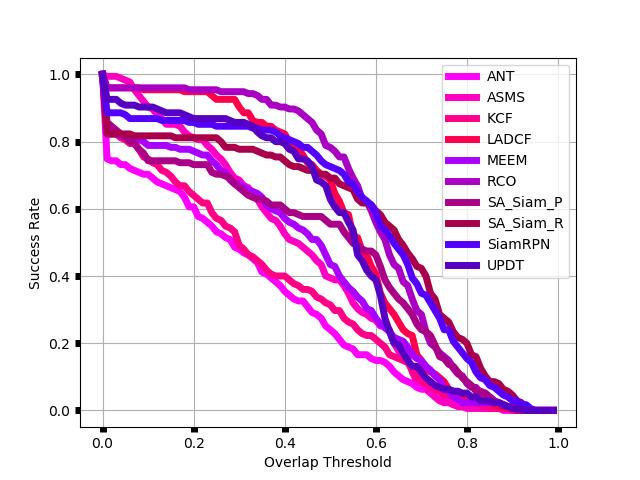, width=0.32\textwidth,height=0.2\textwidth}}
\hspace{0.001\textwidth}
\tcbox[sharp corners, size = tight, boxrule=0.2mm, colframe=black, colback=white]{
\psfig{figure=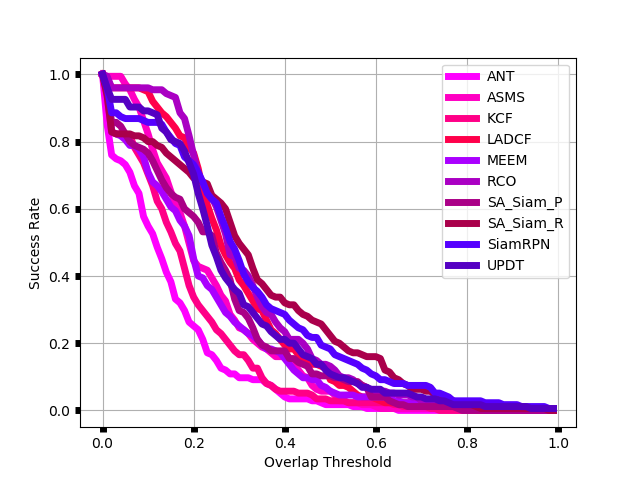, width=0.32\textwidth,height=0.2\textwidth}}}
\centerline{\hspace{0.03\textwidth} Book sequence}
\vspace{0.001\textwidth}
\centerline{ 
\tcbox[sharp corners, size = tight, boxrule=0.2mm, colframe=black, colback=white]{
\psfig{figure=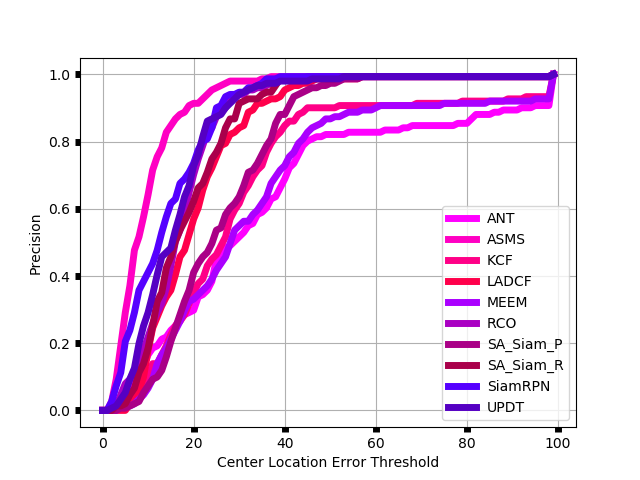, width=0.32\textwidth,height=0.2\textwidth}}
\hspace{0.001\textwidth}
\tcbox[sharp corners, size = tight, boxrule=0.2mm, colframe=black, colback=white]{
\psfig{figure=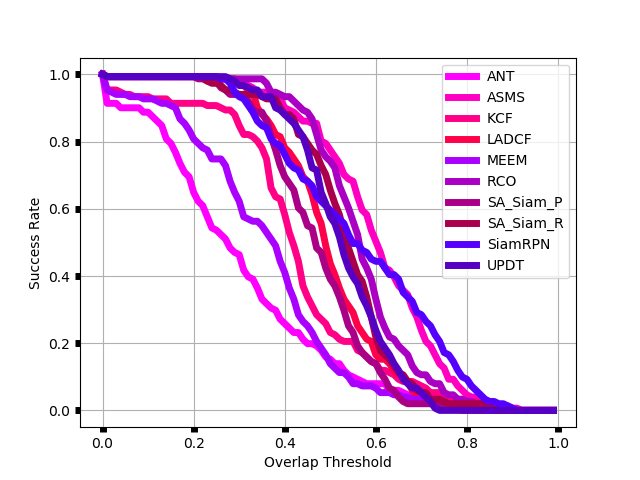, width=0.32\textwidth,height=0.2\textwidth}}
\hspace{0.001\textwidth}
\tcbox[sharp corners, size = tight, boxrule=0.2mm, colframe=black, colback=white]{
\psfig{figure=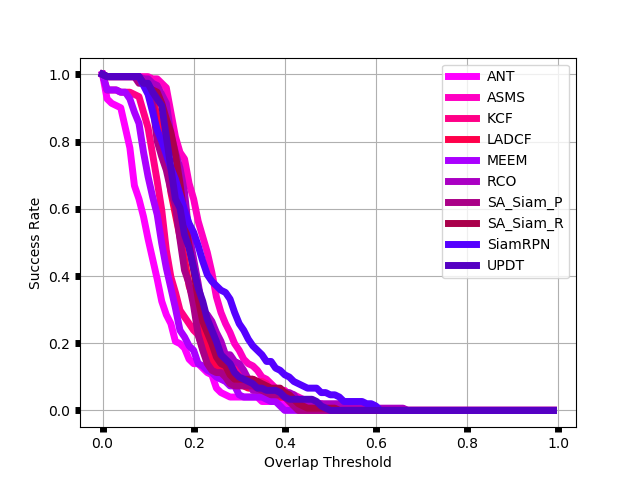, width=0.32\textwidth,height=0.2\textwidth}}}
\centerline{\hspace{0.03\textwidth} Butterfly sequence}
\caption{Illustrates the precision and success plots popularly used for evaluating a tracking algorithm and the success plot based on the proposed new measure - matching score of ten different methods on Bag, Basketball, Blanket, Bmx, Book, and Butterfly sequences. Each plot corresponds to one-pass evaluation ({\sc ope}). \label{figure_plot1}}
\end{figure*}
\begin{figure*}[htp!]
\centerline{Precision Plot \hspace{0.23\textwidth} Success Plot \hspace{0.21\textwidth} New Success Plot}
\centerline{ 
\tcbox[sharp corners, size = tight, boxrule=0.2mm, colframe=black, colback=white]{
\psfig{figure=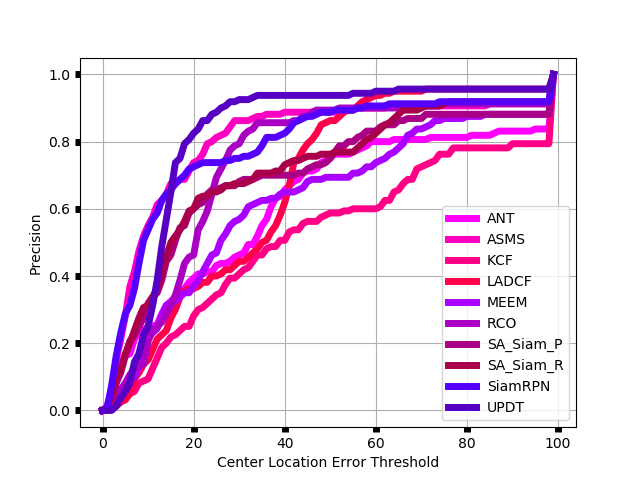, width=0.32\textwidth,height=0.2\textwidth}}
\hspace{0.001\textwidth}
\tcbox[sharp corners, size = tight, boxrule=0.2mm, colframe=black, colback=white]{
\psfig{figure=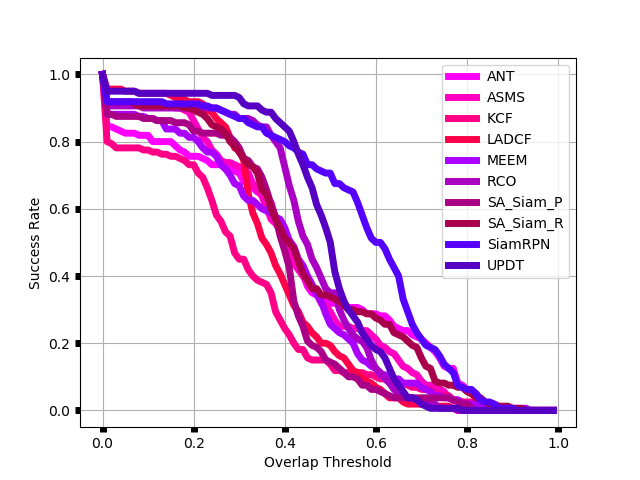, width=0.32\textwidth,height=0.2\textwidth}}
\hspace{0.001\textwidth}
\tcbox[sharp corners, size = tight, boxrule=0.2mm, colframe=black, colback=white]{
\psfig{figure=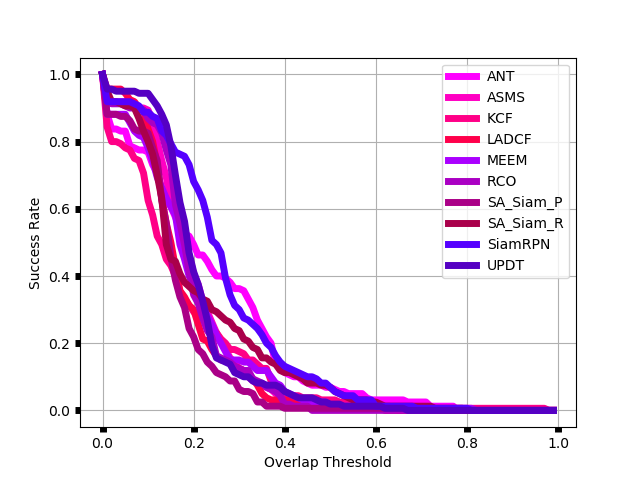, width=0.32\textwidth,height=0.2\textwidth}}}
\centerline{\hspace{0.03\textwidth} Crabs1 sequence}
\vspace{0.001\textwidth}
\centerline{ 
\tcbox[sharp corners, size = tight, boxrule=0.2mm, colframe=black, colback=white]{
\psfig{figure=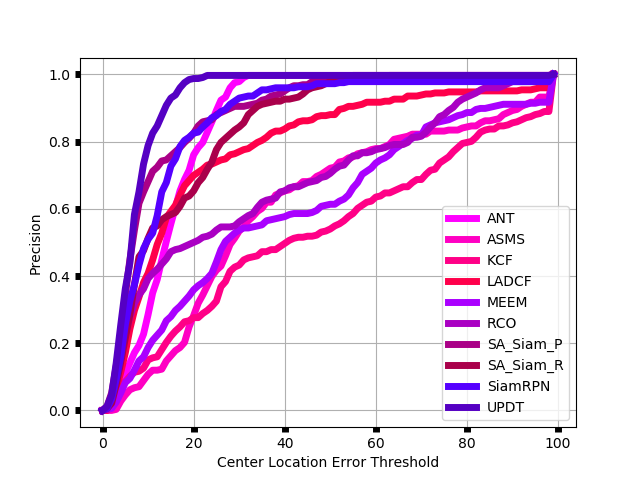, width=0.32\textwidth,height=0.2\textwidth}}
\hspace{0.001\textwidth}
\tcbox[sharp corners, size = tight, boxrule=0.2mm, colframe=black, colback=white]{
\psfig{figure=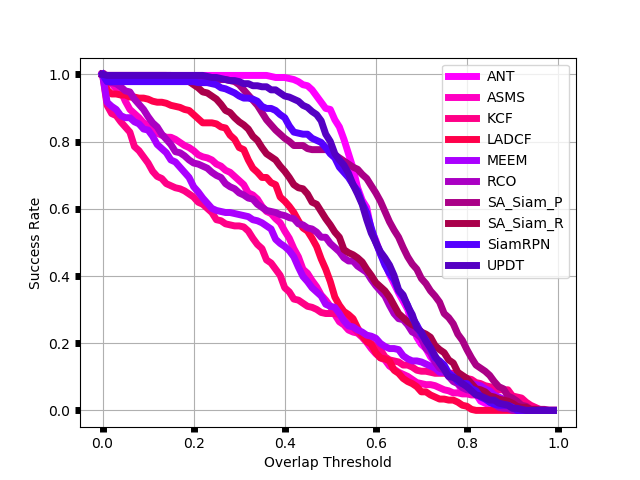, width=0.32\textwidth,height=0.2\textwidth}}
\hspace{0.001\textwidth}
\tcbox[sharp corners, size = tight, boxrule=0.2mm, colframe=black, colback=white]{
\psfig{figure=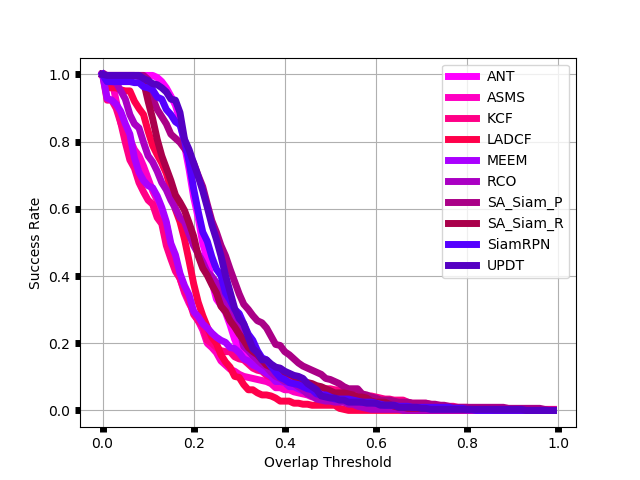, width=0.32\textwidth,height=0.2\textwidth}}}
\centerline{\hspace{0.03\textwidth} Dinosaur sequence}
\vspace{0.001\textwidth}
\centerline{ 
\tcbox[sharp corners, size = tight, boxrule=0.2mm, colframe=black, colback=white]{
\psfig{figure=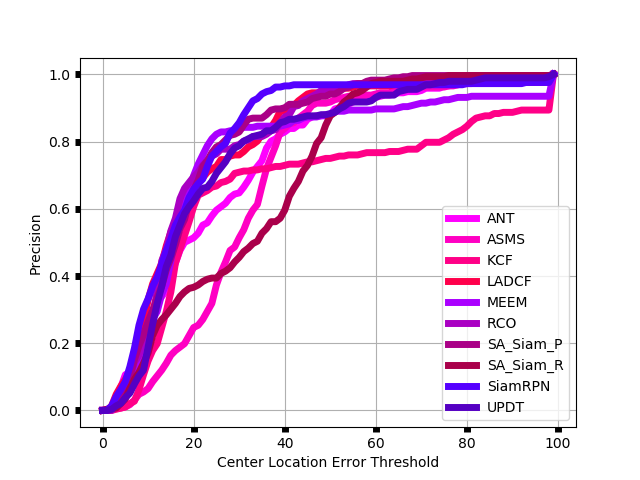, width=0.32\textwidth,height=0.2\textwidth}}
\hspace{0.001\textwidth}
\tcbox[sharp corners, size = tight, boxrule=0.2mm, colframe=black, colback=white]{
\psfig{figure=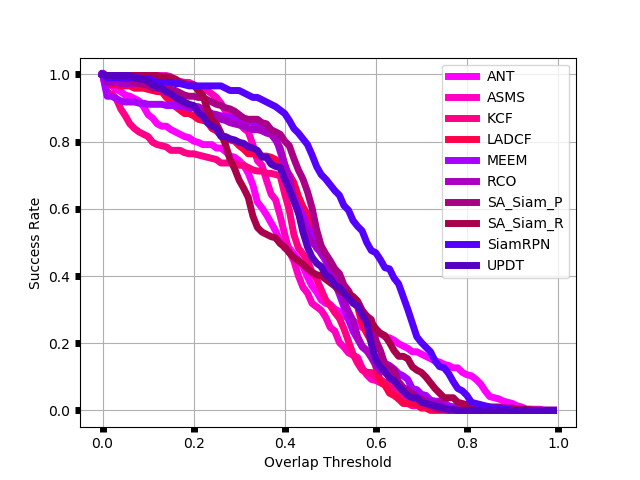, width=0.32\textwidth,height=0.2\textwidth}}
\hspace{0.001\textwidth}
\tcbox[sharp corners, size = tight, boxrule=0.2mm, colframe=black, colback=white]{
\psfig{figure=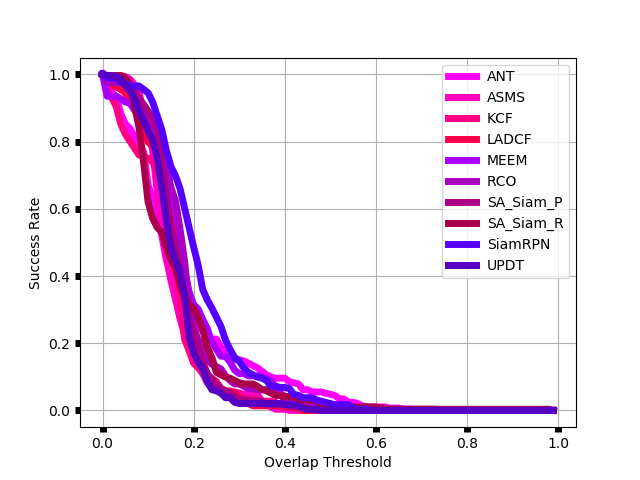, width=0.32\textwidth,height=0.2\textwidth}}}
\centerline{\hspace{0.03\textwidth} Fernando sequence}
\vspace{0.001\textwidth}
\centerline{ 
\tcbox[sharp corners, size = tight, boxrule=0.2mm, colframe=black, colback=white]{
\psfig{figure=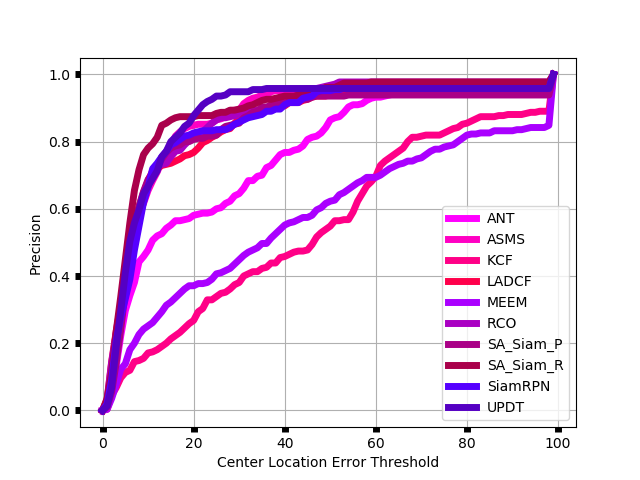, width=0.32\textwidth,height=0.2\textwidth}}
\hspace{0.001\textwidth}
\tcbox[sharp corners, size = tight, boxrule=0.2mm, colframe=black, colback=white]{
\psfig{figure=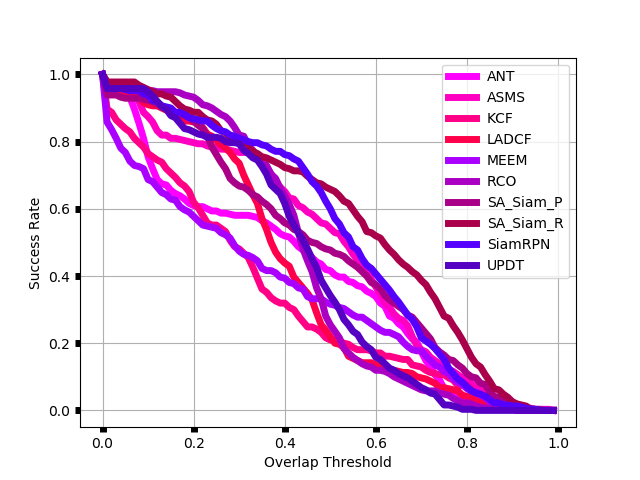, width=0.32\textwidth,height=0.2\textwidth}}
\hspace{0.001\textwidth}
\tcbox[sharp corners, size = tight, boxrule=0.2mm, colframe=black, colback=white]{
\psfig{figure=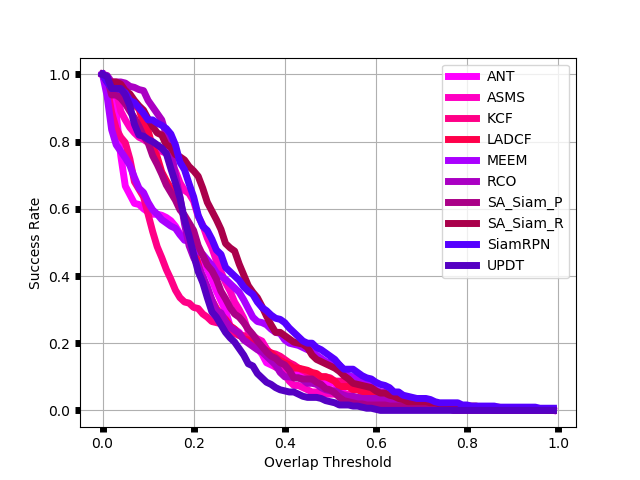, width=0.32\textwidth,height=0.2\textwidth}}}
\centerline{\hspace{0.03\textwidth} Fish2 sequence}
\vspace{0.001\textwidth}
\centerline{ 
\tcbox[sharp corners, size = tight, boxrule=0.2mm, colframe=black, colback=white]{
\psfig{figure=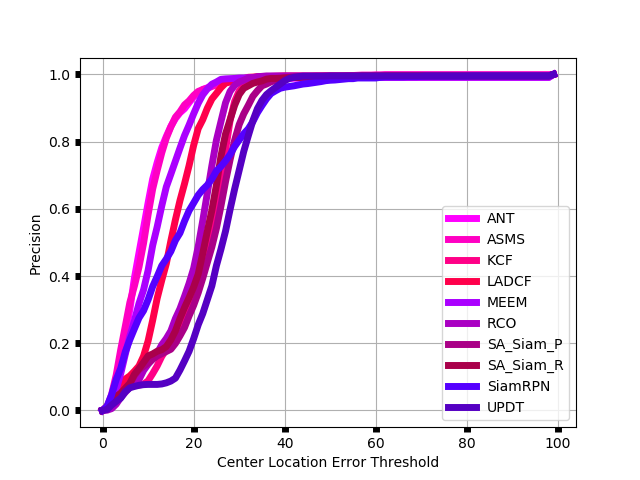, width=0.32\textwidth,height=0.2\textwidth}}
\hspace{0.001\textwidth}
\tcbox[sharp corners, size = tight, boxrule=0.2mm, colframe=black, colback=white]{
\psfig{figure=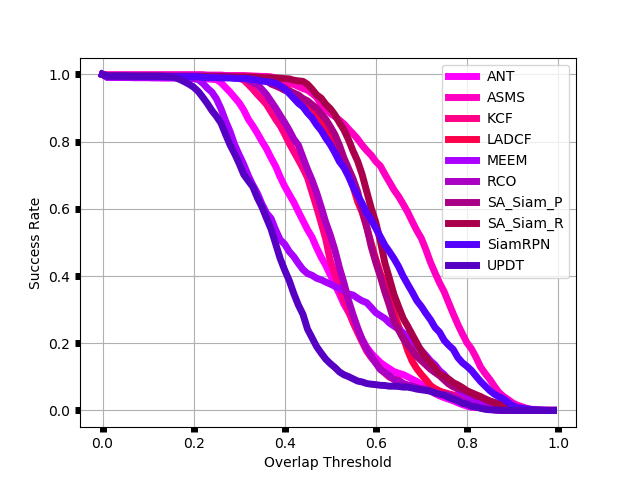, width=0.32\textwidth,height=0.2\textwidth}}
\hspace{0.001\textwidth}
\tcbox[sharp corners, size = tight, boxrule=0.2mm, colframe=black, colback=white]{
\psfig{figure=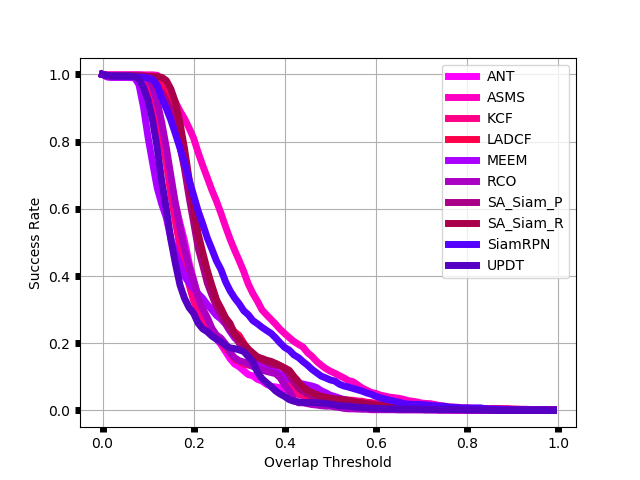, width=0.32\textwidth,height=0.2\textwidth}}}
\centerline{\hspace{0.03\textwidth} Girl sequence}
\vspace{0.001\textwidth}
\centerline{ 
\tcbox[sharp corners, size = tight, boxrule=0.2mm, colframe=black, colback=white]{
\psfig{figure=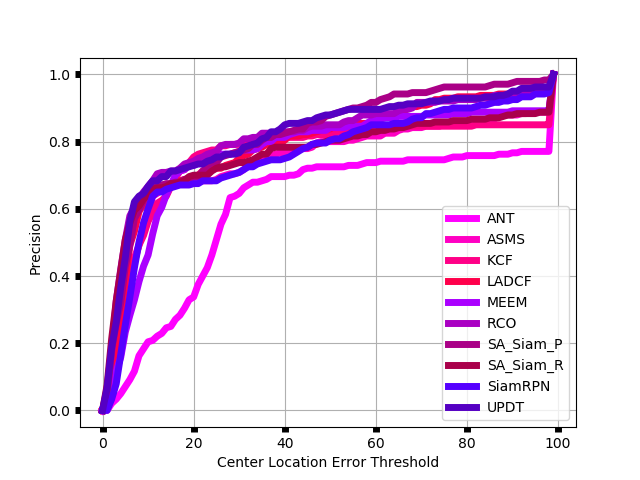, width=0.32\textwidth,height=0.2\textwidth}}
\hspace{0.001\textwidth}
\tcbox[sharp corners, size = tight, boxrule=0.2mm, colframe=black, colback=white]{
\psfig{figure=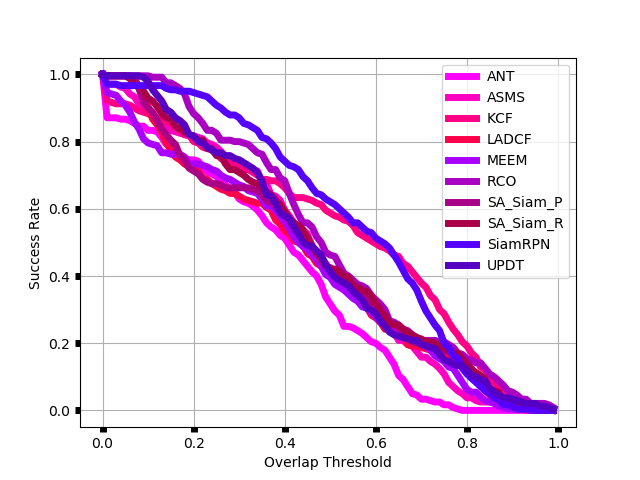, width=0.32\textwidth,height=0.2\textwidth}}
\hspace{0.001\textwidth}
\tcbox[sharp corners, size = tight, boxrule=0.2mm, colframe=black, colback=white]{
\psfig{figure=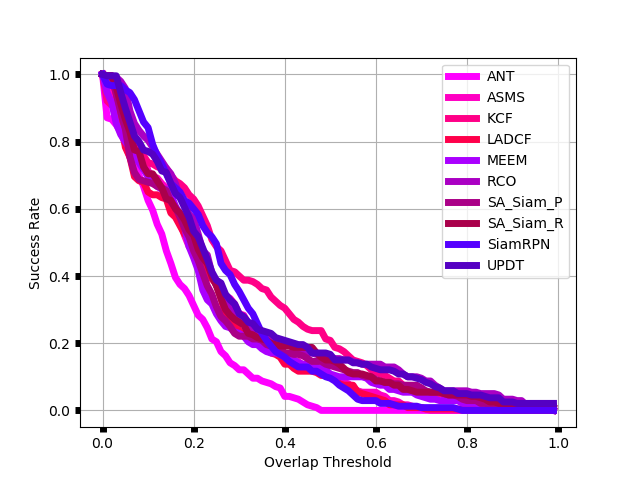, width=0.32\textwidth,height=0.2\textwidth}}}
\centerline{\hspace{0.03\textwidth} Gymnastics2 sequence}
\caption{Illustrates the precision and success plots popularly used for evaluating a tracking algorithm and the success plot based on the proposed new measure - matching score of ten different methods on Crabs1, Dinosaur, Fernando, Fish2, Girl, and Gymnastics2 sequences. Each plot corresponds to one-pass evaluation ({\sc ope}). \label{figure_plot2}}
\end{figure*}
\begin{figure*}[htp!]
\centerline{Precision Plot \hspace{0.23\textwidth} Success Plot \hspace{0.21\textwidth} New Success Plot}
\centerline{ 
\tcbox[sharp corners, size = tight, boxrule=0.2mm, colframe=black, colback=white]{
\psfig{figure=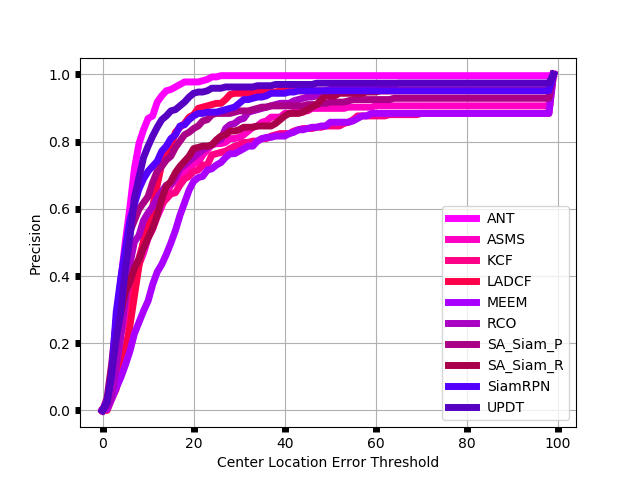, width=0.32\textwidth,height=0.2\textwidth}}
\hspace{0.001\textwidth}
\tcbox[sharp corners, size = tight, boxrule=0.2mm, colframe=black, colback=white]{
\psfig{figure=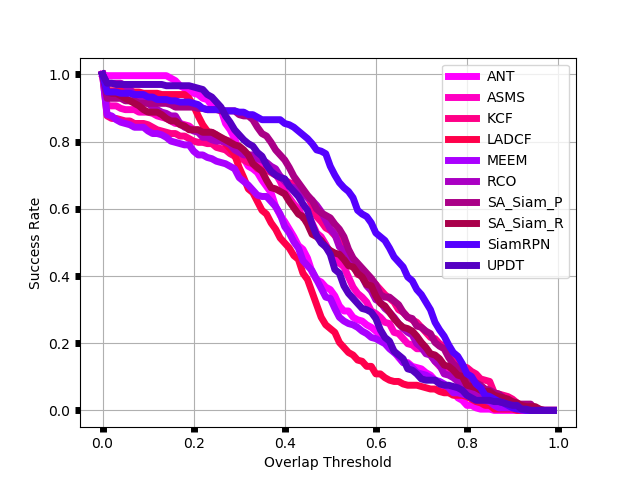, width=0.32\textwidth,height=0.2\textwidth}}
\hspace{0.001\textwidth}
\tcbox[sharp corners, size = tight, boxrule=0.2mm, colframe=black, colback=white]{
\psfig{figure=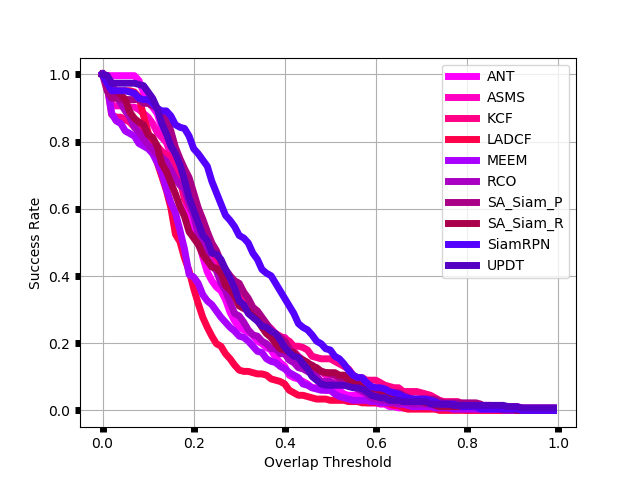, width=0.32\textwidth,height=0.2\textwidth}}}
\centerline{\hspace{0.03\textwidth} Hand sequence}
\vspace{0.001\textwidth}
\centerline{ 
\tcbox[sharp corners, size = tight, boxrule=0.2mm, colframe=black, colback=white]{
\psfig{figure=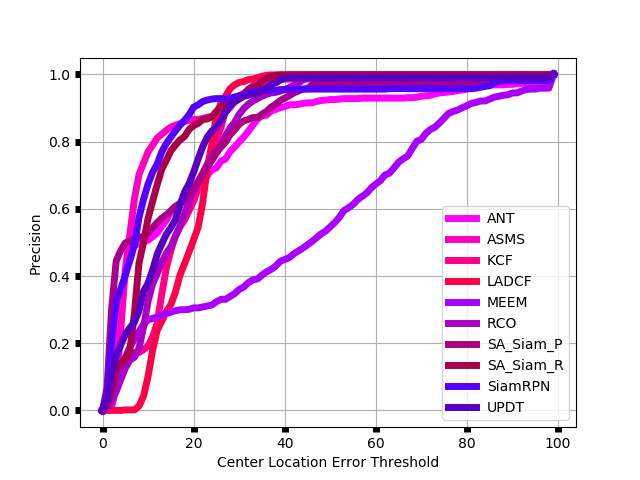, width=0.32\textwidth,height=0.2\textwidth}}
\hspace{0.001\textwidth}
\tcbox[sharp corners, size = tight, boxrule=0.2mm, colframe=black, colback=white]{
\psfig{figure=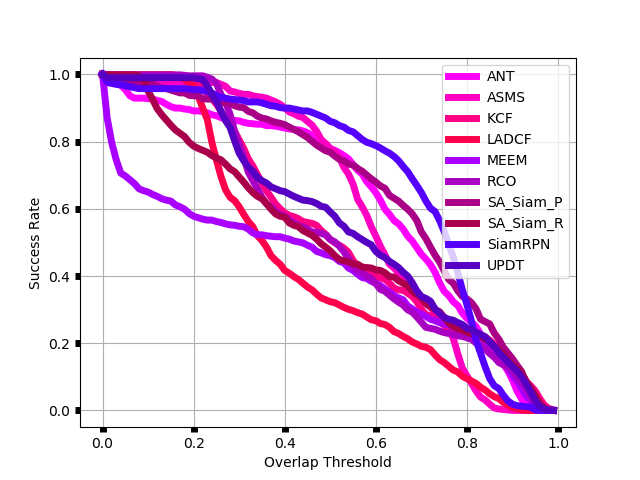, width=0.32\textwidth,height=0.2\textwidth}}
\hspace{0.001\textwidth}
\tcbox[sharp corners, size = tight, boxrule=0.2mm, colframe=black, colback=white]{
\psfig{figure=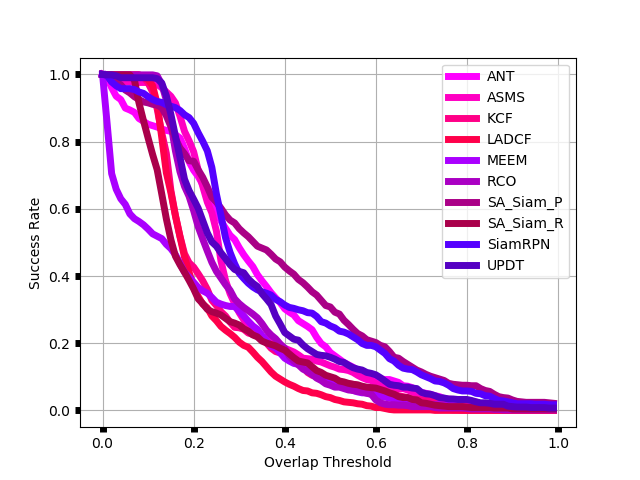, width=0.32\textwidth,height=0.2\textwidth}}}
\centerline{\hspace{0.03\textwidth} Helicopter sequence}
\vspace{0.001\textwidth}
\centerline{ 
\tcbox[sharp corners, size = tight, boxrule=0.2mm, colframe=black, colback=white]{
\psfig{figure=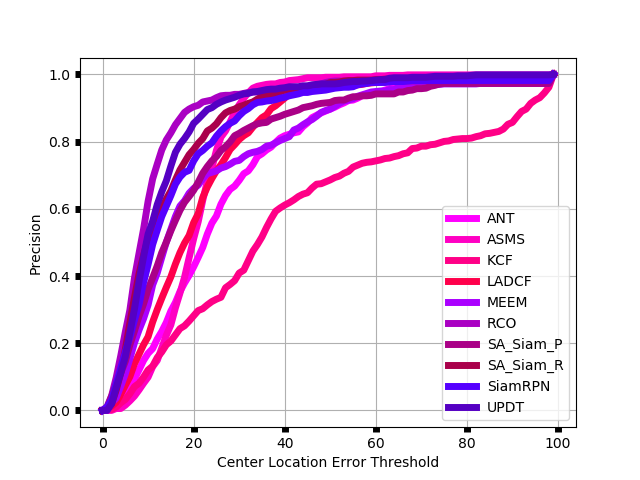, width=0.32\textwidth,height=0.2\textwidth}}
\hspace{0.001\textwidth}
\tcbox[sharp corners, size = tight, boxrule=0.2mm, colframe=black, colback=white]{
\psfig{figure=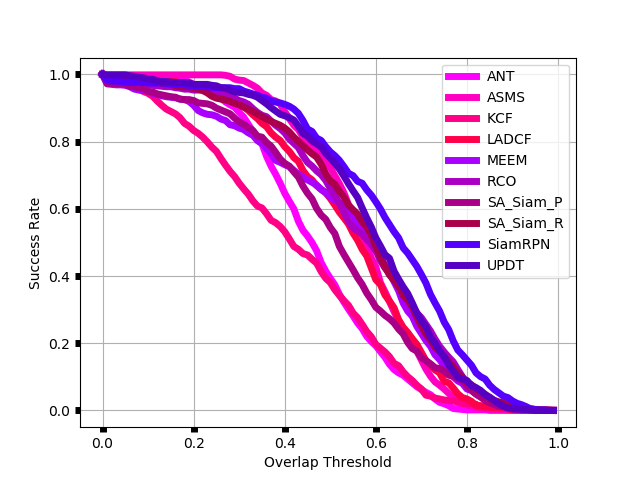, width=0.32\textwidth,height=0.2\textwidth}}
\hspace{0.001\textwidth}
\tcbox[sharp corners, size = tight, boxrule=0.2mm, colframe=black, colback=white]{
\psfig{figure=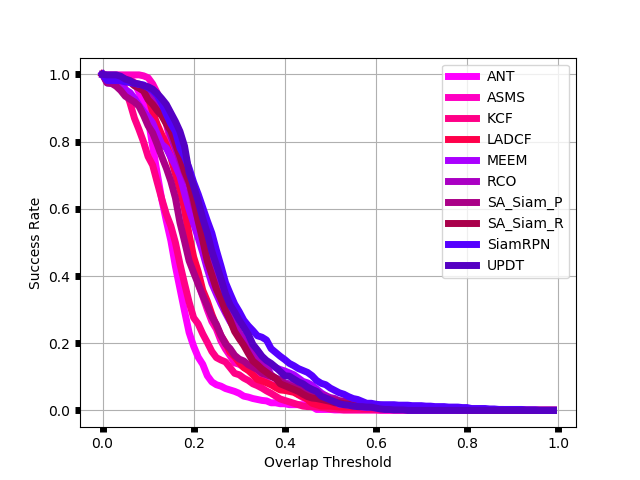, width=0.32\textwidth,height=0.2\textwidth}}}
\centerline{\hspace{0.03\textwidth} Iceskater2 sequence}
\vspace{0.001\textwidth}
\centerline{ 
\tcbox[sharp corners, size = tight, boxrule=0.2mm, colframe=black, colback=white]{
\psfig{figure=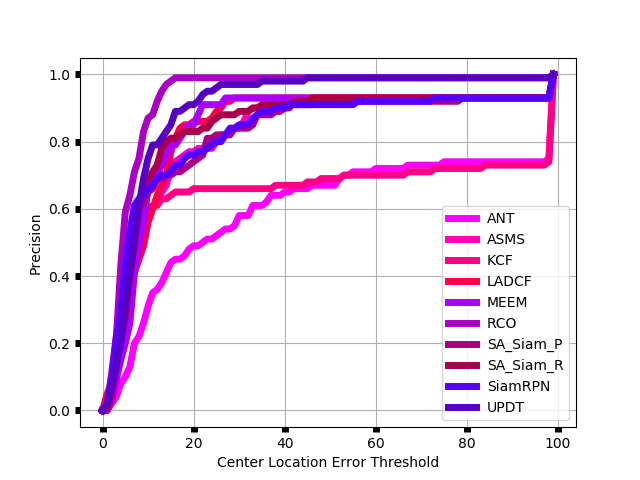, width=0.32\textwidth,height=0.2\textwidth}}
\hspace{0.001\textwidth}
\tcbox[sharp corners, size = tight, boxrule=0.2mm, colframe=black, colback=white]{
\psfig{figure=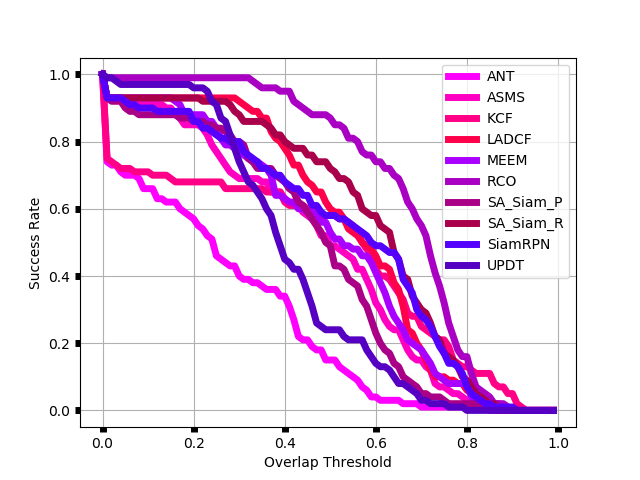, width=0.32\textwidth,height=0.2\textwidth}}
\hspace{0.001\textwidth}
\tcbox[sharp corners, size = tight, boxrule=0.2mm, colframe=black, colback=white]{
\psfig{figure=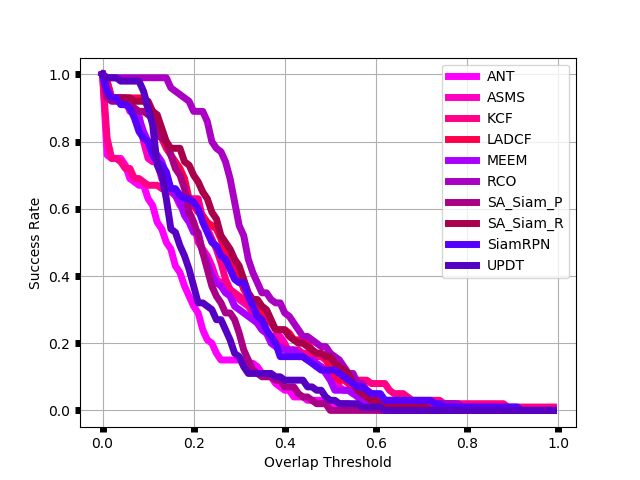, width=0.32\textwidth,height=0.2\textwidth}}}
\centerline{\hspace{0.03\textwidth} Matrix sequence}
\vspace{0.001\textwidth}
\centerline{ 
\tcbox[sharp corners, size = tight, boxrule=0.2mm, colframe=black, colback=white]{
\psfig{figure=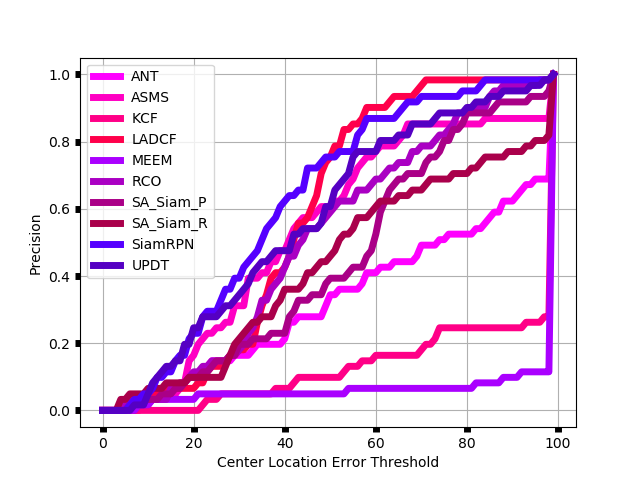, width=0.32\textwidth,height=0.2\textwidth}}
\hspace{0.001\textwidth}
\tcbox[sharp corners, size = tight, boxrule=0.2mm, colframe=black, colback=white]{
\psfig{figure=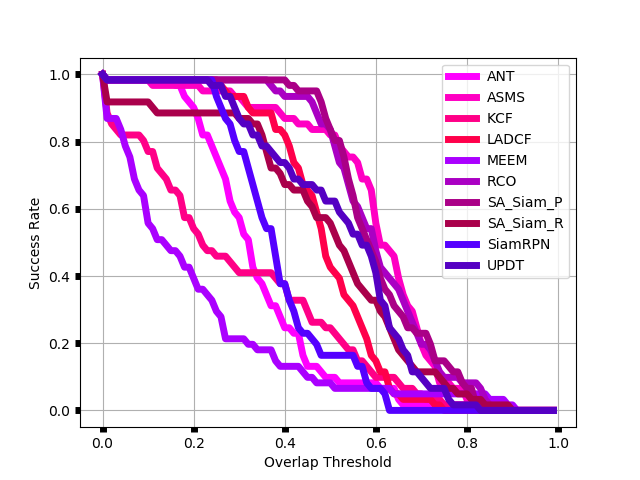, width=0.32\textwidth,height=0.2\textwidth}}
\hspace{0.001\textwidth}
\tcbox[sharp corners, size = tight, boxrule=0.2mm, colframe=black, colback=white]{
\psfig{figure=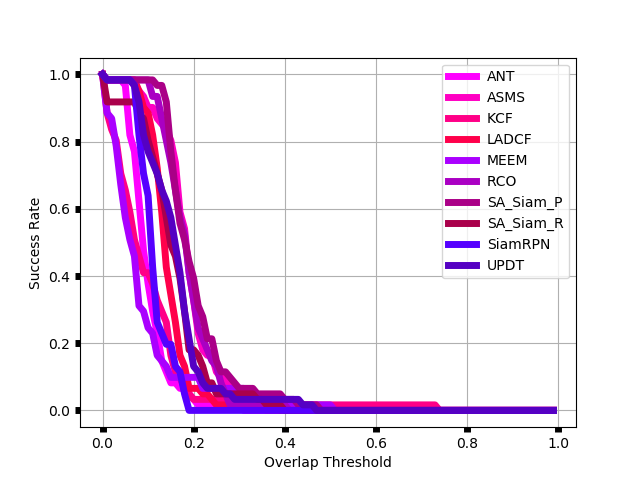, width=0.32\textwidth,height=0.2\textwidth}}}
\centerline{\hspace{0.03\textwidth} Motocross2 sequence}
\vspace{0.001\textwidth}
\centerline{ 
\tcbox[sharp corners, size = tight, boxrule=0.2mm, colframe=black, colback=white]{
\psfig{figure=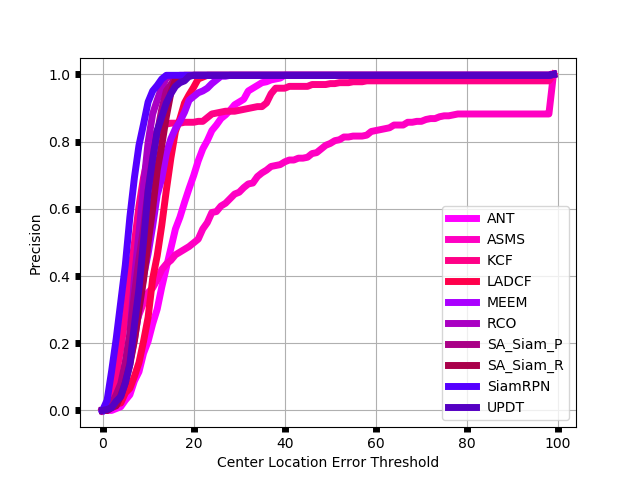, width=0.32\textwidth,height=0.2\textwidth}}
\hspace{0.001\textwidth}
\tcbox[sharp corners, size = tight, boxrule=0.2mm, colframe=black, colback=white]{
\psfig{figure=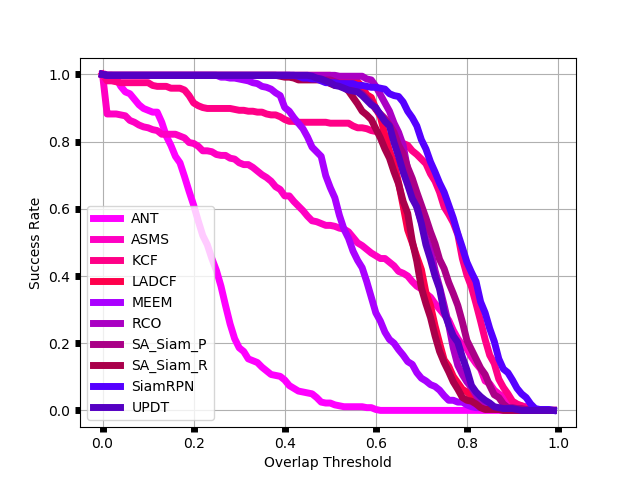, width=0.32\textwidth,height=0.2\textwidth}}
\hspace{0.001\textwidth}
\tcbox[sharp corners, size = tight, boxrule=0.2mm, colframe=black, colback=white]{
\psfig{figure=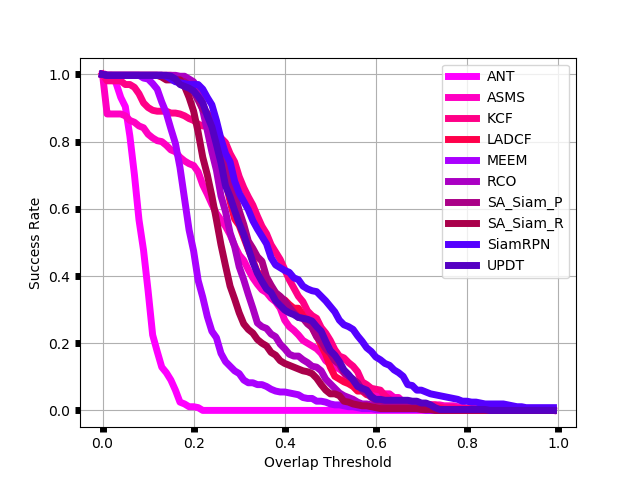, width=0.32\textwidth,height=0.2\textwidth}}}
\centerline{\hspace{0.03\textwidth} Shaking sequence}
\caption{Illustrates the precision and success plots popularly used for evaluating a tracking algorithm and the success plot based on the proposed new measure - matching score of ten different methods on Hand, Helicopter, Iceskater2, Matrix, Motocross2, and Shaking sequences. Each plot corresponds to one-pass evaluation ({\sc ope}). \label{figure_plot3}}
\end{figure*}
\begin{figure*}[htp!]
\centerline{Precision Plot \hspace{0.23\textwidth} Success Plot \hspace{0.21\textwidth} New Success Plot}
\centerline{ 
\tcbox[sharp corners, size = tight, boxrule=0.2mm, colframe=black, colback=white]{
\psfig{figure=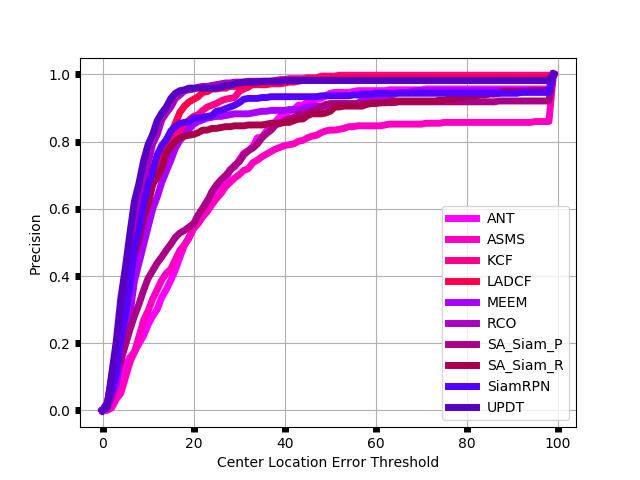, width=0.32\textwidth,height=0.2\textwidth}}
\hspace{0.001\textwidth}
\tcbox[sharp corners, size = tight, boxrule=0.2mm, colframe=black, colback=white]{
\psfig{figure=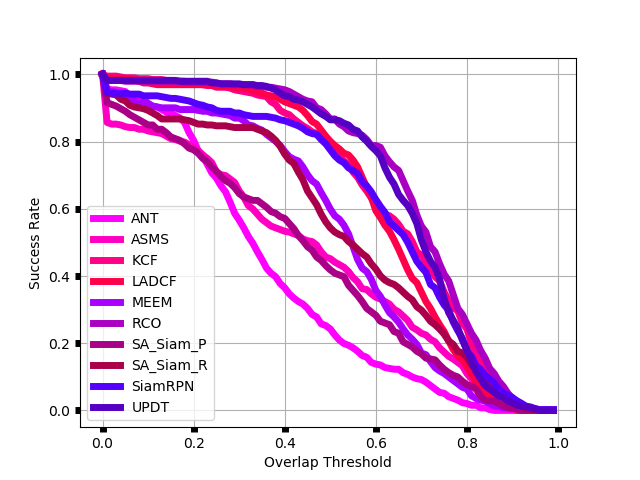, width=0.32\textwidth,height=0.2\textwidth}}
\hspace{0.001\textwidth}
\tcbox[sharp corners, size = tight, boxrule=0.2mm, colframe=black, colback=white]{
\psfig{figure=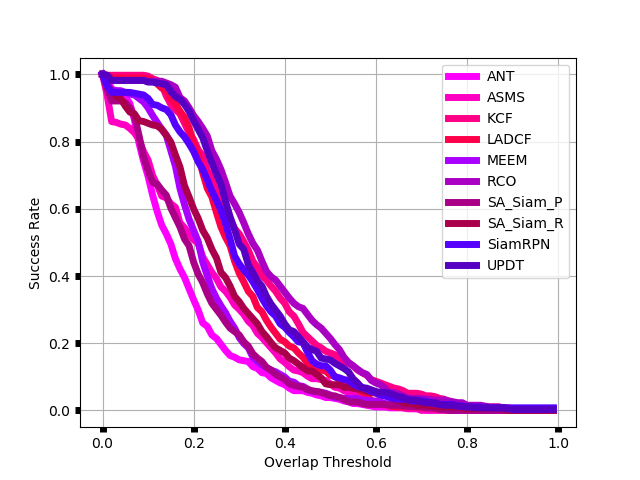, width=0.32\textwidth,height=0.2\textwidth}}}
\centerline{\hspace{0.03\textwidth} Soccer1 sequence}
\vspace{0.001\textwidth}
\centerline{ 
\tcbox[sharp corners, size = tight, boxrule=0.2mm, colframe=black, colback=white]{
\psfig{figure=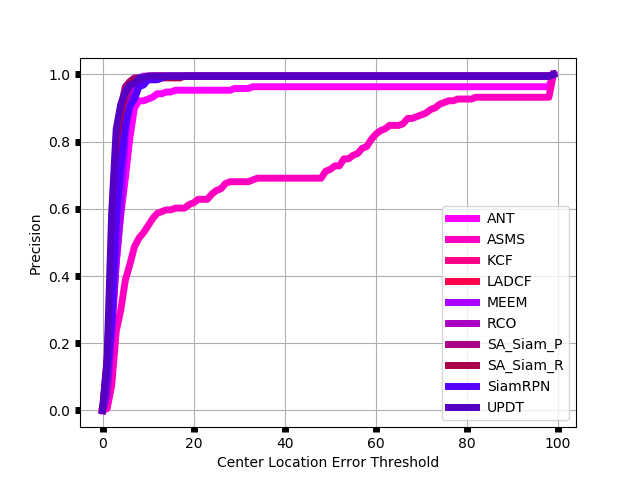, width=0.32\textwidth,height=0.2\textwidth}}
\hspace{0.001\textwidth}
\tcbox[sharp corners, size = tight, boxrule=0.2mm, colframe=black, colback=white]{
\psfig{figure=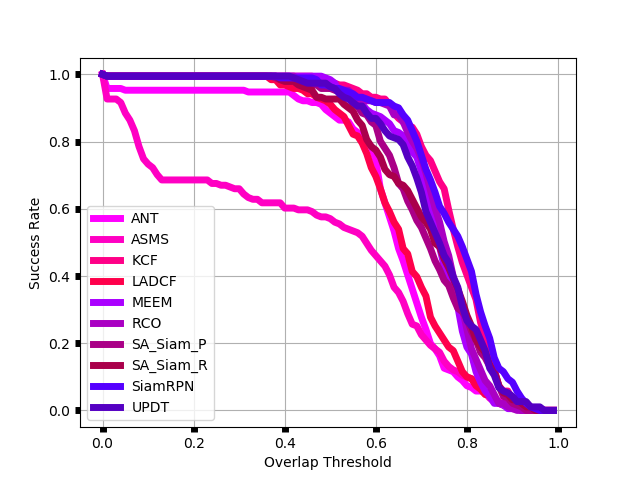, width=0.32\textwidth,height=0.2\textwidth}}
\hspace{0.001\textwidth}
\tcbox[sharp corners, size = tight, boxrule=0.2mm, colframe=black, colback=white]{
\psfig{figure=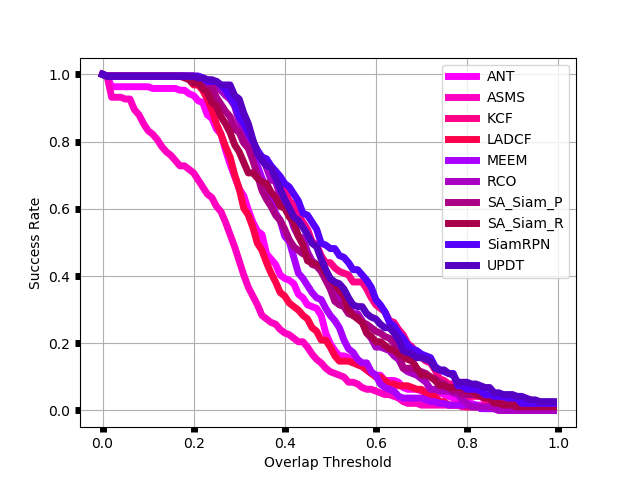, width=0.32\textwidth,height=0.2\textwidth}}}
\centerline{\hspace{0.03\textwidth} Traffic sequence}
\vspace{0.001\textwidth}
\centerline{ 
\tcbox[sharp corners, size = tight, boxrule=0.2mm, colframe=black, colback=white]{
\psfig{figure=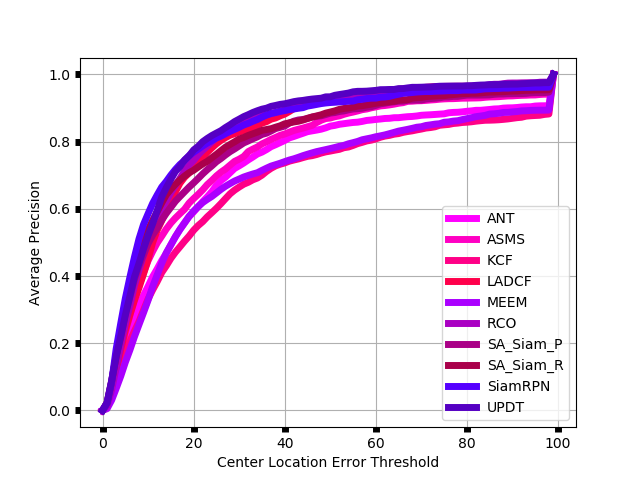, width=0.32\textwidth,height=0.2\textwidth}}
\hspace{0.001\textwidth}
\tcbox[sharp corners, size = tight, boxrule=0.2mm, colframe=black, colback=white]{
\psfig{figure=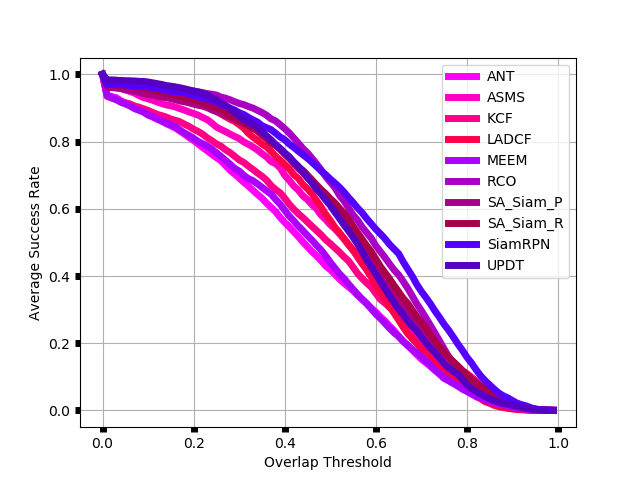, width=0.32\textwidth,height=0.2\textwidth}}
\hspace{0.001\textwidth}
\tcbox[sharp corners, size = tight, boxrule=0.2mm, colframe=black, colback=white]{
\psfig{figure=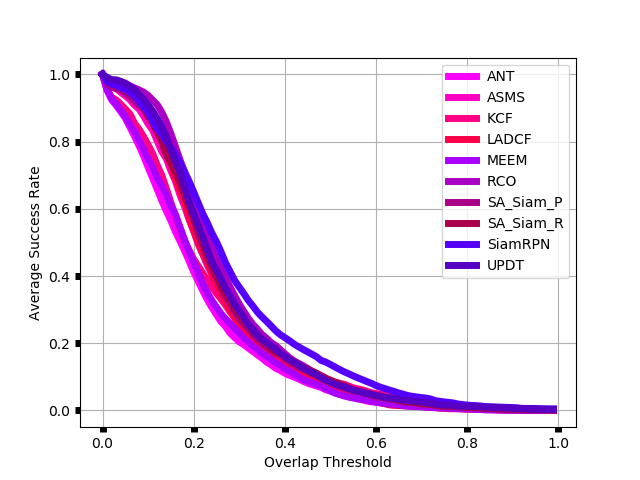, width=0.32\textwidth,height=0.2\textwidth}}}
\centerline{\hspace{0.03\textwidth} Traffic sequence}
\caption{Illustrates the precision and success plots popularly used for evaluating a tracking algorithm and the success plot based on the proposed new measure - matching score of ten different methods on Soccer1 and Traffic sequences. Each plot corresponds to one-pass evaluation ({\sc ope}). \label{figure_plot4}}
\end{figure*}

Figures~\ref{figure_plot1},~\ref{figure_plot2},~\ref{figure_plot3}, and~\ref{figure_plot4} present precision, success, and new success plots correspond to randomly selected video sequences --- Bag, Basketball, Blanket, Bmx, Book, Butterfly, Crabs1, Dinosaur, Fish2, Fernando, Girl, Gymnastics2, Hand, Helicopter, Iceskater2, Matrix, Motocross2, Shaking, Soccer1, and Traffic of {\sc vot-2018} dataset. For these plots, we use {\sc auc} score to summarize the performances of the trackers. Tables~\ref{table_auc_success} and~\ref{table_auc_new_success} highlight {\sc auc} of the success plots of the existing measure and the proposed measure --- matching score. These tables also highlight the {\sc auc} of average success plots over 60 sequences of {\sc vot-2018} dataset. Tables~\ref{table_precision_value},~\ref{table_success_value}, and~\ref{table_new_success_value} highlight the precision at threshold 20 pixels, success rate at threshold value 0.5. The precision and success plots make different conclusions for most of the sequences of {\sc vot-2018} dataset. We make unique conclusion using new success plots of the proposed measure - matching score.

\section{Conclusions}

In this paper, we propose a new measure - matching score to better quantify a tracker for tracking scaled and oriented objects. The proposed measure - matching score is a combination of two existing measures and three new measures. For some sequences containing scaled and oriented objects, two existing measures make two different remarks. It is very difficult to make a conclusion from these remarks. On the other hand, the proposed measure makes a concrete remark which highlights the best tracker. The sufficient examples and results conclude that the proposed measure is more accurate than the existing one to quantify a tracker for tracking scaled and oriented objects.         

\section{Compliance with Ethical Standards}

\subsection{Conflict of Interest}

All authors declare that they have no conflicts of interest.

\subsection{Ethical Approval}

This article does no contain any studies with human participants or animals performed by any of the authors.

\end{document}